\documentclass{article}

\usepackage{datetime2}
\usepackage{arxiv}
\usepackage{float}
\usepackage[utf8]{inputenc} 
\usepackage[T1]{fontenc}    
\usepackage{hyperref}       
\usepackage{url}            
\usepackage{booktabs}       
\usepackage{amsfonts}       
\usepackage{nicefrac}       
\usepackage{microtype}      
\usepackage{lipsum}
\usepackage{graphicx}
\graphicspath{ {./figures/} }
\usepackage{xcolor}
\usepackage{cancel}
\usepackage{amsmath}
\usepackage{bm}
\newcommand{\xdownarrow}[1]{%
  {\left\downarrow\vbox to #1{}\right.\kern-\nulldelimiterspace}
}

\DeclareMathOperator*{\argmin}{arg\,min}

\usepackage[linesnumbered,ruled,vlined]{algorithm2e}

\newcommand{\remark}[1]{\footnote{#1}}

\newcommand{\mbeq}{\overset{!}{=}}

\title{
Towards Rapid Constitutive Model Discovery from Multi-Modal Data: Physics Augmented Finite Element Model Updating (paFEMU)}

\author{
Jingye Tan\\
University of Southern California $\&$\\
Cornell University\\
\texttt{jingyeta@usc.edu}
\And
Govinda Anantha Padmanabha\\
Ecole Polytechnique Federale de Lausanne (EPFL) $\&$\\
Cornell University\\
\texttt{govinda.ananthapadmanabha@epfl.ch}
\And
Steven J. Yang\\
Cornell University\\
\texttt{sjy32@cornell.edu}
\And
Nikolaos Bouklas\\
Cornell University $\&$ Pasteur Labs\\
\texttt{nb589@cornell.edu}
}

\begin{document}
\maketitle
\begin{abstract}

Recent progress in AI-enabled constitutive modeling has concentrated on moving from a purely data-driven paradigm to the enforcement of physical constraints and mechanistic principles, a concept referred to as physics augmentation. Classical phenomenological approaches rely on selecting a pre-defined model and calibrating its parameters, while machine learning methods often focus on discovery of the model itself. Sparse regression approaches lie in between, where large libraries of pre-defined models are probed during calibration. Sparsification in the aforementioned paradigm, but also in the context of neural network architecture, has been shown to enable interpretability, uncertainty quantification, but also heterogeneous software integration due to the low-dimensional nature of the resulting models. Most works in AI-enabled constitutive modeling have also focused on data from a single source, but in reality, materials modeling workflows can contain data from many different sources (multi-modal data), and also from testing other materials within the same materials class (multi-fidelity data). In this work, we introduce physics augmented finite element model updating (paFEMU), as a transfer learning approach that combines AI-enabled constitutive modeling, sparsification for interpretable model discovery, and finite element-based adjoint optimization utilizing multi-modal data. This is achieved by combining simple mechanical testing data, potentially from a distinct material, with digital image correlation-type full-field data acquisition to ultimately enable rapid constitutive modeling discovery. The simplicity of the sparse representation enables easy integration of neural constitutive models in existing finite element workflows, and also enables low-dimensional updating during transfer learning. 
\end{abstract}
\clearpage
\section{Introduction}







Constitutive model identification is essential for solid mechanics simulations, enabling predictive modeling for arbitrarily complex geometries and loading conditions. 
In traditional workflows, engineers select an existing constitutive model form such as a phenomenological model, and fit model parameters of the selected model to experimental data. 
This phenomenological approach relies on user experience and intuition, yet the scientific basis for choosing one model over another is often ad hoc. 
As a result, model development for new materials can be a slow and cumbersome process. 
%
Trustworthiness
\remark{Trustworthiness here, is referred to as the ability of said model to perform in a reasonable (non-erratic) manner when probed in unseen states, whether these interpolate or extrapolate with respect to the training data.} 
of the proposed models is achieved by incorporating strong constraints, from both physics and mechanistic principles, directly in the construction of the models.
Solid mechanics is a data-scarce discipline due to experimental restrictions and cost associated with microscale simulations. Data-sets are usually either small (e.g. few observations/experiments), have restricted observations (e.g. experimentally attainable homogeneous stress states), or partial observations (e.g. displacement observation on specimen surface but no corresponding pointwise stresses).
Machine learning constitutive model discovery aims to automate this process by simultaneously identifying the functional form of a material law and calibrating its parameters from data. 
Such automation is vital for rapid material prototyping in the design cycle, as processing conditions can, for example, significantly affect the response of composites. 
Moreover, discovered models should ideally be interpretable, meaning their parameters and functional forms carry physical meaning. 
Interpretability builds trust in simulations and provides insight (e.g., a learned parameter might correspond to a physical characteristic), which in turn accelerates the design process. 
In summary, a key motivation for data-driven constitutive modeling is to accelerate material characterization for rapid prototyping, while maintaining the trustworthiness and interpretability of classical models.
From a computational perspective, it is useful to distinguish \emph{parameter identification} (calibration within a prescribed constitutive family) from \emph{model discovery} (learning or selecting the constitutive functional form itself). The latter is inherently more ill-posed and typically calls for additional inductive bias, e.g., invariance, potential-based formulations, sparsity, and thermodynamic admissibility, to obtain models that generalize beyond the limited dataset.

%
Over the past few years, significant progress has been made in leveraging machine learning (ML)\cite{goodfellow2016deep, murphy2022probabilistic} to learn constitutive relations from data\cite{Fuhg_2025_DataDrivenReview}. Recent advances in machine learning\cite{louizos2017learning, mcculloch2024sparse} for constitutive models have shown promise in reproducing complex material behavior without assuming a fixed form a priori. 
Purely black-box models can fit data but offer little physical insight
\textcolor{black}{\cite{fuhg2022physics}}. 
%
%
Linka et al. 
\textcolor{black}{\cite{LINKA2023115731}}
introduced Constitutive Artificial Neural Networks (CANNs) that autonomously discover the optimal model form and parameters for biological tissues in the context of hyperelasticity. 
In this referred study, a neural network was trained to identify hyperelastic laws for brain tissue by selecting from a library of classical strain-energy functions (Ogden, Mooney–Rivlin, etc.), effectively blending domain knowledge with deep learning. Such approaches inherit the expressivity of ML while embedding physics-based building blocks to retain interpretability. 
This has spurred interest in techniques that inject physics and/or sparsity into the learning. 
For instance, Fuhg et al. \cite{fuhg2024extreme} demonstrate that Physics-Augmented Neural Networks (PANNs)\footnote{More broadly we refer to PANNs, as neural representations of constitutive laws that have physical information encoded as soft or hard constraints. This term is used broadly throughout the present manuscript.} can be trained on data using $L_0$ regularization\cite{louizos2017learning, yang2025physics} for sparsification, yielding compact, human-readable material laws. This achieves a balance between the flexibility of ML, accuracy and the simplicity of traditional models. 
Complementing the ML-based efforts, sparse regression \cite{tibshirani1996regression,brunton2016discovering} and symbolic regression methods \cite{schmidt2009distilling,doi:10.1126/sciadv.aay2631} directly search for simple algebraic expressions that fit the data.
A notable example is the work of Flaschel et al. 
\textcolor{black}{\cite{flaschel2023automated}}
who proposed the EUCLID framework for unsupervised discovery of constitutive laws. Their method uses only full-field displacements and reaction forces (no stress measurements) to identify an interpretable hyperelastic potential, utilising a curated library of models, that satisfies physics and matches experimental data. 
Their approach was extended for more complex problems in viscoelasticity and plasticity, as well as uncertainty quantification 
\textcolor{black}{\cite{flaschel2023automated, marino2023automated, joshi2022bayesian}}
. 
These approaches
have demonstrated that sparse data-driven models can recover well-known forms 
and discover new ones for new materials, while greatly reducing user bias. 
The emergence of these ML-based and sparse-regression approaches represents a new paradigm in constitutive modeling: rather than tweaking parameters of an assumed model, researchers can now generate candidate models from data and select the best-performing, most interpretable law.


A closely related and highly influential methodology for extracting constitutive information from full-field measurements, often aiming Digital Image Correlation (DIC) experiments \textcolor{black}{\cite{hild2006digital}}, is the \emph{Virtual Fields Method} (VFM) \cite{pierron2012virtual}. VFM is rooted in the principle of virtual work and identifies material parameters by enforcing equilibrium in a weak form using carefully chosen virtual fields. A key practical advantage is that many VFM formulations rely primarily on evaluating equilibrium/virtual-work residuals on measured kinematic fields, thereby reducing the need for repeatedly solving a forward boundary-value problem at every iteration, as is typical in classical Finite Element Model Updating (FEMU)\cite{kumar2025comparative}. Conceptually, EUCLID can be viewed as complementary to VFM: it similarly exploits equilibrium constraints and experimental observables (displacements and global reaction forces), while augmenting the identification step with sparse model selection over a candidate feature library to enable interpretable \emph{model discovery} in addition to parameter calibration. More recent work that invokes neural architectures instead of FEM for the model updating and discovery has been presented in a work termed Automatically Differentiable Model Updating (ADiMU)\cite{ferreira2025automatically}, but also re-casting of FEM in a differentiable setting \cite{regazzoni2026internal}.

Classical adjoint-based inverse methods in solid mechanics provide a foundation for the modern differentiable approach. 
Techniques such as FEMU can formulate parameter identification as a PDE-constrained optimization problem: one seeks to minimize the discrepancy between measured and simulated responses by adjusting model parameters of the PDEs. 
The adjoint method allows efficient computation of the gradient of this discrepancy with respect to all parameters, by solving an auxiliary (adjoint) linear problem
\textcolor{black}{\cite{oberai2003solution}}
.  
%
 Decades of work using FEMU and related methods have shown that incorporating full-field experimental data, such as DIC, can drastically improve the robustness of parameter calibration
 \textcolor{black}{\cite{avril2008overview}}
 . 
%
 In a typical workflow, given a known constitutive model, one iteratively updates parameters (such as elastic constants or hardening moduli) using a gradient-based optimizer, where the gradient is supplied by the adjoint solution at each step. 
This yields local sensitivity information, which is very powerful for refining an initial guess. 
However, being a local method, the success of adjoint-based calibration depends on a good initial model and on the data capturing enough independent deformation modes to constrain all parameters. 
If, for example, only a narrow range of loading conditions is observed, the inverse problem may be ill-posed or lead to a local minimum that fits those conditions but not others. 
Recent studies underline the importance of informative experiments and multi-modal data for unique identification
\textcolor{black}{\cite{nikolov2025variation}}
. 
%
 Approaches like the aforementioned Virtual Fields Method 
 \textcolor{black}{\cite{pierron2012virtual}}
 and Bayesian identification 
 \textcolor{black}{\cite{gaynutdinova2023bayesian}}
 seek to mitigate this by optimal experimental design and uncertainty quantification, respectively. 
In summary, adjoint/PDE-constrained methods offer an elegant and computationally efficient route to calibrate known-form models, but they too encounter difficulties in global model discovery or when confronted with sparse data. These insights set the stage for our work: we aim to improve model discovery by leveraging prior knowledge (to inform the model form and initialization) and by effectively using every bit of data through a transfer learning framework.


%
Classical FEMU typically entails repeated forward solves (and often adjoint solves) inside an outer optimization loop, which can become prohibitive when the parameter space is large and non-convex and/or when the forward model is expensive. 
In contrast, residual-based identification approaches (e.g., the Virtual Fields Method) can partially bypass this bottleneck by operating primarily on equilibrium/virtual-work residual evaluations on measured kinematic fields, rather than requiring a full-field simulation on every iteration.
More broadly, enabling scalable calibration and optimization pipelines in computational mechanics increasingly calls for \emph{end-to-end} differentiability and heterogeneous software tool integration that can seamlessly combine distinct numerical and hardware strategies (e.g., mixing CPU-based solvers with GPU-accelerated kernels, or recasting finite element operators in GPU-native form). Several promising routes have recently been suggested. 
On one end, PINN-type surrogate solvers recast calibration as a joint learning problem over the state field and unknown parameters: a neural network is trained to approximate the displacement (or stress) field while treating constitutive parameters as trainable variables. 
Physics is enforced by coupling the training loss with weighted residual terms of the governing equations (e.g., equilibrium, compatibility, boundary conditions), along with data misfit terms at measurement locations or over full-field observations.
In this way, inverse identification can proceed without repeatedly calling an external FE solver inside an outer optimization loop; instead, the physics is implicitly encoded through the training objective. 
Hamel et al.\cite{hamel2023calibrating} demonstrated this idea for constitutive calibration from full-field data, showing that PINN-based inverse formulations can reduce reliance on dense experimental labeling and can leverage automatic differentiation for sensitivity information, at the cost of nontrivial loss-balancing, optimizer tuning, and potentially expensive training when high-fidelity discretizations or sharp localization features are present.
On the other end, differentiable simulation frameworks and differentiable solvers seek to differentiate through the discretized physics itself, i.e., to expose gradients of quantities of interest with respect to material parameters, loads, geometry, or even design variables by differentiating the discrete residual and its solution map. 

In contrast to PINN-type neural solvers, differentiable solvers typically retain the numerical structure of the PDE discretization, for example, by implementing residual assembly and linear/nonlinear solves in an automatic differentiation (AD) programming model, or by coupling automatic differentiation with implicit differentiation/adjoint ideas at the discrete level. 
Representative directions include differentiable finite-element-style formulations \cite{fan2024differentiable}, differentiable physics and learning pipelines that emphasize stable gradient propagation through time-stepping and PDE operators \cite{holl2024bf}, and GPU-native differentiable physics engines designed for end-to-end optimization with high-throughput kernels (e.g., NVIDIA Warp \cite{macklin2024warp}). 
Collectively, these approaches aim to make gradient information a first-class output of the simulation, enabling large-scale inverse problems and design loops while facilitating heterogeneous compute strategies (CPU/GPU) and modular toolchains.

\begin{figure}[H]
    \centering
    \includegraphics[width=1\linewidth]{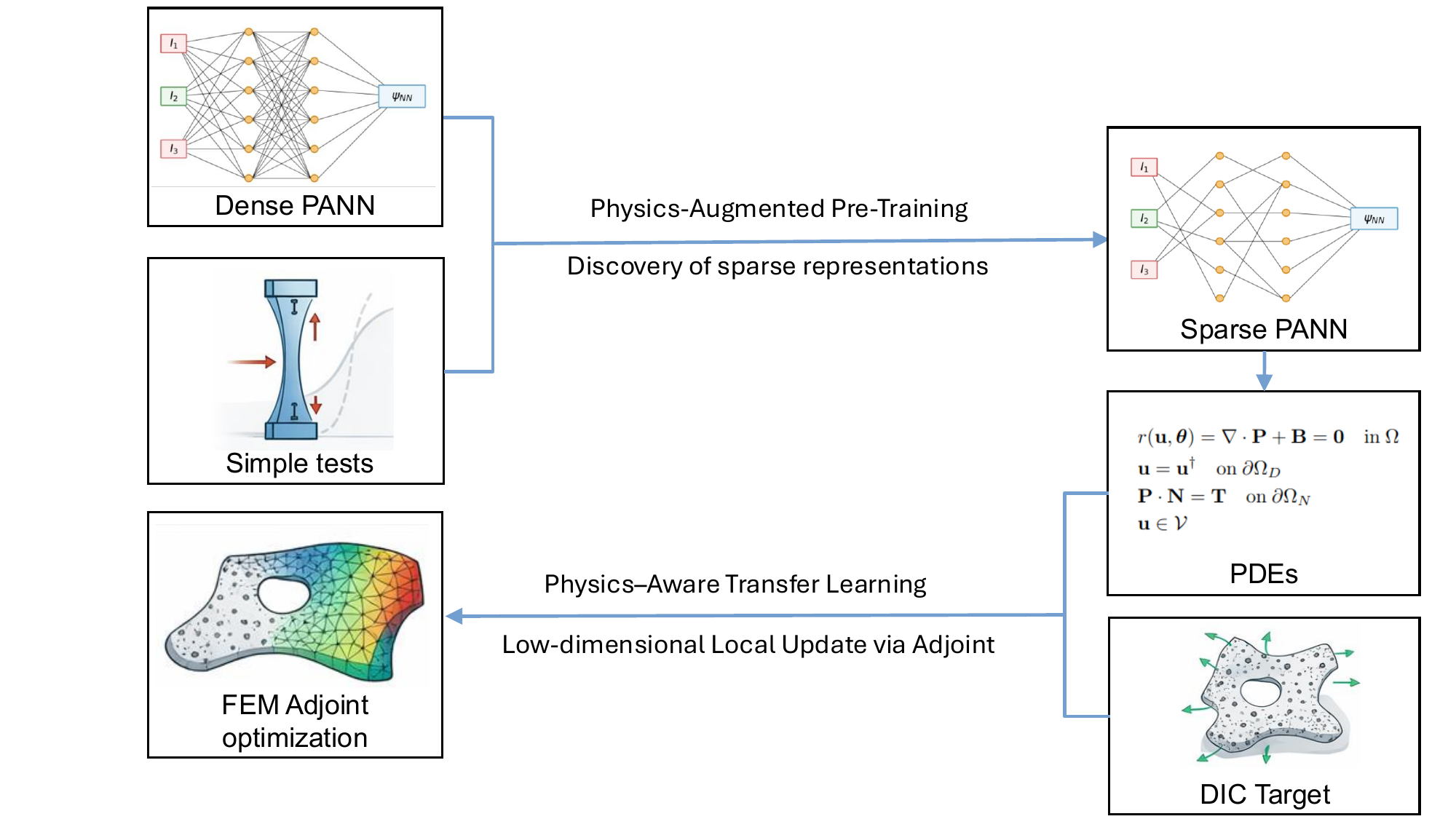}
    \caption{Outline of the paFEMU framework and corresponding multi-modal transfer learning scheme}
    \label{fig:transfer_learning}
\end{figure}

More broadly, differentiable frameworks \cite{baydin2018automatic}, aim to obtain the information captured through the adjoint, in an automated manner. 
The whole construction of ML tools is based on differentiable frameworks that enable efficient training. 
Parallel to advances in material modeling, the computational mechanics community has embraced differentiable finite element methods (FEM) to accelerate simulation and inverse analysis. 
Differentiable FEM solvers, notably those built on automatic differentiation frameworks like JAX 
\textcolor{black}{\cite{xue2023jax}}
,
allow one to compute not only the forward solution of a boundary-value problem, but also exact gradients of any output with respect to material parameters or input conditions 
\textcolor{black}{\cite{xue2023jax, wu2023framework, wu2024jax, hu2025efficient}}
. 
%
For the problem of interest here, this means one can calibrate constitutive model parameters (whether they correspond to a phenomenological or ML-enabled model) by directly minimizing the error between FE-predicted and experimental responses, without manually deriving sensitivity equations. 
The potential of such end-to-end differentiable mechanics solvers has been showcased in topology optimization, multi-scale modeling, and data-driven material law fitting \textcolor{black}{\cite{xue2023jax}}. 
%
However, there are important limitations to note. 
While differentiable solvers provide local gradients that greatly aid optimization, they do not by themselves guarantee a globally optimal model identification. 
If the chosen material model is overly complex and non-convex (e.g., a high-dimensional neural network) and the available data is limited, the calibration may converge to a set of parameters that fits the training cases but fails to generalize (e.g., extrapolation). This is a classic pitfall in small-data regimes: many different material laws might explain a few tests equally well, so a naive gradient-based fit could latch onto a spurious solution. 
%
%
In essence, a differentiable FEM can guide parameter calibration, but it cannot assess model form error or data availability. 
Ensuring generalizable, physics-consistent models from limited data still requires careful regularization (e.g. enforcing thermodynamic laws) or intelligent data collection. 

The aforementioned challenges motivate the approach taken in this work, which couples differentiable solvers with a transfer learning strategy to make the most of limited datasets.
In this context, we pursue transfer learning for constitutive model discovery, utilizing multi-modal and multi-fidelity data. Namely, considering basic mechanical tests in a variety of loading conditions, and advanced imaging tests with complex sample geometries. Transfer learning, a concept rooted in machine learning, utilizes a pre-trained model  and adapts it to a new but related task, rather than starting from scratch. This approach can significantly reduce the required training data and computational cost. In our suggested approach, the initial training utilizes sparsification to enable transitioning from high-dimensional to low-dimensional representations. Following, in the final training stage, the low-dimensional representation is utilized as a starting point for the adjoint-based optimization. 

In mechanics, transfer learning can be extremely powerful: material laws learned from one set of experiments (or from simulated data) can serve as a starting point for new materials that share similar constitutive behavior. For example, one might pre-train a neural constitutive model on a broad class of elastomers (data of lower fidelity), then fine-tune it with a small amount of data from a new polymer to quickly obtain an accurate model for the new material. The rationale is that a materials class might share the same  stress–strain trends, and invariant dependencies, so a model that has captured these patterns is a candidate for re-training for a new case
\textcolor{black}{\cite{pan2009survey}}
. 
%
This is especially helpful in scenarios with limited data due to acquisition complexity, high cost or both. For instance, in biomedical applications where tissue samples are scarce, or in high-rate experiments where instrumentation is challenging \textcolor{black}{\cite{faber2022tissue, faber2022correction, church2014using}}. 
Instead of performing a battery of complex tests on the new material, one could rely on knowledge distilled from previous materials. 
Recent work demonstrates the benefit of transfer learning in constitutive modeling: An RNN-based plasticity model trained on simulated data could be adapted via transfer learning to match experimental soil behavior with a much smaller dataset 
\textcolor{black}{\cite{heidenreich2024transfer}}
%
Similarly, 
Liu et al. \textcolor{black}{\cite{zhu2024transfer}}
proposed a transfer-learning-enhanced physics-informed neural network for identifying soft tissue parameters, achieving faster convergence by starting from a pre-trained network.

In this work, we propose a Physics Augmented Finite Element Model Updating (paFEMU), outlined in Fig. \ref{fig:transfer_learning}. In our approach, we build on this idea by pre-training a PANN model on a set of low-fidelity experiments capturing simple stress states. By employing $L_0$ regularization, we promote sparsity in the learned PANN representation, resulting in a compact and interpretable constitutive description. The model is subsequently fine-tuned using an auto-differentiable finite element-based adjoint formulation on high-fidelity experiments involving complex geometries and heterogeneous stress states measured via Digital Image Correlation (DIC).
%
%
Crucially, our framework is certifiable: we incorporate physics through the architecture of PANNs, and additionally PDE constraints and validation checks so that the transferred model satisfies physical and mechanistic principles, and to fall within error bounds on the target data. By doing so, we address a common concern with transfer learning, namely, that the adapted model might violate physics or predict outside the new training range. In essence, transfer learning in this work serves as a bridge between small-data regimes and complex model requirements, allowing us to calibrate rich constitutive models with minimal additional experiments.

The remainder of this paper is organized as follows: Section 2 introduces the continuum framework for hyperelasticity and develops a polyconvexity indicator that is straightforward to implement within differentiable settings and imposed as a soft constraint. Section 3 presents the design of physics-augmented neural network architectures that enforce thermodynamic consistency, focusing on polyconvexity, and approaches to induce sparsity. Section 4 details the proposed paFEMU framework, combining sparsified pre-training with physics-aware transfer learning across multi-modal datasets. Section 5 outlines the differentiable finite element implementation and adjoint-based optimization strategy used for model updating. Section 6 demonstrates the performance of the framework on representative numerical examples and experimental datasets, highlighting accuracy, interpretability, and data efficiency. Finally, Section 7 concludes the paper with a discussion of key findings, limitations, and future research directions.

\section{Hyperelasticity}\label{sec:hyperelas}
This section, summarizes the continuum framework for hyperelasticity that underpins the proposed learning tasks. Further, polyconvexity is introduced, and an easy to implement polyconvexity indicator is developed. 
\subsection{Kinematics and Thermodynamics}

For a body occupying  $\Omega_0$, we first outline the kinematics of motion. The deformation gradient tensor is defined as
\begin{equation}
    \mathbf{F}(\mathbf{X}) = \frac{\partial \mathbf{x}}{\partial \mathbf{X}},
\end{equation}
where $\mathbf{X}$ and $\mathbf{x}$ represent the position vectors occupying the reference configuration $\Omega_0$ and the current configuration $\Omega$, respectively. 
For the deformation to be physically representative, $\mathbf{F}$ must be invertible. 
A special case of deformation gradient tensor taking the form of the identity tensor, $\mathbf{F} = \mathbf{I}$ indicates the undeformed state.
The right-Cauchy-Green deformation tensor is defined as
\begin{equation}
    \mathbf{C} = \mathbf{F}^T \, \mathbf{F},
\end{equation}
The three principal invariants of the right-Cauchy-Green deformation tensor are,
\begin{equation}
    I_1 = \mathrm{tr}\mathbf{C},\ 
    I_2 = \frac{1}{2}\big[ (\mathrm{tr}\mathbf{C})^2 - \mathrm{tr}(\mathbf{C}:\mathbf{C}) \big],\ 
    I_3 = \det\mathbf{C}.
\end{equation}
Note that the Jacobian $J = \det\mathbf{F} = \sqrt{I_3}$ is more commonly used in place of $I_3$. Further, we follow the use of $J$ for the rest of this manuscript.
%
With the assumption of material isotropy, we require the existence of a specific strain energy density function $\varphi$.
The first law of thermodynamics requires that the rate of change of total energy equals external work for an adiabatic deformable solid,
\begin{equation}
    \dot{\varphi} = \mathbf{P}:\dot{\mathbf{F}},
\end{equation}
where $\mathbf{P}$ denotes the first Piola-Kirchoff stress tensor, the energetic conjugate of $\mathbf{F}$.
The second law of thermodynamics enforces non-negative entropy production by the Clausius-Duhem inequality of the mechanical dissipation,
\begin{equation}
    \mathcal{D} = \mathbf{P}:\dot{\mathbf{F}} - \dot{\varphi} \geq 0\,.
\end{equation}
Thus, in the context of  purely elastic deformation, the dissipation must be zero, which leads to,
\begin{equation}
    \mathbf{P} = \frac{\partial \varphi}{\partial \mathbf{F}},
    \label{eq:stress_from_phi}
\end{equation}
implying that $\varphi$ must be a non-trivial function of $\mathbf{F}$.
By the principle of objectivity, the material description and its response to stress and strain,  is independent of observers, this requires for any orthogonal tensor $\mathbf{R}$ of coordinate transformation, the candidate function $\varphi$ 
\begin{equation}
    \varphi (\mathbf{R}\mathbf{F}) \mbeq \varphi(\mathbf{F})
    , 
    \quad
    \mathbf{P}(\mathbf{R}\mathbf{F}) = \frac{\partial\varphi(\mathbf{R}\mathbf{F})}{\partial\mathbf{F}}
    \mbeq
    \frac{\partial\varphi(\mathbf{F})}{\partial\mathbf{F}}
    =
    \mathbf{R}
    \mathbf{P}(\mathbf{F})\,.
\end{equation}
The aforementioned assumption of isotropy requires that the candidate function $\varphi$ to remain invariant to material symmetry,
\begin{equation}
    \varphi (\mathbf{F}\mathbf{R}^T) \mbeq \varphi(\mathbf{F})
    , 
    \quad
    \mathbf{P}(\mathbf{F}\mathbf{R}^T) = \frac{\partial\varphi(\mathbf{F}\mathbf{R}^T)}{\partial\mathbf{F}}
    \mbeq
    \frac{\partial\varphi(\mathbf{F})}{\partial\mathbf{F}}
    =
    \mathbf{P}(\mathbf{F})
    \mathbf{R}^T\,.
\end{equation}
Additionally, at the undeformed configuration ($\mathbf{F}=\mathbf{I}$), the candidate $\varphi$ can be chosen to be energy-free, 
\begin{equation}
    \varphi(\mathbf{I}) \mbeq 0,
\end{equation}
and must be stress-free,
\begin{equation}
    \mathbf{P}(\mathbf{I}) = \frac{\partial\varphi(\mathbf{I})}{\partial\mathbf{F}}
    \mbeq
    \mathbf{0}.
\end{equation}
While for completeness, at the limit of infinitely large deformation ($\det\mathbf{F}\rightarrow\infty$ for infinite expansion and $\det\mathbf{F}\rightarrow0^+$ for infinite depletion), the candidate function shall grow into infinity coersively, it is uncommon to encounter such deformation modes in practice.
The strain energy density $\varphi(\mathbf{F})$ is  taken to be non-negative for all deformation states $\varphi(\mathbf{F}) \ge 0$ for all admissible $\mathbf{F}$.
This ensures no spontaneous energy release in the absence of external work. 

 \subsection{Polyconvexity}
In nonlinear elasticity, the stored energy function $\varphi(\mathbf{F})$, as formulated for isotropic materials,  is usually not convex in $\mathbf{F}$ over all deformations. 
Polyconvexity is a sufficient (though not necessary) condition to guarantee weak lower-semicontinuity of the energy functional and thus the existence of minimizers (equilibrium solutions) under appropriate growth conditions. Polyconvexity is a weaker convexity condition introduced by Ball (1976) \cite{ball1976convexity} to ensure existence of energy minimizers.
Formally, $\varphi(\mathbf{F})$ is polyconvex if it can be written as a convex function of $\mathbf{F}$ and all of its minors (determinants of sub-matrices).
For example, in 3D there exists a convex $\varphi(\mathbf{F})$ such that $\varphi(\mathbf{F})=\varphi(\mathbf{F},\mathrm{cof}\,\mathbf{F},\det \mathbf{F})$.
 Polyconvexity implies weaker convexity conditions like quasiconvexity and rank-one convexity, which in turn are related to the existence of minimizers and absence of short wavelength instabilities accordingly, but not vice-versa. 
Many standard hyperelastic models (e.g. Mooney–Rivlin, Ogden) satisfy polyconvexity, but others do not (e.g. Gent). Requiring polyconvexity, is a convenient restriction when the material being studied is observed to exhibit a stable response, alleviating issues that could otherwise arise in finite element modeling.

In the present context, polyconvexity plays a dual role: it ensures well-posedness of the underlying boundary value problem, and provides a natural mechanism to regularize neural constitutive models toward physically admissible responses.

\subsection{A Polyconvexity Indicator}

In parallel to strategies that enforce polyconvexity, we derive a reduced set of necessary conditions that can be efficiently evaluated.

If $\varphi$ is a polyconvex function for isotropic materials, it means that it can be written as $\varphi = g(\mathbf{F}, \mathrm{cof}\,\mathbf{F}, \det\mathbf{F})$, and is convex with respect to each argument individually \cite{ball1976convexity}. It is equivalent \cite{steigmann2003isotropic,neff2015exponentiated,tepole2025polyconvex} to recasting this function as 
\begin{equation}
    \varphi = h(\lambda_1, \lambda_2, \lambda_3, \lambda_1\lambda_2, \lambda_1\lambda_3, \lambda_2\lambda_3, J), \label{eq::lambda7}
\end{equation}
or, 
\begin{equation}
    \varphi = h(\lambda_1, \lambda_2, \lambda_3, \lambda_4, \lambda_5, \lambda_6, \lambda_7), 
\end{equation}
where $h$ is convex with respect to its arguments e.g. $\partial^2\varphi/\partial\lambda_i^2 \geq 0$ for $i = 1,2,...,7$, where
\begin{equation}
    \lambda_4 = \lambda_1\lambda_2, \quad
    \lambda_5 = \lambda_1\lambda_3, \quad
    \lambda_6 = \lambda_2\lambda_3, \quad
    \lambda_7 = J,
\end{equation}
but also invariant over permutations of $i = 1,2,3$ and $i = 4,5,6$.

For a potential (commonly seen in NN-based formulations), $\hat{\varphi} (I_1, I_2, J)$ to be a polyconvex function, utilizing \ref{eq::lambda7}, the following identities:
\begin{equation}
    \begin{aligned}
        I_1 &= \lambda_1^2 + \lambda_2^2 + \lambda_3^2,
        \\
    I_2 &= \lambda_1^2\lambda_2^2
    +
    \lambda_1^2\lambda_3^2
    +
    \lambda_2^2\lambda_3^2
    =
    \lambda_4^2+\lambda_5^2+\lambda_6^2
    ,
    \\
    J &= \lambda_1\lambda_2\lambda_3 = \lambda_7,
    \end{aligned}\label{eq:invs}
\end{equation}
the potential $\hat{\varphi}$ must be convex in $\lambda_i$ for $i = 1,2,...,7$, leading to the following inequalities:

\begin{equation}
    \begin{aligned}
        \frac{\partial^2 \hat{\varphi}}{\partial \lambda_1^2}
    &=
    \frac{\partial}{\partial \lambda_1}
    \bigg(
        \frac{\partial \hat{\varphi}}{\partial I_1}
        \frac{\partial I_1}{\partial \lambda_1}
    \bigg)
    =
    \frac{\partial}{\partial \lambda_1}
    \bigg(
        \frac{\partial \hat{\varphi}}{\partial I_1}
        \,2\,\lambda_1
    \bigg)
    \\&
    =
    \frac{\partial^2 \hat{\varphi}}{\partial I_1^2}
    \frac{\partial I_1}{\partial \lambda_1}
    \,2\,\lambda_1
    +
    \frac{\partial \hat{\varphi}}{\partial I_1}
    \frac{\partial (2\,\lambda_1) }{\partial \lambda_1}
    =
    4\,\lambda_1^2\,
    \frac{\partial^2 \hat{\varphi}}{\partial I_1^2}
    +
    2\, \frac{\partial \hat{\varphi}}{\partial I_1}
    \geq 0
    \end{aligned}
    \label{eq:I1_lambda_1}
\end{equation}

\begin{equation}
    \begin{aligned}
        \frac{\partial^2 \hat{\varphi}}{\partial \lambda_2^2}
    &=
    \frac{\partial}{\partial \lambda_2}
    \bigg(
        \frac{\partial \hat{\varphi}}{\partial I_1}
        \frac{\partial I_1}{\partial \lambda_2}
    \bigg)
    =
    \frac{\partial}{\partial \lambda_2}
    \bigg(
        \frac{\partial \hat{\varphi}}{\partial I_1}
        \,2\,\lambda_2
    \bigg)
    \\&
    =
    \frac{\partial^2 \hat{\varphi}}{\partial I_1^2}
    \frac{\partial I_1}{\partial \lambda_2}
    \,2\,\lambda_2
    +
    \frac{\partial \hat{\varphi}}{\partial I_1}
    \frac{\partial (2\,\lambda_2) }{\partial \lambda_2}
    =
    4\,\lambda_2^2\,
    \frac{\partial^2 \hat{\varphi}}{\partial I_1^2}
    +
    2\, \frac{\partial \hat{\varphi}}{\partial I_1}
    \geq 0
    \end{aligned}
    \label{eq:I1_lambda_2}
\end{equation}

\begin{equation}
    \begin{aligned}
        \frac{\partial^2 \hat{\varphi}}{\partial \lambda_3^2}
    &=
    \frac{\partial}{\partial \lambda_3}
    \bigg(
        \frac{\partial \hat{\varphi}}{\partial I_1}
        \frac{\partial I_1}{\partial \lambda_3}
    \bigg)
    =
    \frac{\partial}{\partial \lambda_3}
    \bigg(
        \frac{\partial \hat{\varphi}}{\partial I_1}
        \,2\,\lambda_3
    \bigg)
    \\&
    =
    \frac{\partial^2 \hat{\varphi}}{\partial I_1^2}
    \frac{\partial I_1}{\partial \lambda_3}
    \,2\,\lambda_3
    +
    \frac{\partial \hat{\varphi}}{\partial I_1}
    \frac{\partial (2\,\lambda_3) }{\partial \lambda_3}
    =
    4\,\lambda_3^2\,
    \frac{\partial^2 \hat{\varphi}}{\partial I_1^2}
    +
    2\, \frac{\partial \hat{\varphi}}{\partial I_1}
    \geq 0
    \end{aligned}
    \label{eq:I1_lambda_3}
\end{equation}

\begin{equation}
    \begin{aligned}
        \frac{\partial^2 \hat{\varphi}}{\partial \lambda_4^2}
    &=
    \frac{\partial}{\partial \lambda_4}
    \bigg(
        \frac{\partial \hat{\varphi}}{\partial I_2}
        \frac{\partial I_2}{\partial \lambda_4}
    \bigg)
    =
    \frac{\partial}{\partial \lambda_4}
    \bigg(
        \frac{\partial \hat{\varphi}}{\partial I_2}
        \,2\,\lambda_4
    \bigg)
    \\&
    =
    \frac{\partial^2 \hat{\varphi}}{\partial I_2^2}
    \frac{\partial I_2}{\partial \lambda_4}
    \,2\,\lambda_4
    +
    \frac{\partial \hat{\varphi}}{\partial I_2}
    \frac{\partial (2\,\lambda_4) }{\partial \lambda_4}
    =
    4\,\lambda_4^2\,
    \frac{\partial^2 \hat{\varphi}}{\partial I_2^2}
    +
    2\, \frac{\partial \hat{\varphi}}{\partial I_2}
    \geq 0
    \end{aligned}
    \label{eq:I2_lambda_4}
\end{equation}

\begin{equation}
    \begin{aligned}
        \frac{\partial^2 \hat{\varphi}}{\partial \lambda_5^2}
    &=
    \frac{\partial}{\partial \lambda_5}
    \bigg(
        \frac{\partial \hat{\varphi}}{\partial I_2}
        \frac{\partial I_2}{\partial \lambda_5}
    \bigg)
    =
    \frac{\partial}{\partial \lambda_5}
    \bigg(
        \frac{\partial \hat{\varphi}}{\partial I_2}
        \,2\,\lambda_5
    \bigg)
    \\&
    =
    \frac{\partial^2 \hat{\varphi}}{\partial I_2^2}
    \frac{\partial I_2}{\partial \lambda_5}
    \,2\,\lambda_5
    +
    \frac{\partial \hat{\varphi}}{\partial I_2}
    \frac{\partial (2\,\lambda_5) }{\partial \lambda_5}
    =
    4\,\lambda_5^2\,
    \frac{\partial^2 \hat{\varphi}}{\partial I_2^2}
    +
    2\, \frac{\partial \hat{\varphi}}{\partial I_2}
    \geq 0
    \end{aligned}
    \label{eq:I2_lambda_5}
\end{equation}

\begin{equation}
    \begin{aligned}
        \frac{\partial^2 \hat{\varphi}}{\partial \lambda_6^2}
    &=
    \frac{\partial}{\partial \lambda_6}
    \bigg(
        \frac{\partial \hat{\varphi}}{\partial I_2}
        \frac{\partial I_2}{\partial \lambda_6}
    \bigg)
    =
    \frac{\partial}{\partial \lambda_6}
    \bigg(
        \frac{\partial \hat{\varphi}}{\partial I_2}
        \,2\,\lambda_6
    \bigg)
    \\&
    =
    \frac{\partial^2 \hat{\varphi}}{\partial I_2^2}
    \frac{\partial I_2}{\partial \lambda_6}
    \,2\,\lambda_6
    +
    \frac{\partial \hat{\varphi}}{\partial I_2}
    \frac{\partial (2\,\lambda_6) }{\partial \lambda_6}
    =
    4\,\lambda_6^2\,
    \frac{\partial^2 \hat{\varphi}}{\partial I_2^2}
    +
    2\, \frac{\partial \hat{\varphi}}{\partial I_2}
    \geq 0
    \end{aligned}
    \label{eq:I2_lambda_6}
\end{equation}

\begin{equation}\label{eq:polyconvex_J}
    \begin{aligned}
        \frac{\partial^2 \hat{\varphi}}{\partial \lambda_7^2}
    &=
    \frac{\partial}{\partial \lambda_7}
    \bigg(
        \frac{\partial \hat{\varphi}}{\partial J}
        \cancelto{1}{\frac{\partial J}{\partial \lambda_7}}
    \bigg)
    =
    \frac{\partial^2 \hat{\varphi}}{\partial J^2}
    \cancelto{1}{\frac{\partial J}{\partial \lambda_7}}
    =
    \frac{\partial^2 \hat{\varphi}}{\partial J^2}
    \geq 0
    \end{aligned}
\end{equation}


By combining Eqs.\eqref{eq:I1_lambda_1}-\eqref{eq:I1_lambda_3} and  Eqs.\eqref{eq:I2_lambda_4}-\eqref{eq:I2_lambda_6}, and consequently utilizing Eq.\eqref{eq:invs} the two following inequalities can be obtained
\begin{equation}\label{eq:polyconvex_I1}
\sum_{k=1}^{k=3}
\frac{\partial^2 \hat{\varphi}}{\partial \lambda_k^2}
=
\sum_{k=1}^{k=3}
\bigg(
    4\,\lambda_k^2\,
    \frac{\partial^2 \hat{\varphi}}{\partial I_1^2}
    +
    2\, \frac{\partial \hat{\varphi}}{\partial I_1}
\bigg)
=
\frac{\partial^2 \hat{\varphi}}{\partial I_1^2}
+
\frac{3}{2\,I_1}\, \frac{\partial \hat{\varphi}}{\partial I_1}
    \geq 0
\end{equation}

\begin{equation}\label{eq:polyconvex_I2}
\sum_{j=4}^{j=6}
\frac{\partial^2 \hat{\varphi}}{\partial \lambda_j^2}
=
\sum_{j=4}^{j=6}
\bigg(
    4\,\lambda_j^2\,
    \frac{\partial^2 \hat{\varphi}}{\partial I_2^2}
    +
    2\, \frac{\partial \hat{\varphi}}{\partial I_2}
\bigg)
=
\frac{\partial^2 \hat{\varphi}}{\partial I_2^2}
+
\frac{3}{2\,I_2}\, \frac{\partial \hat{\varphi}}{\partial I_2}
    \geq 0
\end{equation}

The resulting reduced set of inequalities Eq.\eqref{eq:polyconvex_I1} and Eq.\eqref{eq:polyconvex_I2} are necessary but not sufficient for the simultaneous fulfillment of all of Eqs.\eqref{eq:I1_lambda_1}-\eqref{eq:I2_lambda_6}, however, are very convenient to compute in the context of a NN using automatic differentiation (as will be further discussed in the upcoming section) as they don't involve eigenvalue computation from $\mathbf{F}$ or explicitly checking for loss of ellipticity. 
Thus, Eq.\eqref{eq:polyconvex_I1} and Eq.\eqref{eq:polyconvex_I2} can be taken together with Eq.\eqref{eq:polyconvex_J} to form an indicator for polyconvexity for the trainable function $\hat{\varphi}$. 
%


\section{Data-Driven Constitutive Models}\label{sec:NN}

In place of traditional constitutive models, neural networks offer significant flexibility to capture the function forms of the underlying constitutive relationship thanks to over-parametrization. However, not all neural networks are suitable candidates for constitutive modeling. In this section, we discuss a methodology to designing  physics augmented neural networks for learning constitutive models.

\subsection{Input Convex Neural Network Architectures and Polyconvexity}

Input-Convex Neural Networks (ICNNs) \cite{amos2017input} provide a neural architecture whose output is guaranteed to be convex with respect to its inputs.  In the present work, this property is used to construct strain-energy potentials with built-in stability constraints. We consider a feed-forward architecture with input pass-through (skip connections), defining a nonlinear map  $\mathcal{N}:\mathbb{R}^{n^{0}}\rightarrow\mathbb{R}^{n^{L}}$ as

\begin{equation}\label{eq:FNN}
    \boldsymbol{y} = \mathcal{N}(\boldsymbol{x}) \equiv
    \begin{cases}
            \boldsymbol{z}_{0} &= 0 
            \\
            \boldsymbol{z}_{l+1} 
            &= 
            f_{l} \left(  \boldsymbol{W}_{l} \boldsymbol{z}_{l}
            + 
             \boldsymbol{\mathcal{W}}_{l} \boldsymbol{x}
            + 
            \boldsymbol{b}_{l} \right), \qquad l=0, \ldots, L-1 
            \\
            \boldsymbol{y} &=  \boldsymbol{W}_{L} \boldsymbol{z}_{L}
            +   \boldsymbol{\mathcal{W}}_{L} \boldsymbol{x}
            + \boldsymbol{b}_{L}.
    \end{cases}
\end{equation}
Here, $\boldsymbol{W}_l$ propagates hidden features across layers, while  $\boldsymbol{\mathcal{W}}_l$ injects the input directly into each layer through skip connections.
Convexity of $\mathcal{N}(\boldsymbol{x})$ with respect to $\boldsymbol{x}$ is sufficiently ensured when (i) the activation functions $f_l$ are convex and non-decreasing, and 
(ii) the hidden-state weights satisfy $\boldsymbol{W}_l \ge 0$ element-wise. Under these conditions, the network output is convex in the input \cite{amos2017input}.
Convexity is preserved because each layer forms a non-negative linear combination of convex functions of the previous layer together with affine functions of the input, followed by composition with a monotone convex activation. ICNNs, therefore, approximate convex functions while maintaining convexity by construction.
\par

Polyconvex Neural Networks (PCNNs) \cite{klein2022polyconvex,kleinpolyconvex} are obtained by combining Monotonic Neural Networks (MNNs) \cite{fuhg2023modular,klein2025neural}, which enforce monotonicity without requiring convex activation functions, with ICNNs. Within this invariant representation, stability requirements such as polyconvexity can be promoted by enforcing monotonic and convex dependence of the strain-energy function with respect to the isochoric invariants $I_1$ and $I_2$, while the volumetric invariant $I_3$ (or $J=\operatorname{det} \bm{F}$ ) remains convex.
Within the architecture Eq.\eqref{eq:FNN}, PCNNs are obtained by constraining the input and skip-connection weights associated with those invariant inputs with respect to which convexity is enforced. In the present formulation, convexity is imposed with respect to the isochoric invariants $I_1$ and $I_2$, which is achieved by enforcing element-wise non-negativity of the corresponding weights, i.e., $\bm{W}_l^T \geq 0$. The volumetric invariant $I_3$ (or equivalently $J=\operatorname{det} \bm{F}$ ) is introduced as an monotonic input, and therefore the weights associated with this variable are not restricted to be positive. 
%
%
A comparison of these neural network variants is summarized in Table~\ref{tab:nn_compare}.  In this work, we investigate relaxations of these constraints to improve model expressivity while retaining essential stability properties.

\begin{table}[h]
\centering
\caption{Comparison of feed-forward neural  (FNN) variants in the context of constitutive model discovery using invariant-based formulations. }
\label{tab:nn_compare}
\renewcommand{\arraystretch}{1.25}
\begin{tabular}{p{1.5cm} p{2.4cm} p{3.4cm} p{2.4cm} p{3.4cm}}
\hline
 & \textbf{FNN~\cite{goodfellow2016deep}} & \textbf{ICNN~\cite{amos2017input}} & \textbf{MNN~\cite{klein2025neural}} & \textbf{PCNN} \\
\hline
$f_l(\cdot)$ &
Unconstrained & Convex, nondecreasing  & Nondecreasing  & Convex, nondecreasing  \\

$\boldsymbol{W}_l$ & Unconstrained & $\ge \boldsymbol{0}$ & $\ge \boldsymbol{0}$ & $\ge \boldsymbol{0}$ \\

$\boldsymbol{\mathcal{W}}_{l}^{(I_1,I_2)}$ & Unconstrained & Unconstrained & $\ge \boldsymbol{0}$ & $\ge \boldsymbol{0}$ \\

$\boldsymbol{\mathcal{W}}_{l}^{(J)}$ & Unconstrained & Unconstrained & Unconstrained & Unconstrained \\

$\boldsymbol{b}_l$ & Unconstrained & Unconstrained & Unconstrained & Unconstrained 
\\
\hline
Polyconvex guarantee
&
Not sufficient
&
Necessary but not sufficient
&
Necessary but not sufficient
&
Sufficient but not necessary
\\
\hline
\end{tabular}
\end{table}

In the context of ICNNs, the second derivatives of the network w.r.t. its inputs are always positive, thus Eq.\eqref{eq:polyconvex_J} is inherently satisfied; 
while with the additional imposition of monotonicity on $I_1$ and $I_2$, the first derivatives of the network w.r.t. inputs $I_1$ and $I_2$ are always positive, leading to automatic fulfillment of Eq.\eqref{eq:polyconvex_I1}, and Eq.\eqref{eq:polyconvex_I2} and therefore of polyconvexity. 
However, 
monotonicity is a sufficient but not necessary condition for polyconvexity in ICNN, as the inequalities in Eqs.\eqref{eq:polyconvex_I1} and \eqref{eq:polyconvex_I2} could potentially hold without positive first derivatives if the term involving the strictly positive second derivative dominates. 
%

To address the gap between polyconvexity and over-restriction, in this work, we propose to relax the restrictive monotonicity constraints for the invariants $I_1$ and $I_2$ to improve expressiveness while utilize the polyconvexity indicator that was introduced in 
Eq.\eqref{eq:polyconvex_I1}-\eqref{eq:polyconvex_J}. 
In particular, while the internal connections of the network are constrained to have nonnegative weights to preserve the convexity of softplus compositions, we allow signed weights for $I_1$ and $I_2$, as well as their corresponding skip-connectors. 
This means that the learned strain-energy $\hat{\varphi}_{\mathrm{NN}}(I_1,I_2,J)$ is not guaranteed to be polyconvex by construction,
while it has significantly greater flexibility to fit the data. 
The rationale is that as monotonicity being a sufficient but not necessary condition for polyconvexity, 
its relaxation does not inherently imply full violation of polyconvexity a priori.  
The strict convex requirement of the composition of softplus layers may allow Eq.~\eqref{eq:polyconvex_I1} and Eq.~\eqref{eq:polyconvex_I2} to remain non-negative, meeting the key convexity conditions empirically, and yielding a trained function that is polyconvex in practice. Later in the manuscript the use of the polyconvexity indicator as a soft constraint during training will also be discussed.



\subsection{Consistency and Normalization}

%

\begin{figure}[h]
    \centering
    \includegraphics[width=0.9\linewidth]{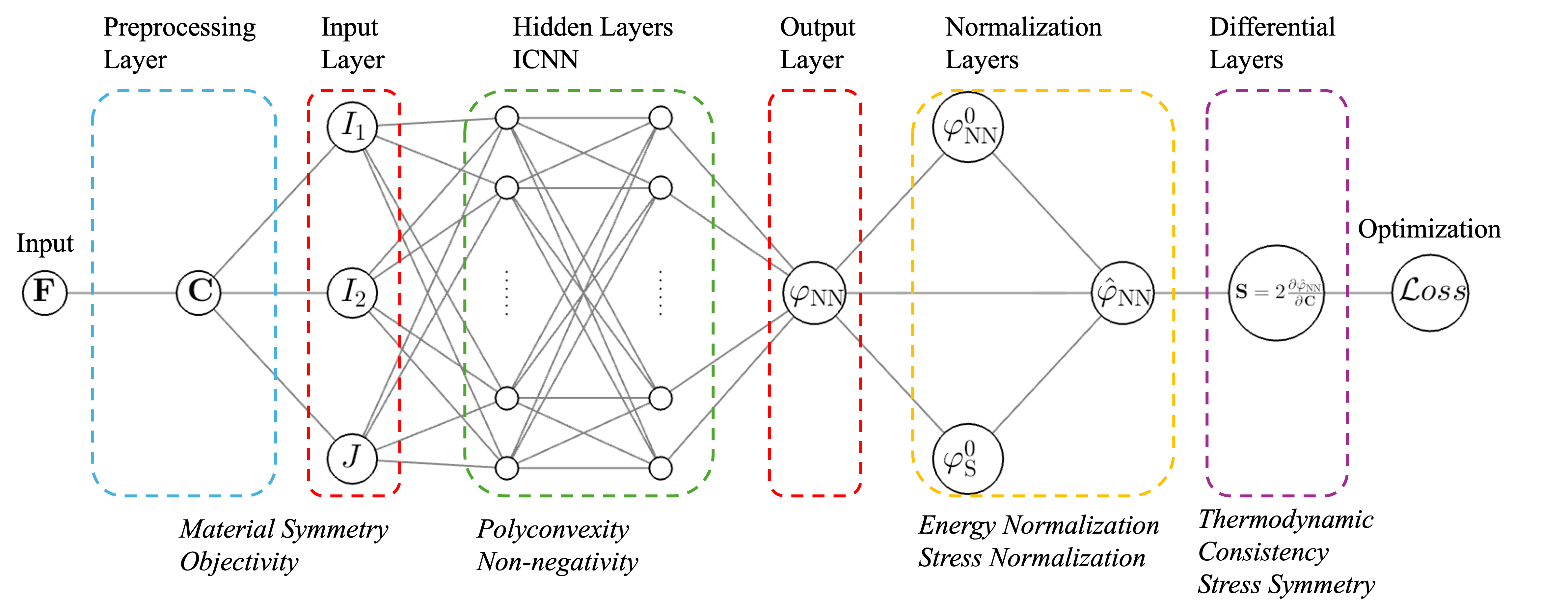}
    \caption{Illustration of the proposed physics-augmented neural network framework integrating an input convex neural network (ICNN) architecture with physics-based augmentation layers.}
    \label{fig:pann}
\end{figure}

The proposed 
physics-augmented training
framework incorporates three physics-augmentation layers  into the neural network training workflow in Fig.\ref{fig:pann}. These layers include preprocessing layer, normalization layer and differential layer.

%
A preprocessing layer transforming the input deformation tensor $\mathbf{F}$ into the right Cauchy-Green tensor $\mathbf{C}$, then into its three scalar invariants $(I_1, I_2, J)$ before passing them as the input layer to neural networks is mechanistically necessary to ensure the fulfillment of objectivity and material symmetry.
We then augment the training process of candidate neural networks with the differential relationship of Eq.\eqref{eq:stress_from_phi} as an intermediate processing layer to obtain the stress output for the calculation of the loss function against the train data. 
This design imposes a hard physics constraint: any stress predictions will derive from a potential, ensuring conservative (path-independent) responses by construction. 
Additionally, 
to ensure physical consistency, further augmentations are necessary. One important step is enforcing that the stress is zero in the reference configuration and that the energy is zero at zero strain.
We achieve so by computing the normalization terms for the energy function $\varphi_{\mathrm{NN}}^0$ and for the computed stress $\varphi_{\mathrm{S}}^0$,
\begin{equation}\label{eq:NNnormalized}
    \begin{aligned}
        \hat{\varphi}_{\mathrm{NN}} (I_1, I_2, J)
    &=
    \varphi_{\mathrm{NN}}(I_1, I_2, J)
    -\varphi_{\mathrm{NN}}^0
    -\varphi_{\mathrm{S}}^0
    \\
    &=
    \varphi_{\mathrm{NN}}(I_1, I_2, J)
    -
    \varphi_{\mathrm{NN}}(3,3,1)
    -
    n(J-1),
    \end{aligned}
\end{equation}
where 
\begin{equation}
    n=  \left. \left( 
    2 
    \frac{\partial \varphi_{\mathrm{NN}}}{\partial I_{1}}  
    + 4 
    \frac{\partial \varphi_{\mathrm{NN}}}{\partial I_{2}} 
    + 
    \frac{\partial \varphi_{\mathrm{NN}}}{\partial J}  
    \right) \right\rvert_{(I_1, I_2, J) = (3,3,1)}
\end{equation}
is the stress normalization constant, for more detail on its analytical derivation, see \cite{fuhg2024extreme}.

In summary, physics augmented training interweaves data-driven learning with enforcement of physics: invariants as inputs (objectivity), an energy-network architecture (hyperelasticity and convexity), normalization adjustments (stress-free reference), and analytic differentiation (to obtain consistent stresses). These steps ensure the learned model not only fits the data but also adheres to fundamental principles of mechanics.

\subsection{Smoothed $L_0$ Sparsification}
A recurring theme in data-driven material modeling is the trade-off between model complexity and interpretability. 
Traditional phenomenological models are simple and interpretable, typically involving only a small number of physically meaningful parameters. In contrast, neural networks offer high expressive capacity and can approximate complex responses directly from data. However, their typically over-parameterized structure and weakly constrained hypothesis space may result in poor out-of-distribution generalization and unreliable predictions when extrapolating beyond the training regime.
Evaluation of complex  NN-based constitutive laws can be highliy demanding computationally, and has motivated the utilization of vectorized architectures for finite elemnet implementation in GPUs\cite{alheit2026commet}. 
\par
Regularization techniques are commonly employed to control model complexity. $L_2$ regularization penalizes large weights and promotes smooth parameter distributions but does not induce sparsity.  $L_1$ regularization promotes sparse solutions by penalizing parameter magnitudes, encouraging parameters with limited contribution to shrink toward zero. More generally, $L_\alpha$ penalties ($0 < \alpha < 2$] interpolate between smooth shrinkage and sparsity-promoting behavior.  Therefore, a common choice to measure neural network complexity is the $L_{0}$-norm, which simply counts the number of non-zero weights. 
Minimizing $L_0$ directly would yield the sparsest network that fits the data, but this leads to a non-differentiable combinatorial optimization. 

Sparse regularization~\cite{louizos2017learning, mcculloch2024sparse, fuhg2024extreme} addresses this trade-off by promoting low-dimensional neural network parameterizations in which a subset of weights is driven exactly to zero while preserving predictive accuracy. Such sparsity enforcements has the potential to  mitigate overfitting in limited-data regimes and enhances interpretability, as the resulting network corresponds to a reduced functional representation. In practice, inducing sparsity in neural constitutive models can help the network ``\textit{discover}''  material laws or identifiable terms (e.g. a particular invariant combination), especially when data is limited.

Louizos et al.\cite{louizos2017learning} introduced a strategy to differentiate through the $L_{0}$ norm by using stochastic gates for the NN weights. This method re-parameterizes each trainable parameter as follows:

\begin{equation}
    \bm{\theta} = \bar{\bm{\theta}} \odot \mathsf{z},
\end{equation}

with $\mathsf{z} = \min (\bm{1}, \max (\bm{0}, \overline{\bm{s}}))$
where $\odot$ denotes the Hadamard product and
$ \overline{\bm{s}} = \bm{s} (\zeta- \gamma) + \gamma \bm{1},$
\begin{equation}
\bm{s} = \operatorname{sig}\Big(\frac{\log \bm{u} - \log (1-\bm{u})+ \log \bm{\alpha}}{\beta}\Big)
\end{equation}

The relaxation of the binary gate to a hard-concrete distribution with a continuous variable $\alpha$ provides a differentiable Monte-Carlo approximated complexity loss:

\begin{equation}
    \mathcal{R}_{\ell_0}(\bm{\theta}) 
    =
    \sum_{j=1}^{|\bm{\theta|}} \text{Sigmoid}(\log \alpha_{j} - \beta \log \frac{-\gamma}{\zeta}), \label{eq:l0}
\end{equation}
with hyperparameters $(\gamma,\zeta,\beta)$ of the stochastic gates. Intuitively, each term in the summation is the probability that trainable parameter $\theta_j$ is non-zero, so the summation estimates the expected number of non-active parameters.
By adding Eq.\eqref{eq:l0} to the training loss, the training will encourage many gates $z_j$ to go to $0$, effectively pruning the network. 
%

Enforcing sparsity in a PANN not only reduces overfitting but also yields a model that is much easier to interpret and generalize (extrapolate), but also simple to deploy in existing finite element workflows, and not computationally demanding. In the realm of material modeling, an interpretable model is often one where the functional form can be examined \cite{fuhg2024extreme}.

\section{Adjoint-Based Finite Element Updates}\label{sec:adj}

Experiments and computational models are associated by observables:
\begin{equation}
    \mathbf{D} = \boldsymbol{\mathcal{F}}(\boldsymbol{\theta}) + \boldsymbol{\xi}
\end{equation}
where $\mathbf{D}$ is the experimental observable (a.k.a. \textit{data}), the computational action $\boldsymbol{\mathcal{F}}$ on model parameters $\boldsymbol{\theta}$ produces computational observable (a.k.a. \textit{state}), and $\boldsymbol{\xi}$ represents the discrepancy between the observables.
The minimization of $|\xi|$ by identifying the optimal values of $\boldsymbol{\theta}$ is a case of an inverse problem, namely parameter identification. Furthermore, one can also perform the minimization of $|\xi|$ by finding the optimal computational action $\boldsymbol{\mathcal{F}}$, this is a case of an optimal model selection problem, for more information, we direct the readers to \cite{tan2022toward}. In this work, we focus on the aforementioned parameter identification problem.

In the case of $\boldsymbol{\mathcal{F}}$ taking the form of a system of time-dependent nonlinear partial differential equations (strong form) derived from first principles and constitutive laws, we can denote the model in an abstract form,
\begin{equation}
    \boldsymbol{\mathcal{R}}(\mathbf{u}, \boldsymbol{\theta}) = \mathbf{0} \quad \mathrm{in\ } \mathcal{V}^\prime.
\end{equation}
This notation is interpreted as given model parameter $\boldsymbol{\theta}\in\Theta$, find the state variable $\mathbf{u}\in\mathcal{V}$ such that the strong residual $\boldsymbol{\mathcal{R}}$ vanishes. We can translate the parameter identification problem into a PDE-constrained optimization problem,
\begin{equation}
    \begin{aligned}
        \argmin_{\boldsymbol{\theta} \in \Theta} & \quad
        \mathcal{J}(\mathbf{u},\boldsymbol{\theta})
    \\
    \mathrm{s.t.} &\quad   
    \boldsymbol{\mathcal{R}}(\mathbf{u},\boldsymbol{\theta}) = \mathbf{0} \quad \mathrm{in\ } \mathcal{V}^\prime
    \end{aligned}
\end{equation}
Find the optimal values of $\boldsymbol{\theta}\in\Theta$ to minimize the value of the scalar objective functional $\mathcal{J}$, while the $\mathbf{u}$ is a physically admissible solution of $\boldsymbol{\mathcal{R}}$ at the given parameter values. While the specific choice of $\mathcal{J}$ varies, it generally takes the form of an error measure between $\mathbf{D}$ and $\mathbf{u}$, known as data-model misfit.

The PDE constraint can be combined with the optimization objective by the use of an arbitrary Lagrange multiplier $\mathbf{v}\in\mathcal{V}_0$,
\begin{equation}
    \mathcal{L}(\mathbf{u}, \mathbf{v}, \boldsymbol{\theta}) 
    =  
    \mathcal{J}(\mathbf{u},\boldsymbol{\theta})
    +
    \big\langle
    \mathbf{v}, \boldsymbol{\mathcal{R}}(\mathbf{u},\boldsymbol{\theta}) 
    \big\rangle.
\end{equation}
By expanding the $L^2$-inner product operation $\big\langle\cdot,\cdot\big\rangle$, it is equivalent to obtaining the weak form of $\boldsymbol{\mathcal{R}}$ via weighted residuals, and $\mathbf{v}$ is the test function, or the adjoint variable. The Lagrangian in weak form,
\begin{equation}
    \mathcal{L}(\mathbf{u}, \mathbf{v}, \boldsymbol{\theta}) 
    =  
    \mathcal{J}(\mathbf{u},\boldsymbol{\theta})
    +
    r(\mathbf{u}, \mathbf{v}, \boldsymbol{\theta}), 
\end{equation}
where $r()$ is the weak form scalar residual, and the minimization of the Lagrangian,
\begin{equation}
    \argmin_{\boldsymbol{\theta} \in \Theta}  \quad
        \mathcal{L}(\mathbf{u},\mathbf{v},\boldsymbol{\theta}),
\end{equation}
is equivalent to the original PDE-constrained optimization problem.
The optimality condition of $\mathcal{L}$ requires that the first variations of the Lagrangian functional $\mathcal{L}$ with respect to all variables must equal to zero. Taking the first variation of $\mathcal{L}$ with respect to the adjoint variable $\mathbf{v}$ in an arbitrary direction $\delta\mathbf{v}$ and setting it to zero,
\begin{equation}
    \big\langle  
    \partial_\mathbf{v} \mathcal{L}, \delta\mathbf{v} 
    \big\rangle
    =
    r(\mathbf{u}, \delta\mathbf{v}, \boldsymbol{\theta})
    =0
    ,
    \quad \forall \delta\mathbf{v} \in \mathcal{V}_0,
    \label{eq:state}
\end{equation}
this is equivalent to solving the governing PDEs $\boldsymbol{\mathcal{R}}(\mathbf{u}, \boldsymbol{\theta})=\mathbf{0}$ for the state variable at the given $\boldsymbol{\theta}$, this is the state (forward) problem. Similarly, taking the first variation of $\mathcal{L}$ with respect to the state variable $\mathbf{u}$ in an arbitrary direction $\delta\mathbf{u}$ and setting it to zero,
\begin{equation}
    \big\langle  
    \partial_\mathbf{u} \mathcal{L}, \delta\mathbf{u} 
    \big\rangle
    =
    \big\langle  
    \partial_\mathbf{u} \mathcal{J}, \delta\mathbf{u} 
    \big\rangle
    +
    r(\delta\mathbf{u}, \mathbf{v}, \boldsymbol{\theta})
    =0
    ,
    \quad \forall \delta\mathbf{u} \in \mathcal{V},
    \label{eq:adjoint}
\end{equation}
this is equivalent to solving the linear adjoint operator $\boldsymbol{\mathcal{R}}_0(\mathbf{v},\mathbf{u}, \boldsymbol{\theta})=-\frac{\partial\mathcal{J}}{\partial\mathbf{u}}$ at the given $\boldsymbol{\theta}$ and the corresponding $\mathbf{u}$ from \eqref{eq:state}, this is the adjoint (backward) problem. With the state solution $\mathbf{u}$ from \eqref{eq:state} and the adjoint solution $\mathbf{v}$ from \eqref{eq:adjoint}, we can take the variation of $\mathcal{L}$ with respect to the model parameters $\boldsymbol{\theta}$ in an arbitrary direction $\delta\boldsymbol{\theta}$ to obtain the weak form of the gradient:
\begin{equation}
        \big\langle  
    \partial_{\boldsymbol{\theta}} \mathcal{L}, \delta\boldsymbol{\theta}
    \big\rangle
    =
    \big\langle  
    \partial_{\boldsymbol{\theta}} \mathcal{J}, \delta\boldsymbol{\theta}
    \big\rangle
    +
    r(\mathbf{u}, \mathbf{v}, \delta\boldsymbol{\theta})
    \quad \forall \delta\boldsymbol{\theta} \in \Theta,
    \label{eq:grad}
\end{equation}
and in strong form, 
\begin{equation}
    \mathrm{grad}_{\boldsymbol{\theta}} \mathcal{L} \bigg|_{\boldsymbol{\theta}}
    =
    \frac{\partial \mathcal{J}}{\partial \boldsymbol{\theta}} \bigg|_{(\mathbf{u}, \boldsymbol{\theta})}
    +
    \frac{\partial r}{\partial \boldsymbol{\theta}} \bigg|_{(\mathbf{u}, \mathbf{v}, \boldsymbol{\theta})}
\end{equation}











\subsection{Objective function for full-field DIC-type calibration}\label{subsec:adj}
The objective function for a DIC-type full field dataset, can be cast as \begin{equation}
    \begin{aligned}
        \argmin_{\boldsymbol{\theta} \in \boldsymbol{\mathbb{R}^+}} & \quad
        J(\mathbf{u},\boldsymbol{\theta})
        =
        \frac{1}{2}\int_\Omega |\mathbf{u}-\mathbf{u}^\mathrm{data}|_2 \,\mathrm{d}\Omega
        +
        \frac{\alpha_1}{2}\bigg(  F(\mathbf{u},\boldsymbol{\theta})-F^\mathrm{data}  \bigg)^2
        + 
        \alpha_2\,\mathcal{R}(\boldsymbol{\theta})
    \\
    \mathrm{s.t.} &\quad   
    r(\mathbf{u},\boldsymbol{\theta}) = \nabla\cdot \mathbf{P}+\mathbf{B} = \mathbf{0} \quad \mathrm{in\ }\Omega 
    \\
    &\quad   \mathbf{u}=\mathbf{u}^\dagger  \quad \mathrm{on\ }\partial\Omega_D 
    \\
    &\quad   \mathbf{P\cdot N} = \mathbf{T}  \quad \mathrm{on\ }\partial\Omega_N 
    \\
    &\quad   \mathbf{u} \in \mathcal{V}
    \end{aligned}
    \label{eq:optimization}
\end{equation}
where $\alpha_1,\,\alpha_2$ are weight for the multi-objective optimization problem, $\mathbf{P}(\mathbf{u},\boldsymbol{\theta})$ is the first Piola–Kirchhoff stress tensor given a strain energy density function $\varphi(I_1,I_2,J, \boldsymbol{\theta})$ as in \eqref{eq:stress_from_phi} and normalized (zero stress in reference configuration) through \eqref{eq:NNnormalized} applied in the last term of the objective function. Consequently, the scalar force is computed as follows, given the unit direction of loading $\hat{n}$:
\begin{equation}
    F(\mathbf{u},\boldsymbol{\theta})
    =
    \int_{\partial\Omega}
    (\boldsymbol{P\cdot N})\cdot \hat{n} \,\mathrm{d} \partial\Omega
\end{equation}
where $\boldsymbol{N}$ is the surface normal at the deformed configuration.  
In this setting the first set of the objective function integrates the discrepancy from the full-field observable, in this case the displacement field $\mathbf{u}$,  
while the second term evaluates the discrepancy of the scalar reaction force in the standard mean-square error form. 
Lastly, the third term in the objective represents regularization function(s) of choice and/or inequality penalties such as \eqref{eq:polyconvex_I1} and \eqref{eq:polyconvex_I2}.

To solve the PDE-Constrained optimization problem, we define the Lagrangian
\begin{equation}
    \mathcal{L}(\mathbf{u}, \mathbf{v}, \boldsymbol{\theta}) 
    = 
    J(\mathbf{u},\boldsymbol{\theta})
    +
    \langle \mathbf{v}, r(\mathbf{u},\boldsymbol{\theta}) \rangle
\end{equation}
where $\mathbf{v} \in \mathcal{V}_0$ is the adjoint variable, and $\langle\cdot,\cdot\rangle$ denotes the $L^2$-inner product
\begin{equation}
    \langle  \mathbf{u}(\mathbf{x}) , \mathbf{v}(\mathbf{x})  \rangle = \int_\Omega \mathbf{u}\cdot\mathbf{v} \,\mathrm{d}\Omega
\end{equation}
By expanding the $L^2$-inner product, it is essentially the weak form of the PDE residual. The full Lagrangian functional is as follow:
\begin{equation}
    \begin{aligned}
        \mathcal{L}(\mathbf{u}, \mathbf{v}, \boldsymbol{\theta}) 
    &= 
    \frac{1}{2}\int_\Omega |\mathbf{u}-\mathbf{u}^\mathrm{data}|_2 \,\mathrm{d}\Omega
        +
        \frac{\alpha_1}{2}\bigg(  \int_{\partial\Omega}
    (\boldsymbol{P\cdot N})\cdot \hat{n} \,\mathrm{d} \partial\Omega-F^\mathrm{data}  \bigg)^2
        + 
        \alpha_2\,\mathcal{R}(\boldsymbol{\theta})
        \\
    &+
    \int_\Omega \mathbf{P}:\nabla \mathbf{v} \,\mathrm{d}\Omega 
    - 
    \int_\Omega \mathbf{B} \cdot \mathbf{v} \,\mathrm{d}\Omega
    -
    \int_{\partial\Omega} \mathbf{T} \cdot \mathbf{v} \,\mathrm{d}\partial\Omega
    \end{aligned}
    \label{eq:lagrange}
\end{equation}

The variations of the Lagrangian functional \eqref{eq:lagrange} with respect to all variables must vanish in order to solve the optimization problem in \eqref{eq:optimization}.

Vanishing the variation of the Lagrangian functional \eqref{eq:lagrange} with respect to the adjoint variable $\mathbf{v}$ in an arbitrary direction $\delta\mathbf{v}$ results in the state problem:
\begin{equation}
    \langle  \partial_\mathbf{v} \mathcal{L}, \delta\mathbf{v} \rangle
    =
    \int_\Omega \mathbf{P}:\nabla (\delta\mathbf{v}) \,\mathrm{d}\Omega 
    - 
    \int_\Omega \mathbf{B} \cdot (\delta\mathbf{v}) \,\mathrm{d}\Omega
    -
    \int_{\partial\Omega} \mathbf{T} \cdot (\delta\mathbf{v}) \,\mathrm{d}\partial\Omega
    =0,
    \quad \forall \delta\mathbf{v} \in \mathcal{V}_0
    \label{eq:state}
\end{equation}

Vanishing the variation of the Lagrangian functional \eqref{eq:lagrange} with respect to the state variable $\mathbf{u}$ in an arbitrary direction $\delta\mathbf{u}$ results in the adjoint problem:
\begin{equation}
    \begin{aligned}
        \langle  \partial_\mathbf{u} \mathcal{L}, \delta\mathbf{u} \rangle
    &=
    \int_\Omega (\mathbf{u}-\mathbf{u}^\mathrm{data}) \cdot \delta\mathbf{u} \,\mathrm{d}\Omega
        \\&
        +
        \alpha_1\bigg(  \int_{\partial\Omega}
    (\boldsymbol{P\cdot N})\cdot \hat{n} \,\mathrm{d} \partial\Omega-F^\mathrm{data}  \bigg)
    \bigg(  \int_{\partial\Omega}
    ((\frac{\partial\mathbf{P}}{\partial \mathbf{u}}:\delta\mathbf{u})\cdot\boldsymbol{N})\cdot \hat{n} \,\mathrm{d} \partial\Omega  \bigg)
    \\&
    +
    \int_\Omega \mathbf{P}(\delta\mathbf{u}):\nabla\mathbf{v} \,\mathrm{d}\Omega 
    - 
    \int_{\partial\Omega} \mathbf{T}(\delta\mathbf{u}) \cdot \mathbf{v} \,\mathrm{d}\partial\Omega
    =0,
    \quad \forall \delta\mathbf{u} \in \mathcal{V}_0
    \end{aligned}
    \label{eq:adjoint}
\end{equation}

With the state solution $\mathbf{u}$ from \eqref{eq:state} and the adjoint solution $\mathbf{v}$ from \eqref{eq:adjoint}, we can take the variation of the Lagrangian functional \eqref{eq:lagrange} with respect to the model parameters $\boldsymbol{\theta}$ in an arbitrary direction $\delta\boldsymbol{\theta}$ to obtain the weak form of the gradient:
\begin{equation}
    \begin{aligned}
        \langle  \partial_{\boldsymbol{\theta}} \mathcal{L}, \delta\boldsymbol{\theta} \rangle
    &=
    \alpha_1\bigg(  \int_{\partial\Omega}
    (\boldsymbol{P\cdot N})\cdot \hat{n} \,\mathrm{d} \partial\Omega-F^\mathrm{data}  \bigg)
    \bigg(  \int_{\partial\Omega}
    ((\frac{\partial\mathbf{P}}{\partial \boldsymbol{\theta} }:\delta\boldsymbol{\theta} )\cdot\boldsymbol{N})\cdot \hat{n} \,\mathrm{d} \partial\Omega  \bigg)
    \\&
    +
    \int_\Omega \mathbf{P}(\delta\boldsymbol{\theta}, \mathbf{u}):\nabla\mathbf{v} \,\mathrm{d}\Omega 
    - 
    \int_{\partial\Omega} \mathbf{T}(\delta\boldsymbol{\theta}, \mathbf{u}) \cdot \mathbf{v} \,\mathrm{d}\partial\Omega
    =0,
    \quad \forall \delta\boldsymbol{\theta} \in \mathbb{R}^+
    \end{aligned}
\end{equation}

\section{A Transfer Learning Scheme} 

Despite recent developments in data-driven modeling,    deployment, training and testing procedures for neural networks are often computationally exhausting, especially due to the high-dimensionality of neural networks. This is especially valid if the neural networks have to be evaluated repeatedly for a single evaluation of a complicated loss (e.g. a loss necessitating the numerical solution of a PDE).
The training of neural networks often relies on stochastic gradient descent (SGD) and its variants. 
SGD is computationally efficient, especially when dealing with high-dimensional spaces, because it performs updates more frequently and processes only partial updates (partial gradients) per iteration, leading to faster steps toward a global solution exploring a highly nonconvex space. 
Additionally, backpropagation through automatic differentiation provides direct access to analytical gradients, further enhancing the computational efficiency of the optimization process.
However, the updates are noisy and can fluctuate significantly, causing the algorithm to oscillate, which may result in a high iteration count before finding an acceptably optimal solution.

In the context of discovery of constitutive laws with the suggested NN-based approaches, training a NN within the PDE-constrained optimization in Sec.\ref{sec:adj} can be rather challenging due to the repeated evaluation of the NN at integration points and also the adjoint evaluation and backpropagation requirements for this potentially high-dimensional representation. The adjoint method, comparing to a direct differentiable implementation of a fully auto-differentiable solver, provides direct access to full local gradients independent of the dimensional of model parameters nor of the high nonlinearity of the physics while precisely incorporating physics constraints. 
However, this approach requires solving the full and often nonlinear state problem at least once per iteration, as well as the associated linear adjoint problems, which could pose significant computational challenges.

To this end, we propose a 
transfer learning strategy which facilitates accelerated computations and targeted discovery. This strategy combines physics constraints for the constitutive law discovery along with the versatility of the finite element method for probing experiments that enable access to full field information.
Instead of combining neural networks and finite element solvers as one single end-to-end differentiable pipeline, the core idea is to utilize multi-modal data and decouple model discovery into two stages with distinct objectives and methods.

In stage 1, which is referred as the pre-training stage,  training data from simple mechanical test (e.g. uniaxial, biaxial, etc.) corresponding to homogeneous stress states are utilized to train and sparsify NN variants with different physics augmentation. This could be done  on the specific material that will be further tested with DIC imaging, or utilizing a different material from the same material class (expecting quantitative, but no significant qualitative differences). Extreme sparsification has been shown to push these physics-augmented NN models to very low-dimensional representations which in turn is beneficial in allowing connecting to robust FE solvers; in this case FEniCS is utilized in the next stage.
In stage 2, this pretrained model is inserted into a FE-based adjoint framework and further adapted to recapitulate more complex full field data. 
By separating the training scheme into two stages, the approach exploits the strengths of each. Stage 1: pre-training does not necessitate using an FE solver as the data for that stage is simple, and the model can be efficiently augmented with physical constraints and sparsified moving from an initially high-dimensional representation to a low-dimensional interpretable and compact expression. 
Stage 2: transfer learning utilizing the adjoint based PDE-constrained optimization scheme, enables fine-tuning the low-dimensional model while adapting the physical constraints for the sparsified PANN expression directly in the constraing of the adjoint framework. Additionally, PDE-constrained optimization excels when the search space is small and well-initialized.
Crucially, sparsification of the PANN during pre-training enables closing the loop with an FE-based adjoint framework, aleviating difficulties that can arise during optimization due to the  fungibility of a higher dimensional representation. In the following result section, we demonstrate the proposed transfer learning strategy with numerical experiments spanning through the entire transfer learning pipeline, starting from data preparation, to both aforemention stages, and lastly with a demonstration of fine-tuned model deployment.

\section{Results}
\subsection{Multi-modal data preparation}


This work employs a multi-modal data generation strategy designed to reflect the heterogeneous nature of information typically available in experimental solid mechanics. In practice, constitutive modeling rarely relies on a single type of observation. Instead, material behavior is inferred from a combination of conventional mechanical tests that apply single loading modes (e.g., uniaxial tension or shear) together with full-field observations acquired through techniques such as digital image correlation (DIC). These sources provide complementary insights, conventionally mechanical tests yield homogenized stress–strain responses under controlled loading paths, whereas full-field observations capture heterogeneous deformation patterns that indirectly encode constitutive behavior.

\par
Accordingly, two distinct but connected datasets are constructed. First, a synthetic dataset (containing labeled data pairs of invariant triplets and stresses) is generated to emulate simple mechanical tests and is used for physics-informed pre-training of the neural constitutive model. This stage enables the network to learn physically admissible stress responses within a broad region of invariant space. Second, a synthetic full-field displacement dataset is generated through virtual DIC experiments in a finite element setting. These data serve as observational inputs during transfer learning, where the pretrained model is adapted to an unknown target material using experimentally realistic measurements that do not directly provide stresses.
\par
The separation between these modalities reflects realistic experimental workflows in material characterization: well-understood materials or simplified tests are often available to establish constitutive priors, whereas complex prototype materials are typically characterized through limited full-field experiments. The following subsections describe the generation of the pre-training dataset and the full-field dataset used for transfer learning. This section focuses on: data-set preparation,  physics-augmented NNs pre-training under physics and sparsity constraints, transfer learning via PDE-constrained optimization, and finally deployment of the model in predictive simulations.


\subsubsection{Pre-training dataset}

\begin{figure}[H]
    \centering
    \includegraphics[width=0.32\linewidth, clip=true, trim = 10mm 0mm 15mm 0mm]{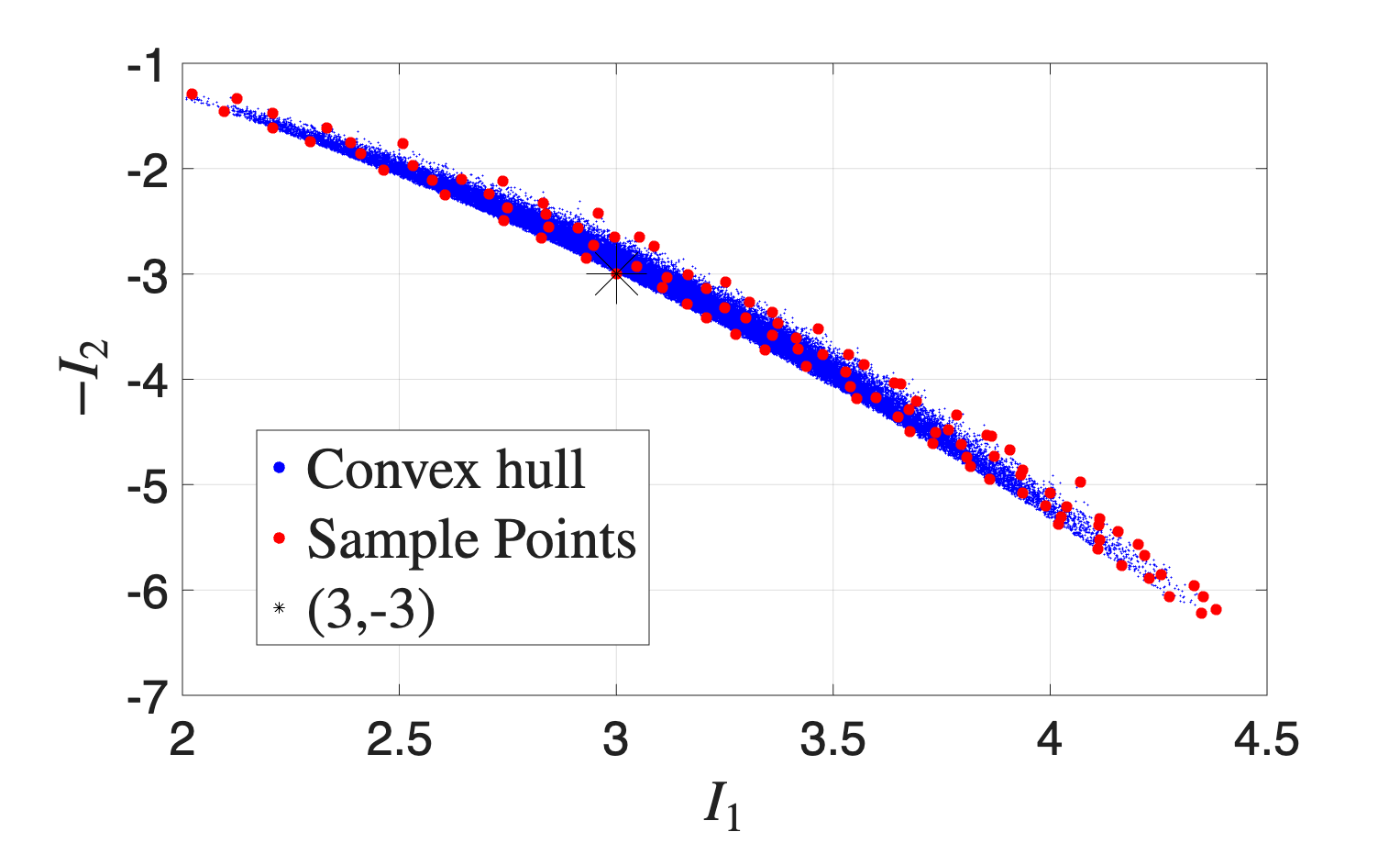}
    ~
    \includegraphics[width=0.32\linewidth, clip=true, trim = 0mm 0mm 15mm 0mm]{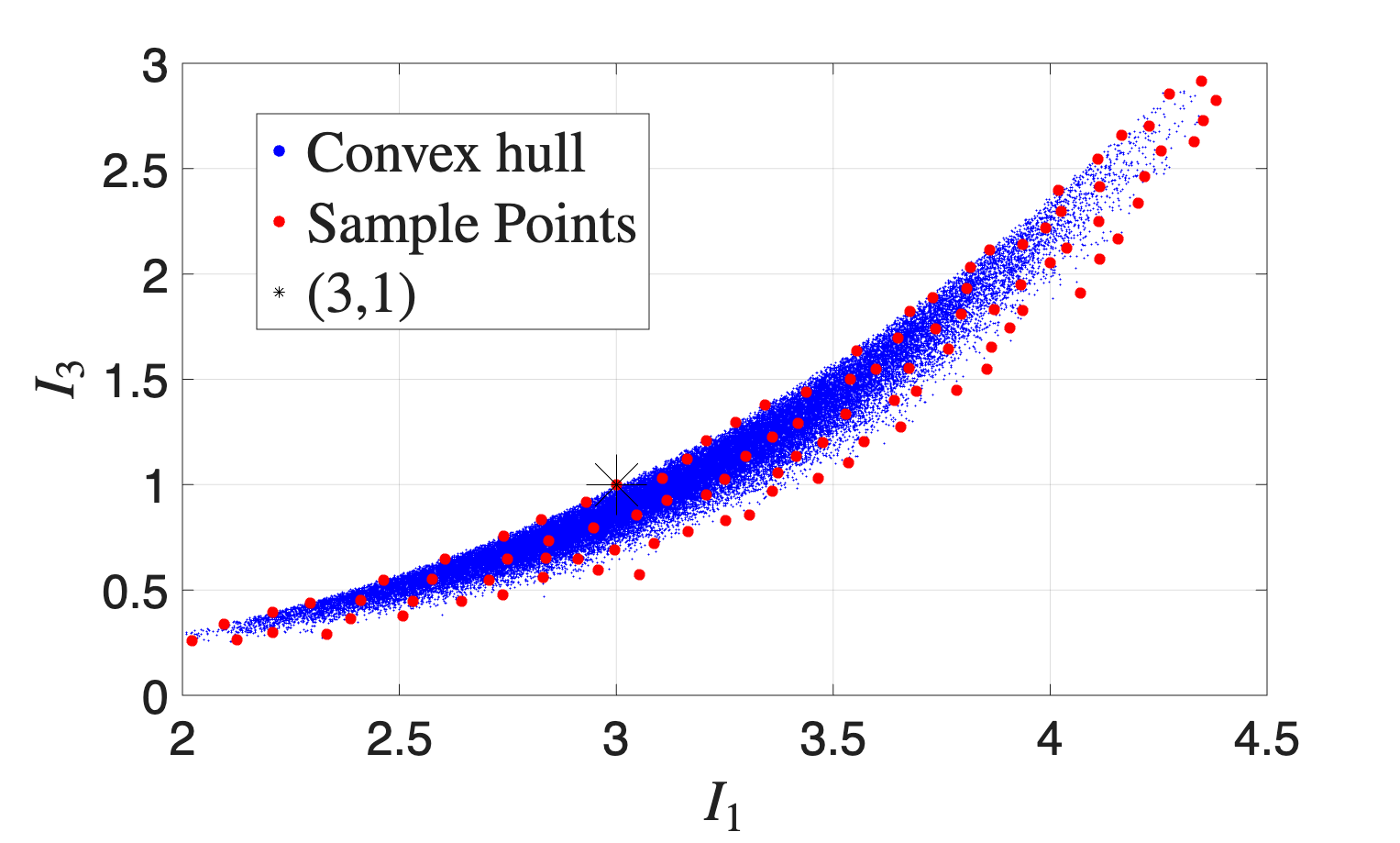}
    ~
    \includegraphics[width=0.32\linewidth, clip=true, trim = 0mm 0mm 15mm 0mm]{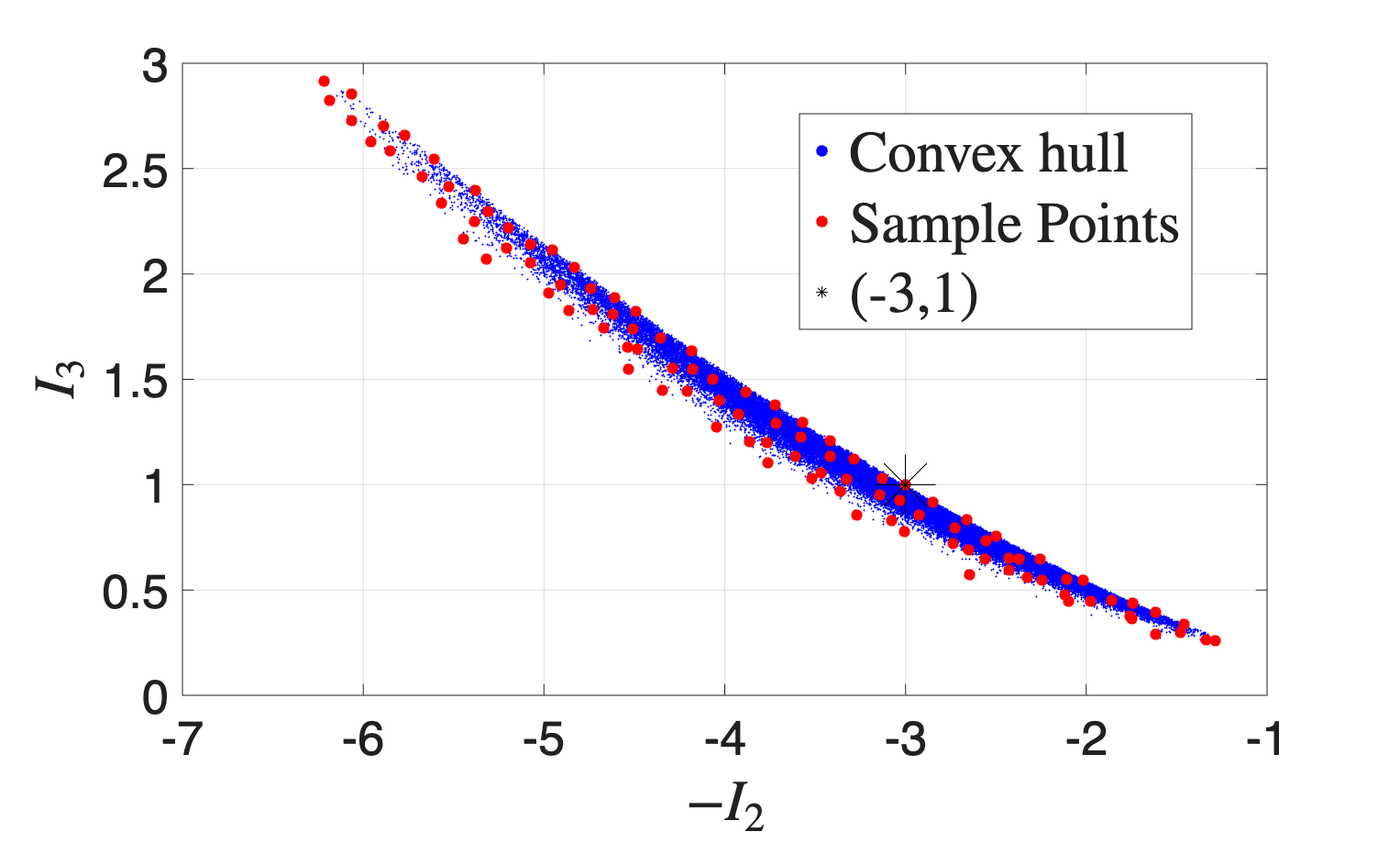}
    \caption{The invariant space showing the convex hull and the selected invariant triplets used for synthetic data generation.}
    \label{fig:invariant_samples}
\end{figure}

Following Fuhg et al.\cite{fuhg2022physics, fuhg2024stress}, we generate data on complex heterogeneous stress states, motivated on how one would obtain data from the perspective of computational homogenization, interogating an RVE in specific loading paths. In the context of compressible hyperelasticity, we first obtain a convex hull of physically permissible invariant space points through Latin Hypercube Sampling in the deformation gradient space, based on the invariants of 50000 samples of the deformation gradient tensor with a bound of $\delta=0.2$, and 
\begin{equation}
    F_{ij} \in \begin{cases}
        1+\mathbb{U}(-\delta,\delta), \ \mathrm{if\ }i=j
        \\
        \mathbb{U}(-\delta,\delta), \ \mathrm{else}
    \end{cases}.
\end{equation}
Next, in a heuristic fashion, we run a hybrid optimization scheme, farthest-point sampling (FRS) and simulated annealing (SA) (see Appendix \ref{Appdx:FRS+SA}) to \textit{optimally select} 100 invariant triplets [$I_1^i, I_2^i, I_3^i$] from the 50000-point convex hull such that they are well-spaced in the convex hull. 
Figure \ref{fig:invariant_samples} shows the 2D marginal distributions of the convex hull and the selected invariant triplets. The origin of the invariant space $(3,3,1)$, corresponding to the undeformed state, is manually enforced to the selection. 
Following Burnside \cite{burnside1892theory}, a diagonal right Cauchy-Green tensor consisting solely of principal strains 
\begin{equation}
    \mathbf{C}_{rec}^i = \mathrm{diag}\bigg(  
    (\lambda_1^i)^2, (\lambda_2^i)^2, (\lambda_3^i)^2
    \bigg),
\end{equation}
can be reconstructed for each invariant triplets (see Appendix \ref{appdx:rec_C}). 
%
It is worth to mention that, contrast to the simulated annealing sampling scheme proposed in Fuhg et al.\cite{fuhg2022physics}, the heuristic selection scheme utilized here preserves access to the (non-diagonal) true deformation gradient tensors $\mathbf{F}^i_{\mathrm{true}}$ corresponding to each selected invariant triplets for model verification purposes, though access to these tensors is not necessary for training.

\begin{figure}[H]
    \centering
    \includegraphics[width=0.95\linewidth]{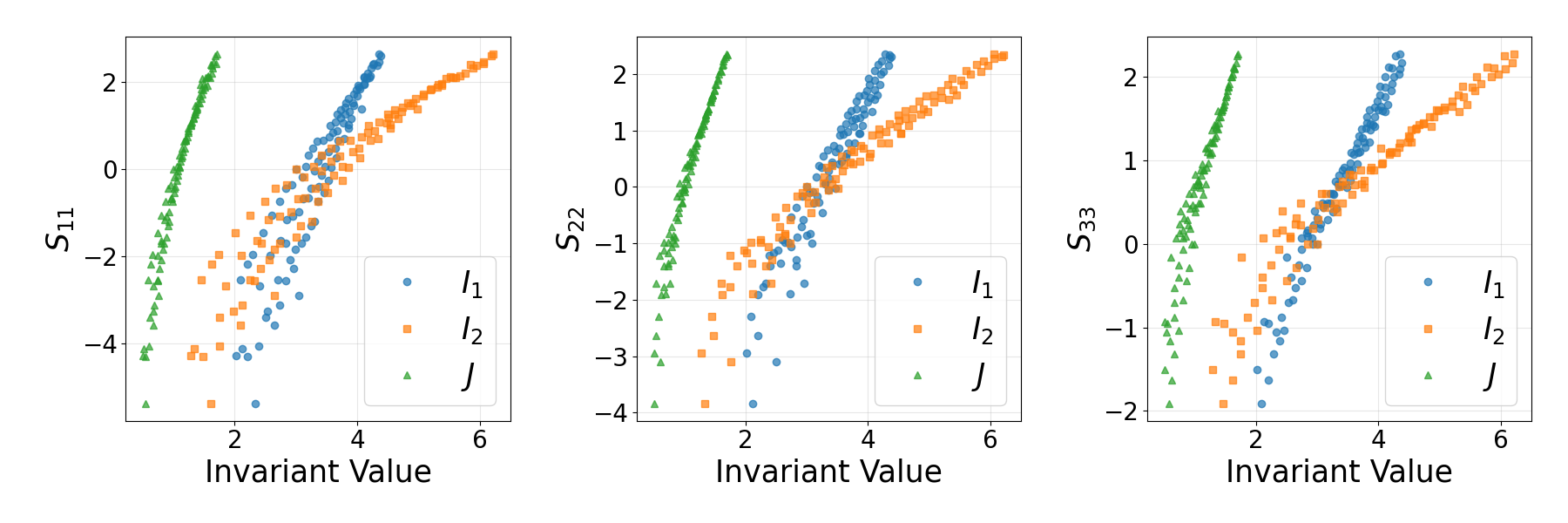}
    \caption{Sampled principal stress points as functions of invariants for synthetic data generated using the Gent model.}
    \label{fig:PANNdata}
\end{figure}

The first target for reconstruction is the Gent-Gent hyperelastic model
\begin{equation}
    \varphi_\mathrm{gent}(I_1,I_2,J)
    =
    -\frac{\mu}{2} \, J_m \, \ln\bigg( 1-\frac{I_1-3}{J_m} \bigg)
    -C_2 \, \ln\bigg( \frac{I_2}{3} \bigg)
    + \kappa \bigg( \frac{1}{2} (J^2-1) - \ln J  \bigg)
    \label{eq:gent}
\end{equation}
with material parameters $\mu=2.4195$, $J_m = 77.931$, $\kappa=1.20975$ and $C_2 = 0.75\mu$,
%
It is utilized as a synthetic stand-in for the true material or microstructure, in order to generate \textit{training data} and \textit{testing data}. 
%
%
%
%
Partial differentiation of \eqref{eq:gent} w.r.t. its input invariants, combining with 
the diagonal right Cauchy-Green tensor $\mathbf{C}_{rec}^i$ implicitly reconstructed from the corresponding invariant triplets,
generates the corresponding diagonal stress tensors $\hat{\mathbf{S}}^i$ with non-zero normal components.
It is worth to note that thanks to the intentional selection scheme for the invariant triplets which preserve validation access to the (non-diagonal) true deformation gradient tensors $\mathbf{F}^i_{\mathrm{true}}$, we can compute the true stress respond $\mathbf{S}^i_{\mathrm{true}}$, which is non-diagonal, for model validation purposes solely.
This concludes the generation of \textit{pre-training data}. 
In Fig.\ref{fig:PANNdata}, each component of the  diagonal synthetic stress tensors is plotted against the corresponding invariant triplets. 
These normal stress responses and the corresponding invariant triplets are the only information needed at the pre-training stage. 
It is noted, that this could also be performed when only simple tests are available as well (e.g. uniaxial, biaxial, simple shear, hydrostatic loading), as previously showcased for the Treloar dataset in Fuhg et al. 
\cite{fuhg2024extreme}.

\begin{figure}[H]
    \centering
    \includegraphics[width=0.3\linewidth]{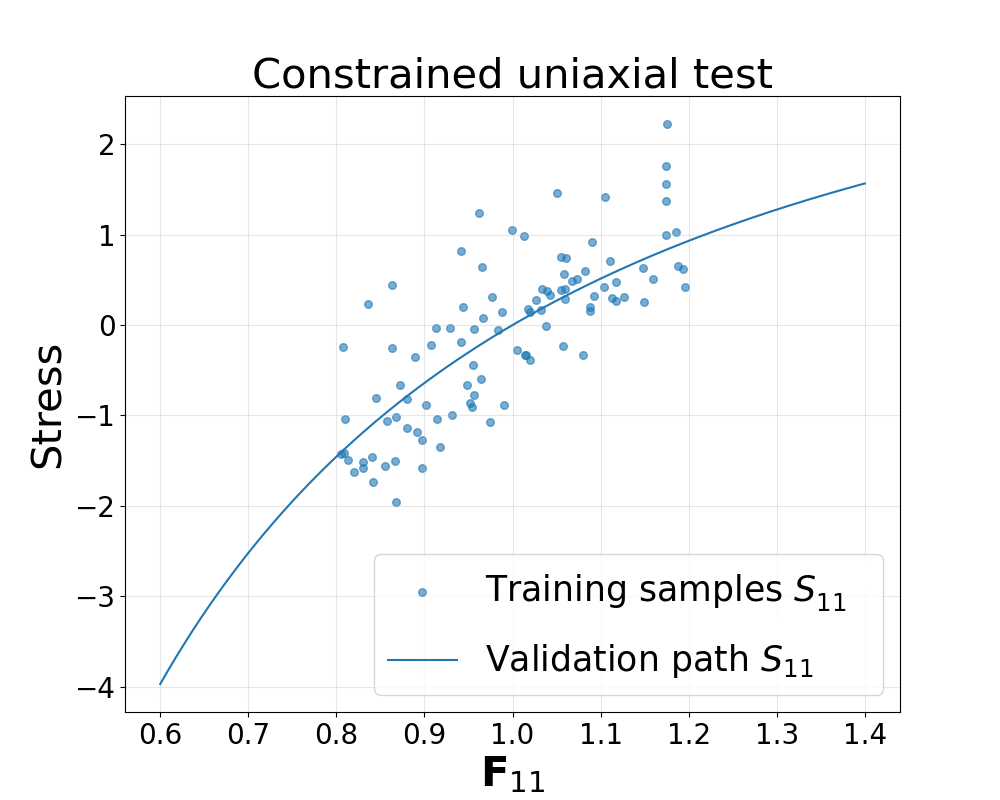}
    ~
    \includegraphics[width=0.3\linewidth]{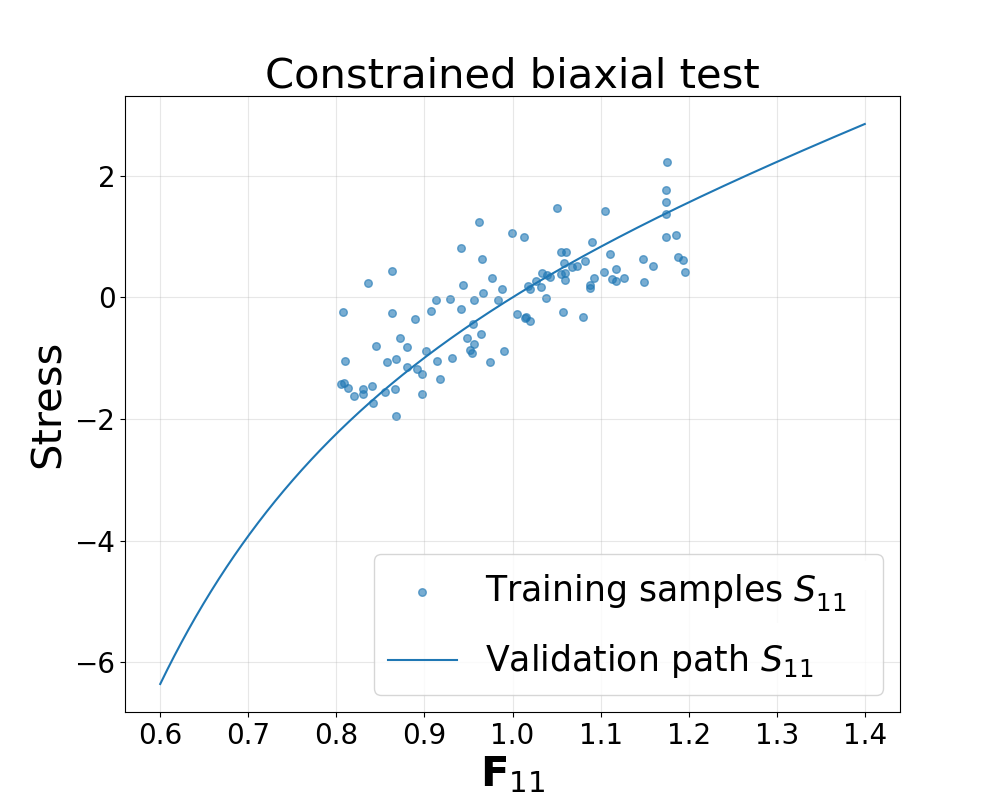}
    ~
    \includegraphics[width=0.3\linewidth]{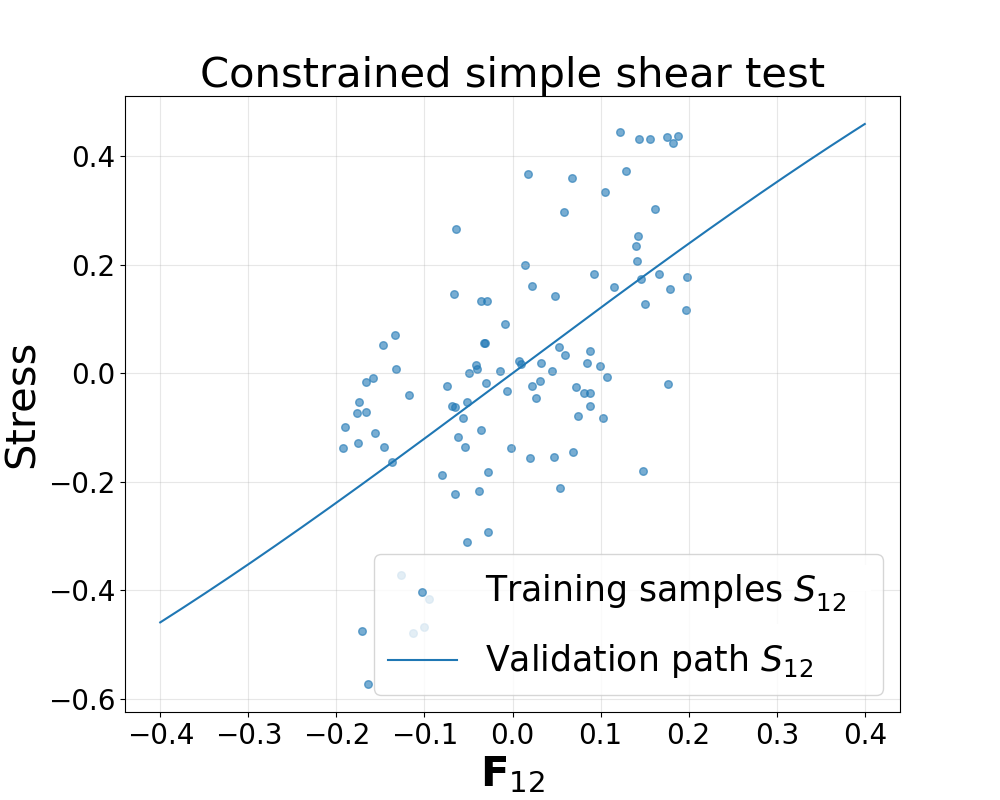}
    \caption{Sampled training data and validation paths in stress–deformation space for constrained uniaxial, biaxial, and simple shear tests.}
    \label{fig:gent_synthetic_data}
\end{figure}

To test the validity of the pretrained model beyond fitting training data, a set of \textit{testing data} is generated with \eqref{eq:gent} with three canonical loading paths: 
: (1) constrained uniaxial tension (with lateral contractions suppressed, $F_{11}$ varies over a large range 0.6–1.4), (2) constrained biaxial tension (simultaneous stretch in two directions while suppressing stretch on the third), and (3) simple shear (with shear component $F_{12}$ varying from –0.4 to 0.4). 
Along each loading path, we sampled stress responses across an extended strain range that exceeds the bounds of the training data.
Figure \ref{fig:gent_synthetic_data} shows the \textit{testing data} for each canonical loading paths (lines) compared to data obtained utilizing our sampling scheme over more complex stress/deformation states (scattered points). Figure \ref{fig:gent_synthetic_data} just showcases the range of data for each stress component for our sampling scheme even though the sampling  range is smaller that that utilized for the canonical examles.
%
It also clearly reveals how limited the training data is relative to the canonical loading paths. For a sufficient representation, the network will need to interpolate and extrapolate between sparse points in the invariant space and the principle stress space. 
The testing data-set will be used after pre-training, as a form of validation to evaluate how well the learned model captures the true behavior in regimes beyond the training samples.

\begin{table}[H]
  \centering
  \caption{Parameters for the generalized Ogden model}
  \label{tab:ogden}
  \begin{tabular}{@{}lr lr lr@{}}
    \toprule
    \textbf{Parameter} & \textbf{Value} &
    \textbf{Parameter} & \textbf{Value} &
    \textbf{Parameter} & \textbf{Value} \\
    \midrule
    $c_{10}$ & 1.302 & $c_{20}$ & 0.261 & $c_{30}$ & 0.246 \\
    $c_{01}$ & 0.668 & $c_{02}$ & 0.245 & $c_{03}$ & 0.143 \\
    $\kappa$ & 0.831 & \multicolumn{2}{c}{} & \multicolumn{2}{c}{} \\
    \bottomrule
  \end{tabular}
\end{table}

\subsubsection{Full field dataset for transfer learning}

To close the loop of the multi-modal transfer learning scheme, we generate agenerate synthetic full-field data-set as a stand-in for DIC experiments. To enable this we deploy the virtual DIC experiments in a finite element setting. 
To this end, we use the compressible Neo-Hookean model,
\begin{equation}
    \varphi_\mathrm{neo}(I_1,I_2,J)
    =
    \frac{\mu}{2}
    (I_1-3)
    -
    \mu \, \ln{J}
    +
    \frac{\lambda}{2} (\ln{J})^2
    \label{eq:neo}
\end{equation}
with material parameters $\mu=1$ and $\lambda=0.333$,
and the generalized Ogden model
\begin{equation}\label{eq:ogden}
    \varphi_{\mathrm{ogden}}(I_1,I_2,J)
    =
    \sum_{i=1}^m 
    c_{i0}
    \bigg(
            \bar{I}_1 - 3
    \bigg)^i
    +
    \sum_{j=1}^n 
    c_{0j}
    \bigg(
            \bar{I}_2^{-3/2}  -  3\sqrt{3}
    \bigg)^j
    +
    \kappa\bigg( J^2 + J^{-2} -2\bigg),
\end{equation}
where
\begin{equation}
    \bar{I}_1 = J^{-2/3}\,I_1, \quad \bar{I}_2 = J^{-4/3}\,I_2,
\end{equation}
and with material parameters as indicated in Table \ref{tab:ogden}.
It is important to note that the choice that the hyperelastic laws differ from the pre-training data (where the Gent-Gent model was chosen) is to emulate unknown material prototypes (\textit{target materials}) encountered in material design and testing. These target materials are oftentimes part of a materials prototyping design cycle, while the pre-training data correspond to well-studied materials of the same materials class.

\begin{figure}
    \centering
    \includegraphics[width=0.5\linewidth, clip=true, trim = 50mm 50mm 50mm 50mm]{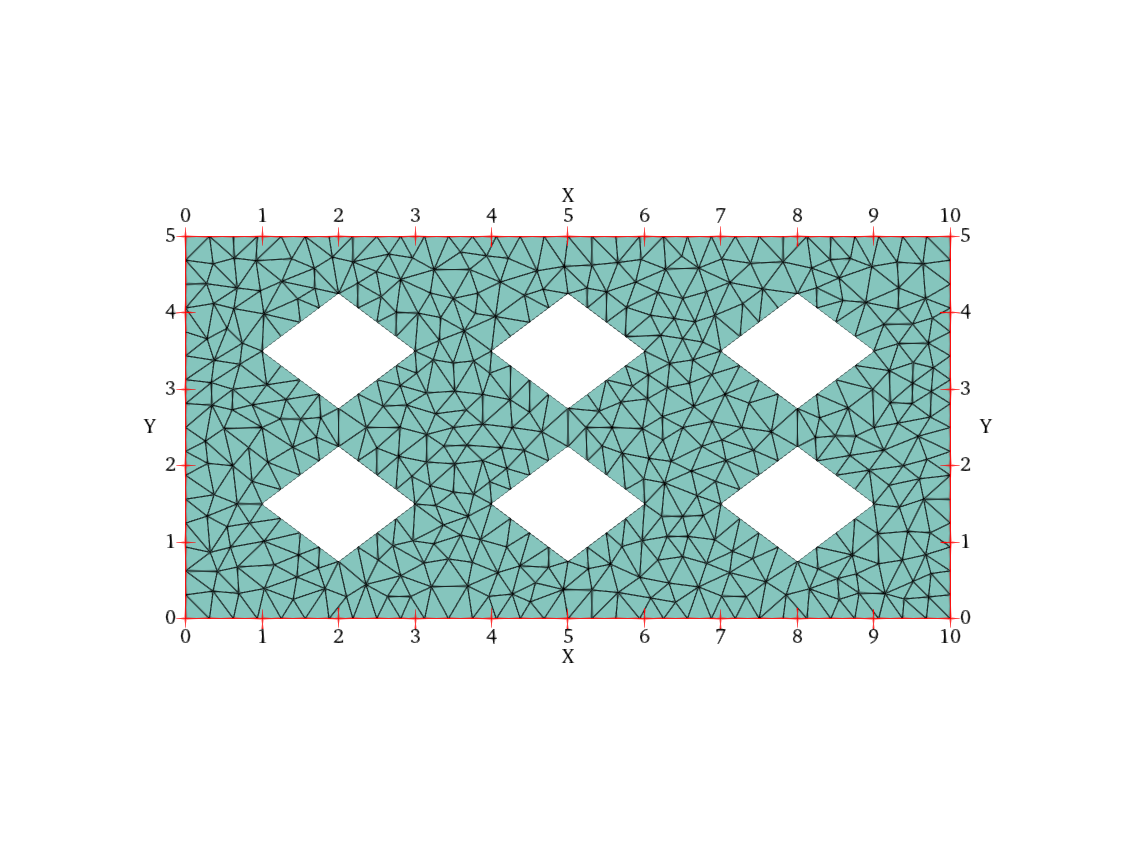}
    \caption{Discretized 2D domain of the digital image correlation speciment, displacement is fixed at $y=0$, a prescribed positive incremental displacement is applied uniformly at $y=5$. The lateral sides and the internal holes are not constrained in any way. }
    \label{fig:mesh}
\end{figure}

Specimen geometry for the (virtual) DIC experiment dictates the level of spatial heterogeneity of the resulting stress state; it is noted that the observable is the full-field displacement field that can be transformed to strain, and there is no direct stress comparison between experiment and the forwards FE solver, as highlighted in the adjoint-based PDE-constrained optimization set-up. Even though in this work we do not focus on optimization of the specimen geometry to obtain richer and more informative full field data-sets, this will be the focus of an upcoming study. Figure \ref{fig:mesh} shows the discretized 2D domain of the synthetic specimen, with several features designed to induce a highly heterogeneous spatial distribution of stress states. For simplicity, the specimen is taken to be in plane strain conditions, and this is replicated in the numerical solution. The specimen is loaded with prescribed displacement up to $\Delta u_y = 2.5$ ($50\%$ strain) in 25 increments, and full-field displacements and the corresponding homogenized reaction forces are recorded at $2\%, 10\%, 20\%, 30\%, 40\%, 50\%$ strains, respectively. 

To mimic experimental DIC data, spatially correlated noise was added to the displacement field. In practice, DIC measurements contain errors arising from camera noise (e.g., gray-level intensity fluctuations) and matching errors during speckle pattern tracking \cite{jones2018good}. Because displacement values are computed over pixel subsets, which often overlap, the resulting errors exhibit spatial correlation with a characteristic length scale related to the subset size and other filtering methods \cite{bornert2009assessment}. 
Accordingly, the measurement noise for the displacement field is represented as a Gaussian random field (GRF) $\mathcal{G}(\mathbf{x})$ specified with a Mat\'ern kernel \cite{lindgren2011explicit} with covariance $\mathcal{C}=\mathcal{A}^{-2}$, sampled through the action of the self-adjoint differential operation $\mathcal{A}$:
\begin{equation}
\mathcal{A}\, \mathcal{G}
=
\begin{cases}
    \gamma\,\nabla\cdot(\nabla\,\mathcal{G}) + \delta\,\mathcal{G}
    \quad
    \mathrm{in\ }
    \Omega
    \\
    (\nabla\,\mathcal{G})\cdot\mathbf{n}
    +
    \frac{\sqrt{\delta\,\gamma}}{1.42}\,\mathcal{G}
    \quad
    \mathrm{on\ }
    \partial\Omega
\end{cases}
\end{equation}
Here, the isotropic spatial correlation length is controlled by the hyperparameters $\delta$ and $\gamma$ to be approximately 0.33 to comply with the finite element mesh size. For relevant applications, see \cite{tan2024scalable}.

For brevity, we do not showcase the FE solutions at this stage as they will be showcased as ground truth in the transfer learning stage. 
By generating data from two separate hyperelastic laws, we will be able to test the adaptability of our learned model to material responses of different complexity in the transfer learning stage.

\subsection{Pre-training on a limited dataset}
Following the NN-based constitutive law construction in Sec.\ref{sec:NN}, three ICNN variants are investigated. 
Each of these three variants consist of 2 layers with 200 hidden units each, and a softplus activation function for each layer. With biases only in the first linear layer, each variant has a maximum parameter count of 41400.  Each network is trained with the Adam optimizer for up to 2800 epochs and with batch size as 10, using a stepping learning rate starting at 0.1 and reduced by one order of magnitude every $700$ epochs. Additionally, the $L_{0}$ regularization for sparsification and the input-dependency penalty remain inactivate for the first 1000 epochs to allow the training to sufficiently explore the optimization space, and are activated with a linear warm-up period of $500$ epochs up to $w_{L_{0}} = 1$ and $w_{\mathrm{input}}=1e4$.

In order to address the balance of trustworthiness and expressivity under thermodynamical constraints, we vary the enforcement of \textit{polyconvexity} in two ways: by hard constraints in the network architecture, and by soft constraints in the loss. 
In particular, a \textbf{polyconvex neural network constitutive model}
guarantees polyconvexity by construction. 
Following PCNN of Sec.\ref{sec:NN}, this is constructed by ensuring all neural weights corresponding to the inputs $I_1$ and $I_2$ to be strictly positive. 
With the weights strictly being forbidden to explore negative values during training, this enforces polyconvexity as a hard constraint. 
Following the derivation of the polyconvex indicator inequalities in Sec.\ref{sec:NN}, 
a \textbf{relaxed ICNN 
constitutive model} is constructed by removing the restrictions of the  positivity constraint (as it pertains to the weights corresponding to the inputs $I_1$ and $I_2$) in the PCNN network architecture, while adding the inequalities \eqref{eq:polyconvex_I1} and \eqref{eq:polyconvex_I2} to the loss function as a penalty over the training data empirically during training.  
This network no longer guarantees polyconvexity, while expanding the optimization space for increased expressivity. 
Instead, the network will attempt to comply with the inequality constraints where possible, promoting that polyconvexity is not violated. The polyconvexity indicator guarantees that polyconvexity is violated when it is not satisfied, while it allows (but does not require) for polyconvexity to hold when it is satisfied.
Lastly, an \textbf{unconstrained NN constitutive model} with neither the hard or soft constraints for polyconvexity is included as a reference model, this is the standard ICNN as defined in Sec.\ref{sec:NN}. 
Note that all three networks are input-convex, while differ in satisfaction of polyconvexity. 

Concluding the pre-training stage, the extreme sparsification of $L_0$, all three ICNNs reduce to compact and interpretable algebraic forms as follow: a 13-parameter 
PCNN constitutive model,
\begin{equation}
    \begin{aligned}
        \varphi_{\mathrm{NN}} ^{\mathrm{polyc}}
        &=
    \theta_{11} \log{\left(e^{2 \theta_{12} \log{\left(e^{2 I_{2} \theta_{13}} + 1 \right)}} + 1 \right)} + \theta_{1} \log{\left(e^{2 \theta_{2} \log{\left(e^{2 J \theta_{3}} + 1 \right)}} + 1 \right)} 
    \\
    &+ \theta_{4} \log{\left(e^{2 \theta_{5} \log{\left(e^{2 J \theta_{6}} + 1 \right)}} + 1 \right)} + \theta_{7} \log{\left(e^{2 \theta_{8} \log{\left(e^{2 I_{1} \theta_{10} + 2 \theta_{9}} + 1 \right)}} + 1 \right)}
    \end{aligned},
    \label{eq:polyconvexNN}
\end{equation}
a 9-parameter relaxed ICNN constitutive model,
\begin{equation}
    \varphi_{\mathrm{NN}}^{\mathrm{rIC}} = 
    \theta_{1} \log{\left(e^{2 I_{1} \theta_{2} + 2 \theta_{3} \log{\left(e^{2 I_{2} \theta_{4}} + 1 \right)}} + 1 \right)} + \theta_{5} \log{\left(e^{2 \theta_{6} \log{\left(e^{2 I_{2} \theta_{7}} + 1 \right)} + 2 \theta_{8} \log{\left(e^{2 J \theta_{9}} + 1 \right)}} + 1 \right)},
\label{eq:relaxedICNN}
\end{equation}
and a 9-parameter unconstrained NN constitutive model, 
\begin{equation}
    \varphi_{\mathrm{NN}}^{\mathrm{uNN}} =
    \theta_{1} \log{\left(e^{2 I_{1} \theta_{2} + 2 \theta_{3} \log{\left(e^{2 I_{2} \theta_{4}} + 1 \right)}} + 1 \right)} + \theta_{5} \log{\left(e^{2 \theta_{6} \log{\left(e^{2 I_{2} \theta_{7}} + 1 \right)} + 2 \theta_{8} \log{\left(e^{2 J \theta_{9}} + 1 \right)}} + 1 \right)}.
    \label{eq:unconstrainedICNN}
\end{equation}
The correponding parameter values are listed in Table \ref{tab:three_sets}. Notably, the relaxed and unconstrained NNs reduce into the same algebraic form, and only differ in parameter values. The training history and R2 scores of all three models are available in Figs. \ref{fig:history_poly}--\ref{fig:history_unc} in Appendix \ref{appdx:params}. Overall, all training losses show stability at their minimum, indicating a good convergence to an accurate fit. 
A notable observation lies on the R2 testing scores between epoch 500 and epoch 1000 (pre-sparsification), and between epoch 1000 and epoch 1500 (peri-sparsification). All three models demonstrate improvements on testing (generalization) performance as sparsification starts and warms up, highlighting the importance of parsimonious representations. Starting from 41400 parameters all three cases showcase efficient extreme sparsification, without requiring an intelligent initial guess. 
\begin{table}[h!]
  \centering
  \caption{Pre-trained model parameter sets (rounded to three decimals, see appendix for full precision). Here, polyc is polyconvex NN, rIC is relaxed ICNN, and uNN is unconstrained NN.}
  \label{tab:three_sets}
  \begin{tabular}{@{}lr lr lr@{}}
    \toprule
    \multicolumn{2}{c}{Set 1: $\theta^{\mathrm{polyc}}_{i}$} &
    \multicolumn{2}{c}{Set 2: $\theta^{\mathrm{rIC}}_{j}$} &
    \multicolumn{2}{c}{Set 3: $\theta^{\mathrm{uNN}}_{k}$} \\
    \cmidrule(lr){1-2}\cmidrule(lr){3-4}\cmidrule(lr){5-6}
    \textbf{Param} & \textbf{Value} &
    \textbf{Param} & \textbf{Value} &
    \textbf{Param} & \textbf{Value} \\
    \midrule
    $\theta^{\mathrm{polyc}}_{1}$  & 0.999 & $\theta^{\mathrm{rIC}}_{1}$ & 0.905 & $\theta^{\mathrm{uNN}}_{1}$ & 0.782 \\
    $\theta^{\mathrm{polyc}}_{2}$  & 5.227 & $\theta^{\mathrm{rIC}}_{2}$ & 0.506 & $\theta^{\mathrm{uNN}}_{2}$ & 0.695 \\
    $\theta^{\mathrm{polyc}}_{3}$  & -0.696 & $\theta^{\mathrm{rIC}}_{3}$ & 1.674 & $\theta^{\mathrm{uNN}}_{3}$ & 3.124 \\
    $\theta^{\mathrm{polyc}}_{4}$  & 0.149 & $\theta^{\mathrm{rIC}}_{4}$ & -0.211 & $\theta^{\mathrm{uNN}}_{4}$ & -0.192 \\
    $\theta^{\mathrm{polyc}}_{5}$  & 1.577 & $\theta^{\mathrm{rIC}}_{5}$ & 1.429 & $\theta^{\mathrm{uNN}}_{5}$ & 1.520 \\
    $\theta^{\mathrm{polyc}}_{6}$  & -0.584 & $\theta^{\mathrm{rIC}}_{6}$ & 1.266 & $\theta^{\mathrm{uNN}}_{6}$ & 1.283 \\
    $\theta^{\mathrm{polyc}}_{7}$  & 0.080 & $\theta^{\mathrm{rIC}}_{7}$ & -0.772 & $\theta^{\mathrm{uNN}}_{7}$ & -0.803 \\
    $\theta^{\mathrm{polyc}}_{8}$  & 1.617 & $\theta^{\mathrm{rIC}}_{8}$ & 3.657 & $\theta^{\mathrm{uNN}}_{8}$ & 2.797 \\
    $\theta^{\mathrm{polyc}}_{9}$  & -3.232 & $\theta^{\mathrm{rIC}}_{9}$ & -0.521 & $\theta^{\mathrm{uNN}}_{9}$ & -0.575 \\
    $\theta^{\mathrm{polyc}}_{10}$ & 1.279 & \multicolumn{2}{c}{} & \multicolumn{2}{c}{} \\
    $\theta^{\mathrm{polyc}}_{11}$ & 0.029 & \multicolumn{2}{c}{} & \multicolumn{2}{c}{} \\
    $\theta^{\mathrm{polyc}}_{12}$ & 0.071 & \multicolumn{2}{c}{} & \multicolumn{2}{c}{} \\
    $\theta^{\mathrm{polyc}}_{13}$ & 0.042 & \multicolumn{2}{c}{} & \multicolumn{2}{c}{} \\
    \bottomrule
  \end{tabular}
\end{table}
\begin{figure}[h!]
    \centering
    \includegraphics[width=0.45\linewidth]{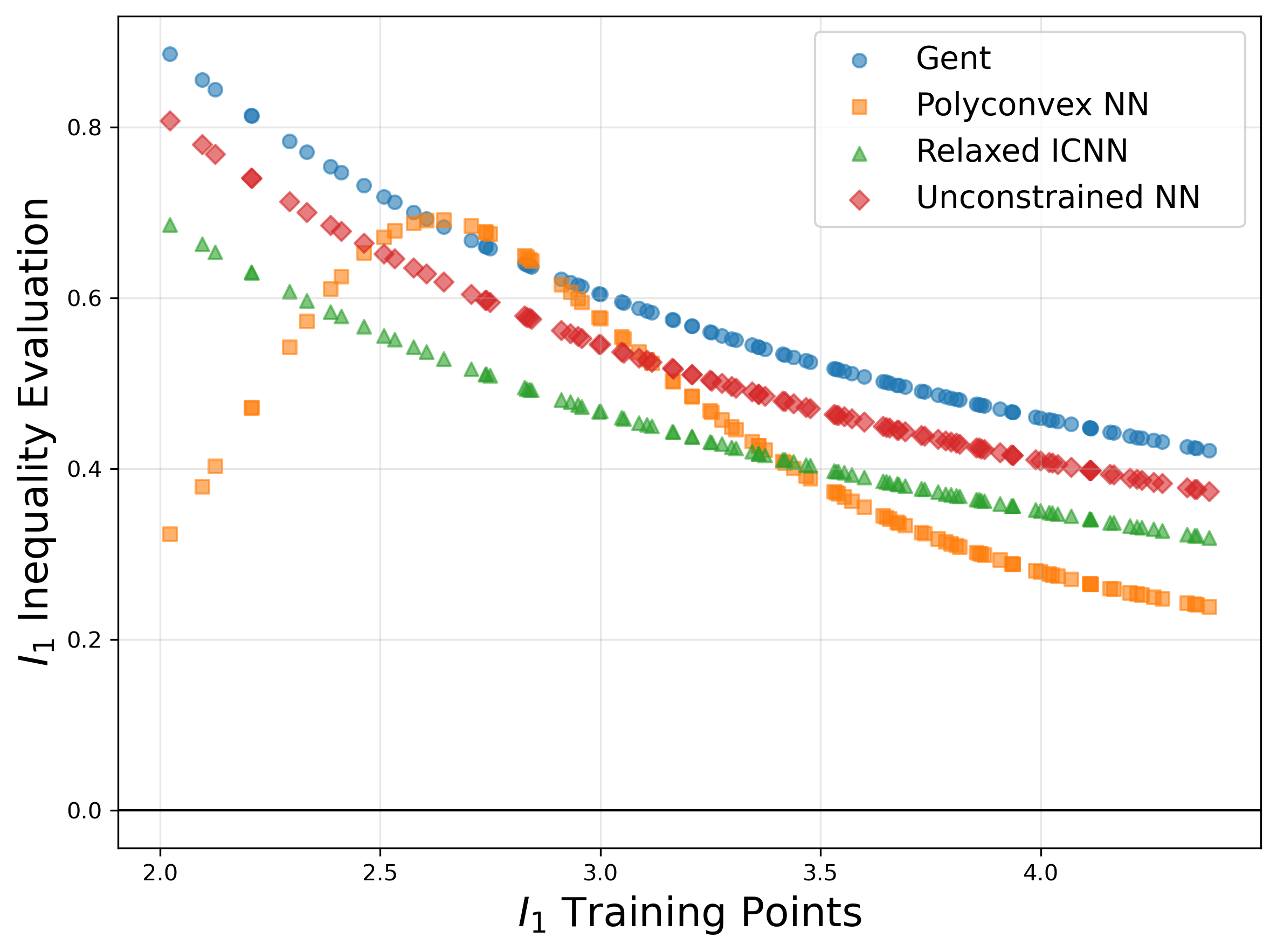}
    ~
    \includegraphics[width=0.45\linewidth]{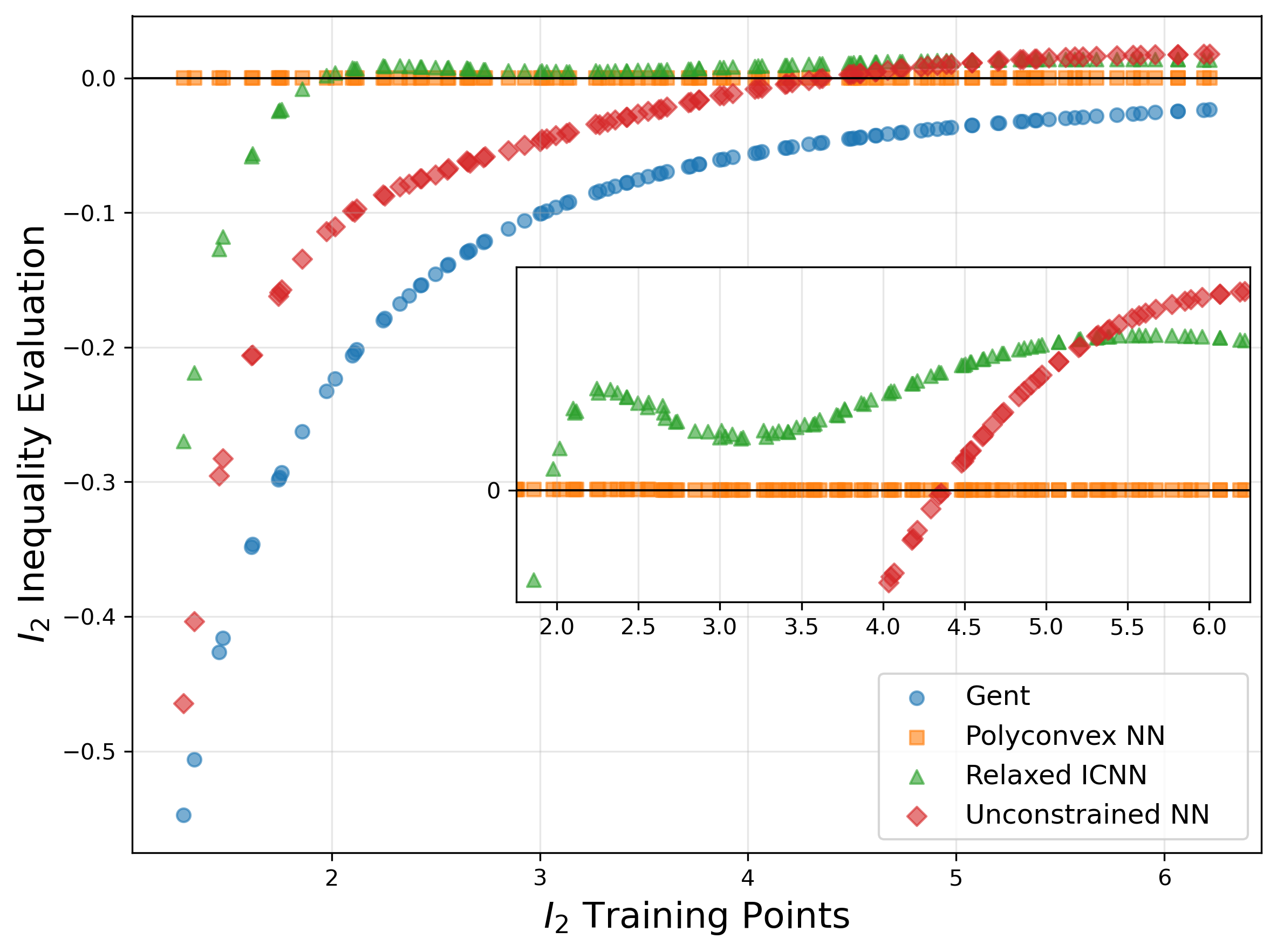}
    \caption{Evaluation of the polyconvexity indicator inequalities over training data. Positive values indicate satisfaction of the indicator inequalities.}
    \label{fig:inequality}
\end{figure}

Figure \ref{fig:inequality} evaluates the polyconvexity indicator inequalities \eqref{eq:polyconvex_I1} and \eqref{eq:polyconvex_I2} --for brevity, each is referred to as $I_1$ and $I_2$ inequalities-- over all training points for all three model variants, as well as for the analytical Gent-Gent model \eqref{eq:gent}, from which the training dataset is developed. 
While all four models satisfy the $I_1$ inequality  \eqref{eq:polyconvex_I1} at the training points, the results for the $I_2$ inequality \eqref{eq:polyconvex_I2} provides a more complex picture. 
%
%
The Gent-Gent model, as expected, is not polyconvex at the training dataset locations in invariant space, as the I2 inequaliy is not satisfied for all the evaluations.
This lack of polyconvexity makes Gent-Gent a challenging model to learn with ICNN approximations that are designed to be polyconvex (PCNN). 
Enforcing such conditions in pre-training to exclude short wavelength instabilities may overly restrict expressivity.
As a result, by enforcing polyconvexity-by-construction, the PCNN model is drastically scaling down its dependence on $I_2$ to near zero, and in fact, removing it completely if the model is trained without the penalty to enforce input dependency. 
On the other hand, both the relaxed and the unconstrained NNs show that the $I_2$ inequality  \eqref{eq:polyconvex_I2} is satisfied for large $I_2$ values. 
This is expected as \eqref{eq:polyconvex_I2} shows a competition between
the negative contribution $\frac{\partial \hat{\varphi}_{\mathrm{NN}}}{\partial I_2}$ inversely scales with $I_2$ in the first term of \eqref{eq:polyconvex_I2} while $\frac{\partial^2 \hat{\varphi}_{\mathrm{NN}}}{\partial I_2^2}$ remains strictly positive due to the convex nature of ICNNs.
\begin{figure}[h!]
    \centering
    \includegraphics[width=\linewidth]{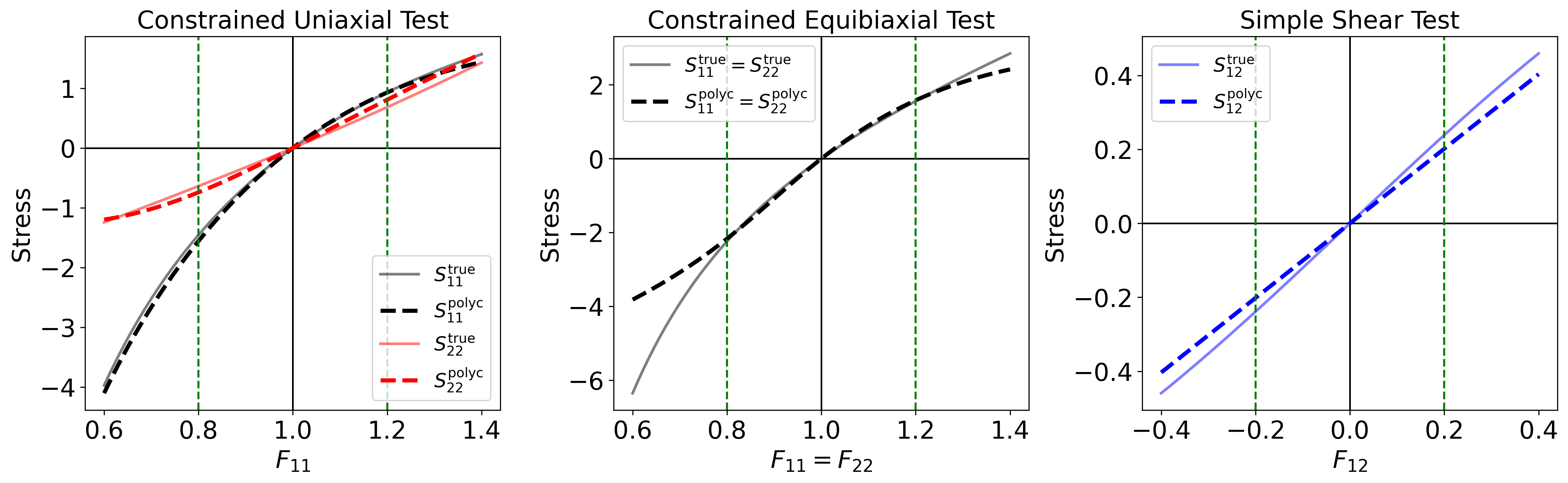}
    ~
    \includegraphics[width=\linewidth]{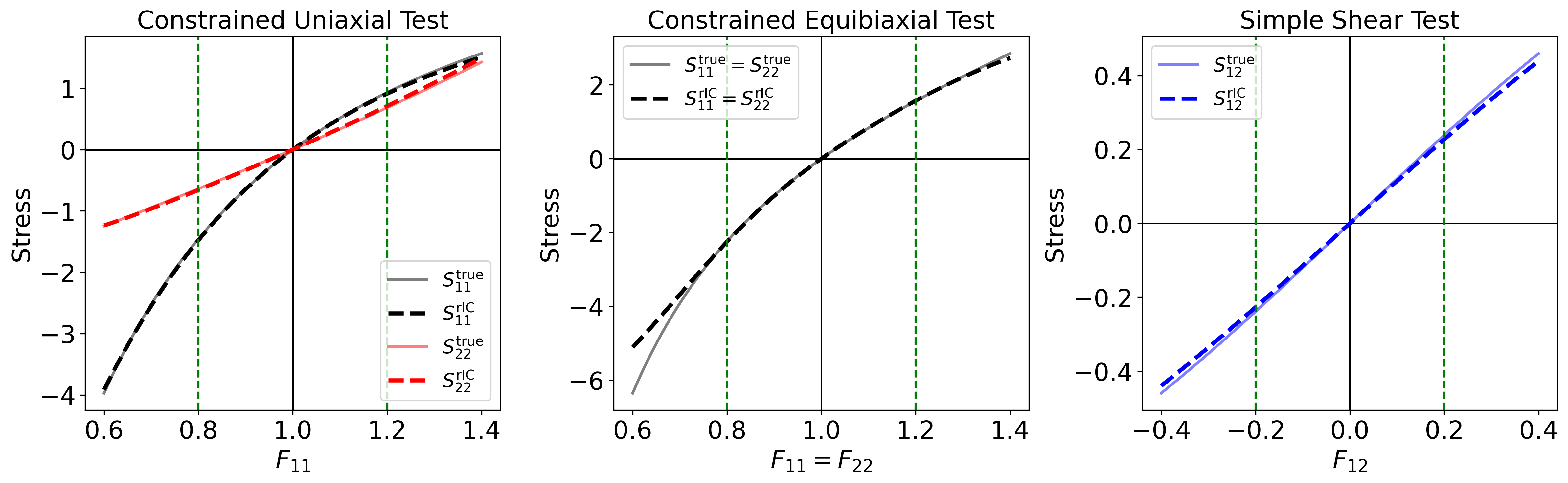}
    ~
    \includegraphics[width=\linewidth]{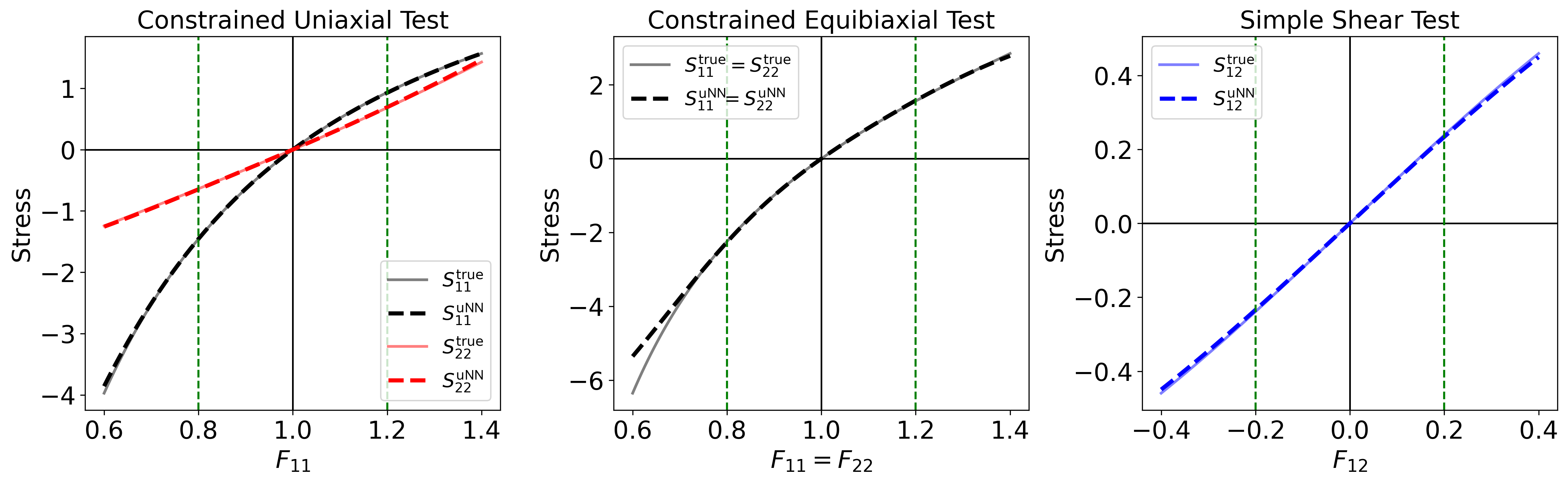}
    \caption{Generalization performance under constrained uniaxial, equibiaxial, and simple shear tests. The top row shows the polyconvex model, the middle row the relaxed model, and the bottom row the unconstrained NN model. Solid lines denote the ground-truth stresses, while dashed lines represent model predictions. Green dotted lines indicate the limits of the training regime used during learning.  Here, polyc is polyconvex NN, rIC is relaxed ICNN, and uNN is unconstrained NN.}
    \label{fig:constrained_tests}
\end{figure}
Furthermore, the use of the $I_2$ inequality \eqref{eq:polyconvex_I2} as a soft constraint guides the relaxed ICNN model to not violate the constraint for smaller $I_2$ values in a heuristic fashion, comparing it to the unconstrained NN model. 
Notably, as previously mentioned, the relaxed ICNN model \eqref{eq:relaxedICNN} and the unconstrained NN model \eqref{eq:unconstrainedICNN} reduce to the same algebraic form. 
The difference in the evaluations of the $I_2$ inequality \eqref{eq:polyconvex_I2} shown in Fig.\ref{fig:inequality} between these two models is merely due to the differences in parameter values (within the discovered 9-parameter models).

Figure \ref{fig:constrained_tests}, examines the effects of sparsification and polyconvexity enforcement in generalization through a validation test.  
The extrapolation ability of all three models is probed for the canonical loading path for constrained uni-axial tension as compared to the previously generated validation data in Fig.\ref{fig:gent_synthetic_data}. 
While all models demonstrates sufficient accuracy within training range, the PCNN model fails to generalize outside of the training range of [0.8, 1.2] for the constrained equibiaxial test, especially on larger compression ($F_{11}<0.8$), but also has issues recapitulating the response even within the training range for the simple shear test. 
On the other hand, both the unconstrained and the relaxed ICNN models behave sufficiently in the training range, but also demonstrate better ability to generalize out of the training range; it is noted that the unconstrained NN model has a slightly better performance. Both the unconstrained and the relaxed ICNN models deviate from the expected response in highly compressive states for the constrained equibiaxial test. 

\begin{figure}[h!]
    \centering
    \includegraphics[width=\linewidth]{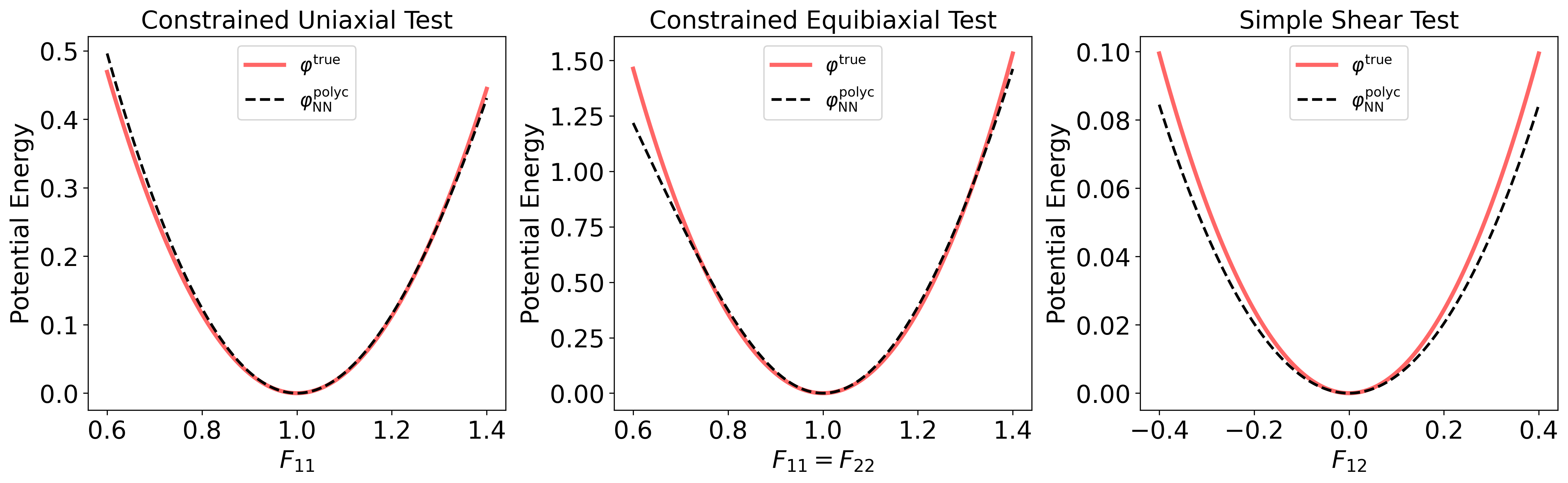}
    ~
    \includegraphics[width=\linewidth]{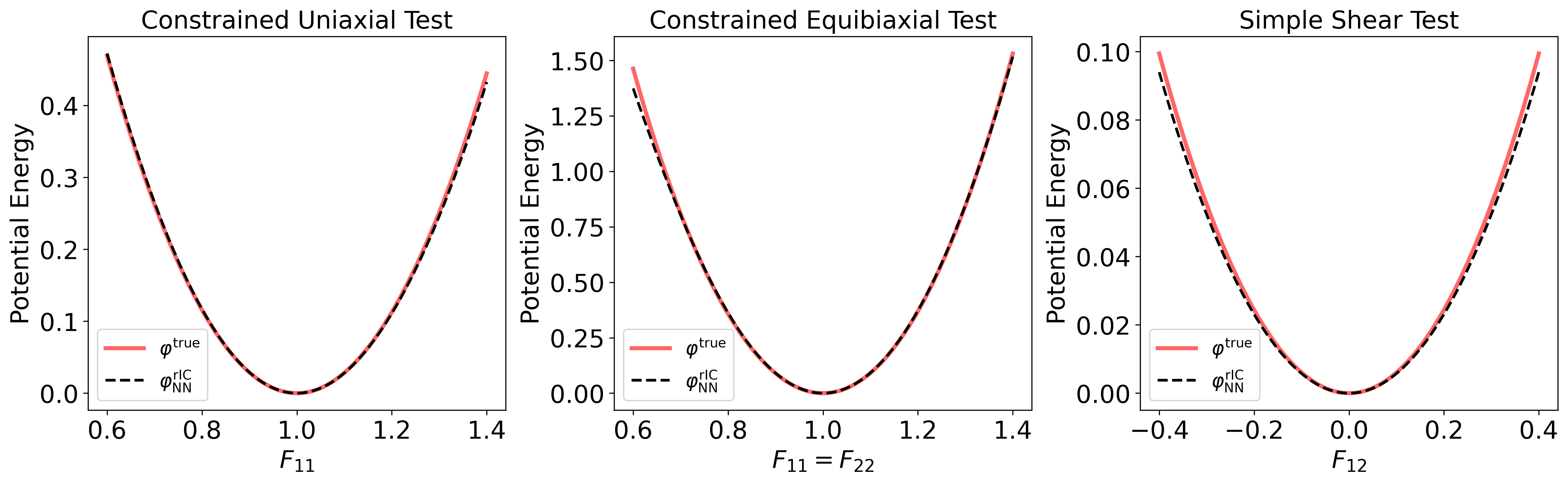}
    ~
    \includegraphics[width=\linewidth]{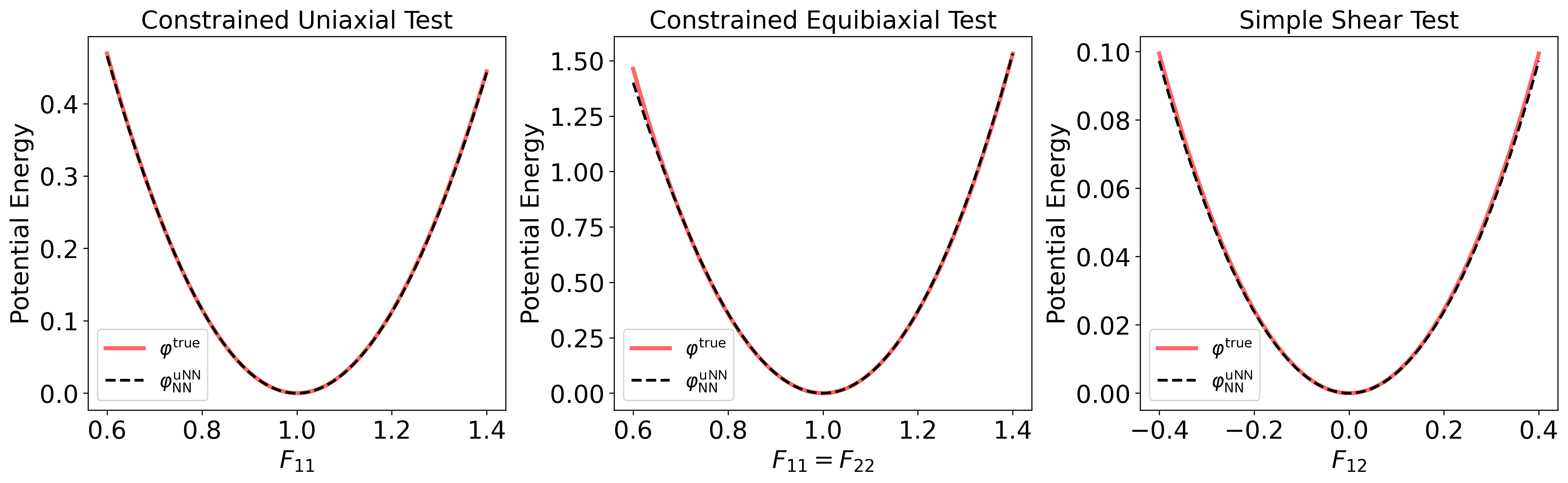}
    \caption{Validation of the learned strain energy functions along canonical loading paths over an extended range. The polyconvex, relaxed, and unconstrained NN models are shown in the top, middle, and bottom rows, respectively. Solid curves denote the ground-truth potential energy, while dashed curves represent the model predictions.  Here, polyc is polyconvex NN, rIC is relaxed ICNN, and uNN is unconstrained NN.}
    \label{fig:constrained_tests_potentials}
\end{figure}

\begin{figure}[h!]
    \centering
    \includegraphics[width=\linewidth]{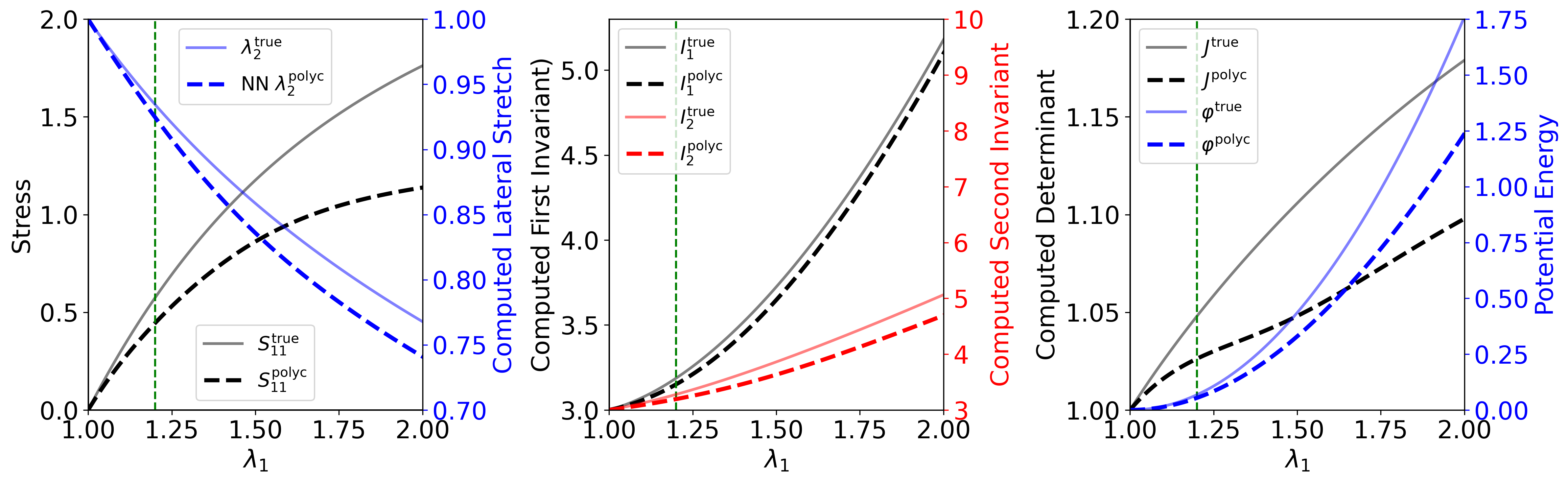}
    ~
    \includegraphics[width=\linewidth]{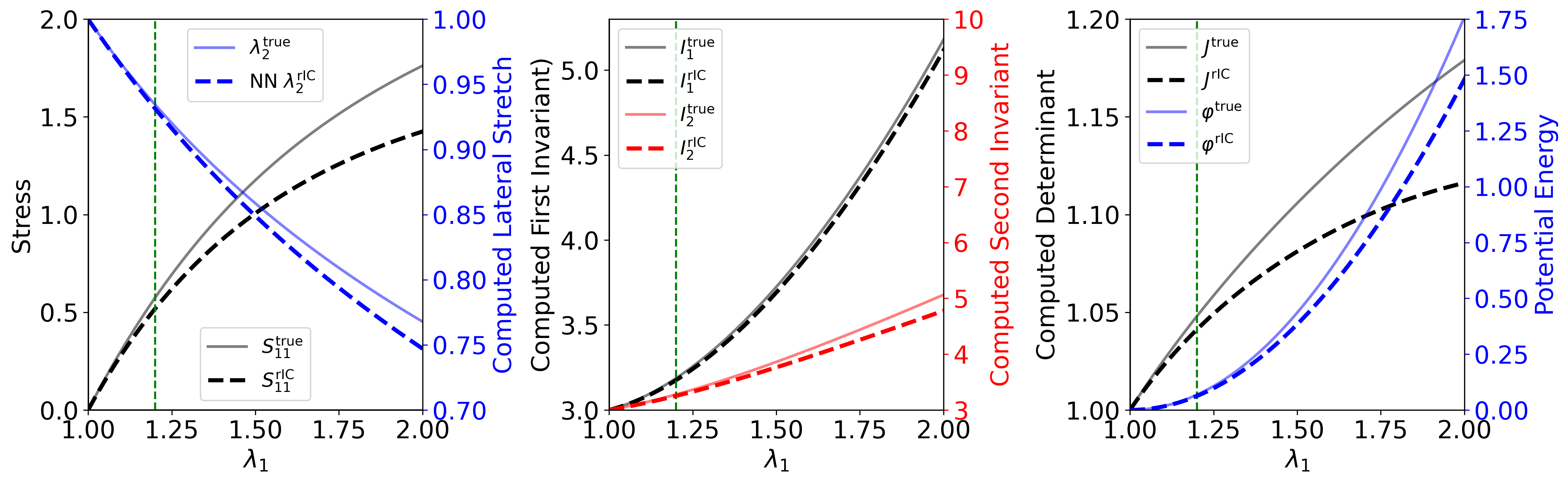}
    ~
    \includegraphics[width=\linewidth]{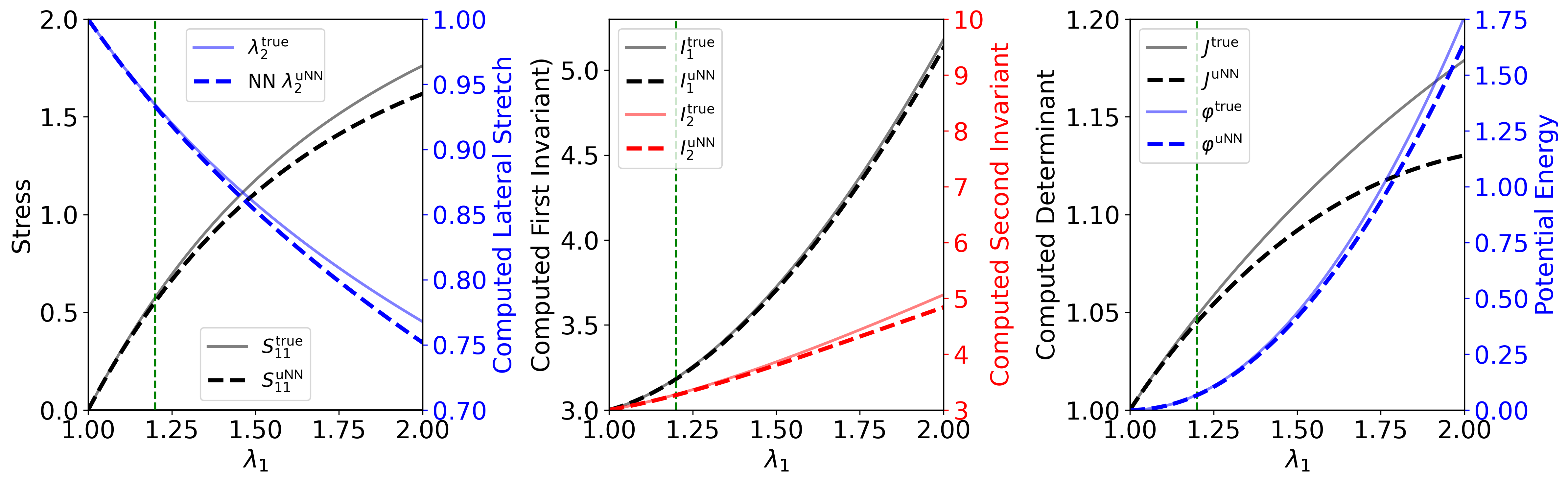}
    \caption{ Uniaxial test for examining performance of the models in a nonlinear solution scheme. Top row: polyconvex; middle row: relaxed; bottom row: unconstrained NN models.  Here, polyc is polyconvex NN, rIC is relaxed ICNN, and uNN is unconstrained NN.}
    \label{fig:NRtests}
\end{figure}

In Fig.\ref{fig:constrained_tests_potentials}, the potential energy values for the canonical paths are compared to the ground truth value, which is not seen directly by the model during training. The same qualitative conclusions  can be reached with what was discuss for Fig.\ref{fig:constrained_tests}.
To investigate potential numerical issues of the trained material models further, we perform a standard uniaxial tension test (not to be confused with the constrained uniaxial test used above) using a Newton-Raphson algorithm This is done outside of an FE setting, solely for a homogeneous state treated as a single integration point. 
This scenario requires accessing second derivatives of the NN potentials with respect to the deformation gradient for the calculation of a consistent material tangent required by the Newton-Raphson scheme.
%
%
In Fig. \ref{fig:NRtests} the normal stress in the direction of the loading  and the lateral stretch are shown with respect to the axial stretch up to $100\%$ tensile strain. 
The second and third columns show the values of the computed 
invariants $I_1, I_2, J$, and the potential $\varphi$, evaluated during this test. All three models do not encounter any numerical issues for the evaluation of the tangent, but the discrepancy of the learned models is more clearly observed compared to the ground truth results (Newton-Raphson over the Gent-Gent model). 
Additionally, it is worth to point out that while all three models demonstrate great accuracy in the training domain, the polyconvex model shows significant error in the determinant $J$ of the deformation gradient, even within the training range. This is showing that the hard enforcement of the polyconvexity constraint is restrictive in terms of expressivity.

\subsection{Transfer learning on full-field data} \label{subsec:transfer}

\begin{figure}[h!]
    \centering
    \includegraphics[width=\linewidth]{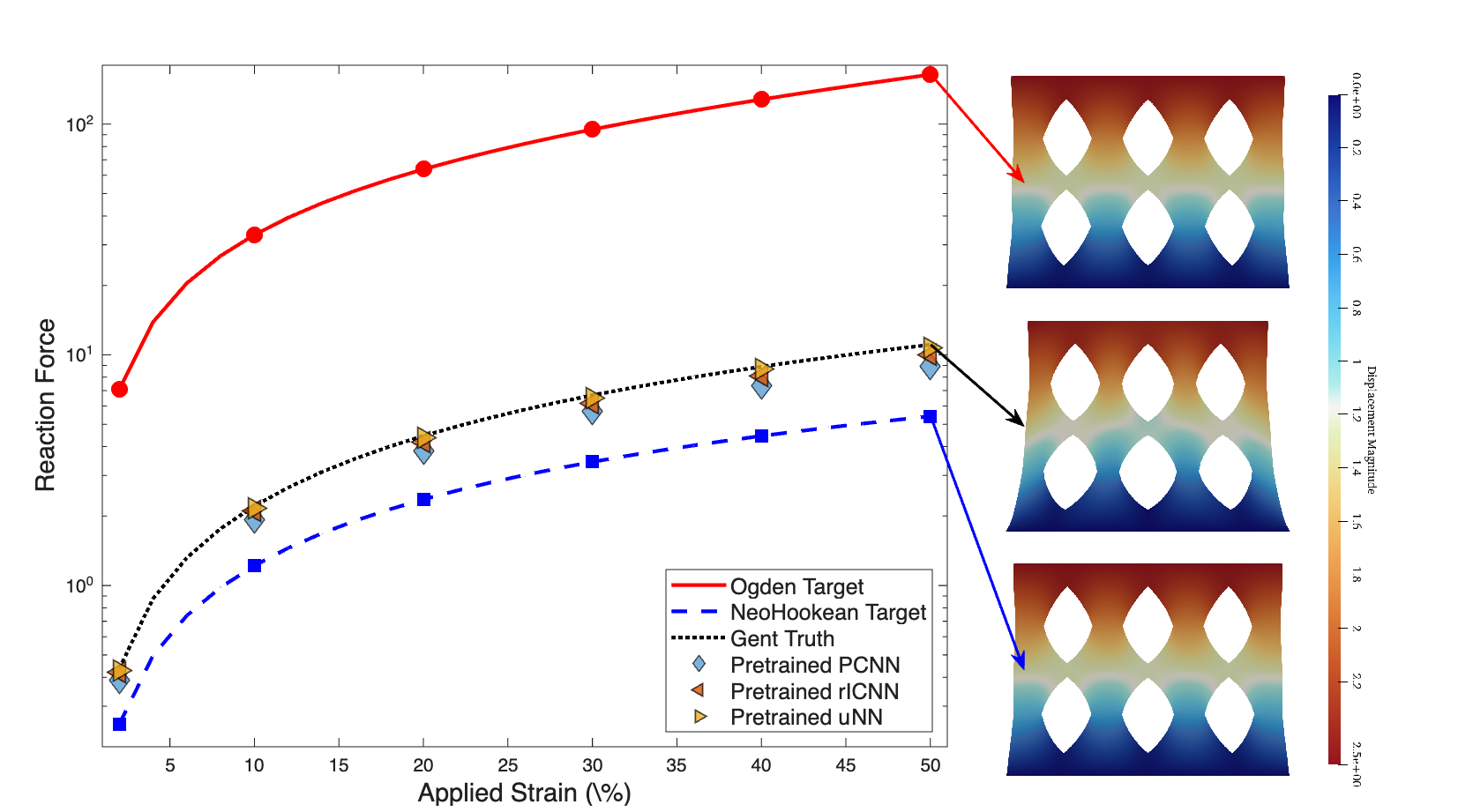}
    \caption{Comparison of transfer learning targets to the pretrained models as well as the pretrain truth at the strain points of which DIC data is generated through FEM. The markers on the Ogden and Neo-Hookean targets indicate where the data points collected for transfer learning. At each data collection point, the scalar-valued reaction force and the full-field displacement are collected for the Ogden and Neo-Hookean targets. For references, the displacement contours of the targets at the last collection point, as well as that of the Gent truth, are shown here. Here, PCNN is polyconvex NN, rICNN is relaxed ICNN, and uNN is unconstrained NN.}
    \label{fig:transfer_targets}
\end{figure}

Having established three ICNN variant of the Gent-Gent type material model through pre-training, the significant reduction of network parameters (from $\mathcal{O}^5$ to $\mathcal{O}^1$) enables a seamless utilization of the discovered constitutive laws in an FE-based adjoint framework utilizing more complex data to capture the target response. Pre-training enabled discovery of sparse representations, and in this transfer learning stage these will be further refined.
raises questions of potential loss of model expressivity, as this is often connected to over-parameterization. This is especially important as one could potentially aim to use a lower dimensional discovered model for transfer learning with a final target of higher complexity.
%
%
%
The pre-trained network now serves as an intelligent initial guess which will be updated (transferred) by assimilating higher-fidelity/full-field observations.
To this end, we use the synthetic DIC dataset  as the target for the transfer learning study.  
The choices of the two targets, utilizing the Neo-Hookean and generalized Ogden models for the generation of the synthetic DIC dataset, are intentionally chosen with model-form discrepancy with respect to the pre-training dataset generation (Gent-Gent model), to challenge the transferability and validity of the pre-trained models. 
In comparison to the material of Gent type \eqref{eq:gent} used for pre-training, the Neo-Hookean \eqref{eq:neo} model does not depend on the second invariant $I_2$, a feature that can be viewed as a reduction in material model complexity. 
On the other hand, the Ogden \eqref{eq:ogden} model probes the expressivity of the pre-trained model with increased complexity.
Fig.\ref{fig:transfer_targets} compares the two transfer targets to the true Gent model used during the pretraining stage, as well as the three pretrained model variants at the synthetic DIC data generating points. The transfer learning targets are chosen to respond with drastically differences from the pretrained models. 

The PDE-constrained optimization framework presented in Section \ref{subsec:adj} is implemented with \texttt{FEniCS} \cite{logg2012automated, alnaes2015fenics} and \texttt{SciPy} \cite{virtanen2020scipy}. 
This framework carries the low-dimensional pre-trained neural constitutive models as inputs to a finite element model coresponding to the  DIC specimen (Fig. \ref{fig:mesh}, and iteratively updates the neural constitutive parameters with analytical gradients computed through the adjoint method to minimize the discrepancy between the FE simulation outputs and the target data. 
This reduced dimensionality of the neural representations, allowes gradient-based optimization to be carried out at the level of the full-field problem within a robust and tested FE platform.

For each target material, the transfer learning scheme starts from the pre-trained neural parameters for each ICNN variant as a warm start, and performs iterative updates until the FE-predicted displacements field and reaction force matching those of the synthetic DIC data for every load-step throughout the nonlinear response. This matching is consideredin a least square sense as seen in Eq. \ref{eq:lagrange}. 
The results of this procedure is a set of updated neural constitutive parameters for each ICNN variant. For each target material scenario, the parameter values are recorded in Tab.\ref{tab:neohooke} for Neo-Hookean target and in Tab.\ref{tab:ogden} for Ogden target.

\begin{figure}[h!]
    \centering
    \includegraphics[width=\linewidth]{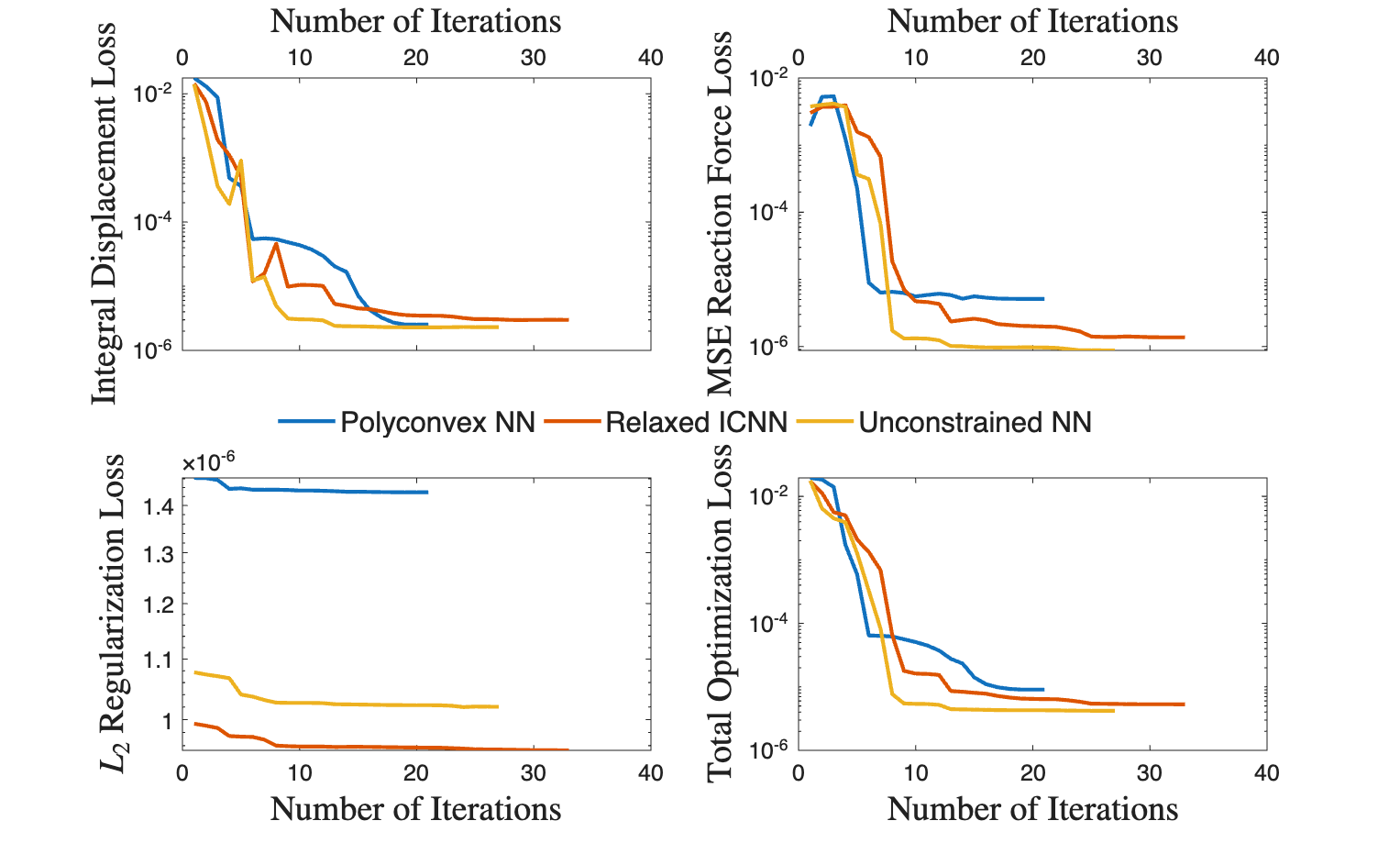}
    \caption{Loss components for transfer learning scheme, utilizing the Neo-Hookean synthetic DIC dataset}
    \label{fig:loss_neohooke}
\end{figure}

\begin{figure}[h!]
    \centering
    \includegraphics[width=0.6\linewidth]{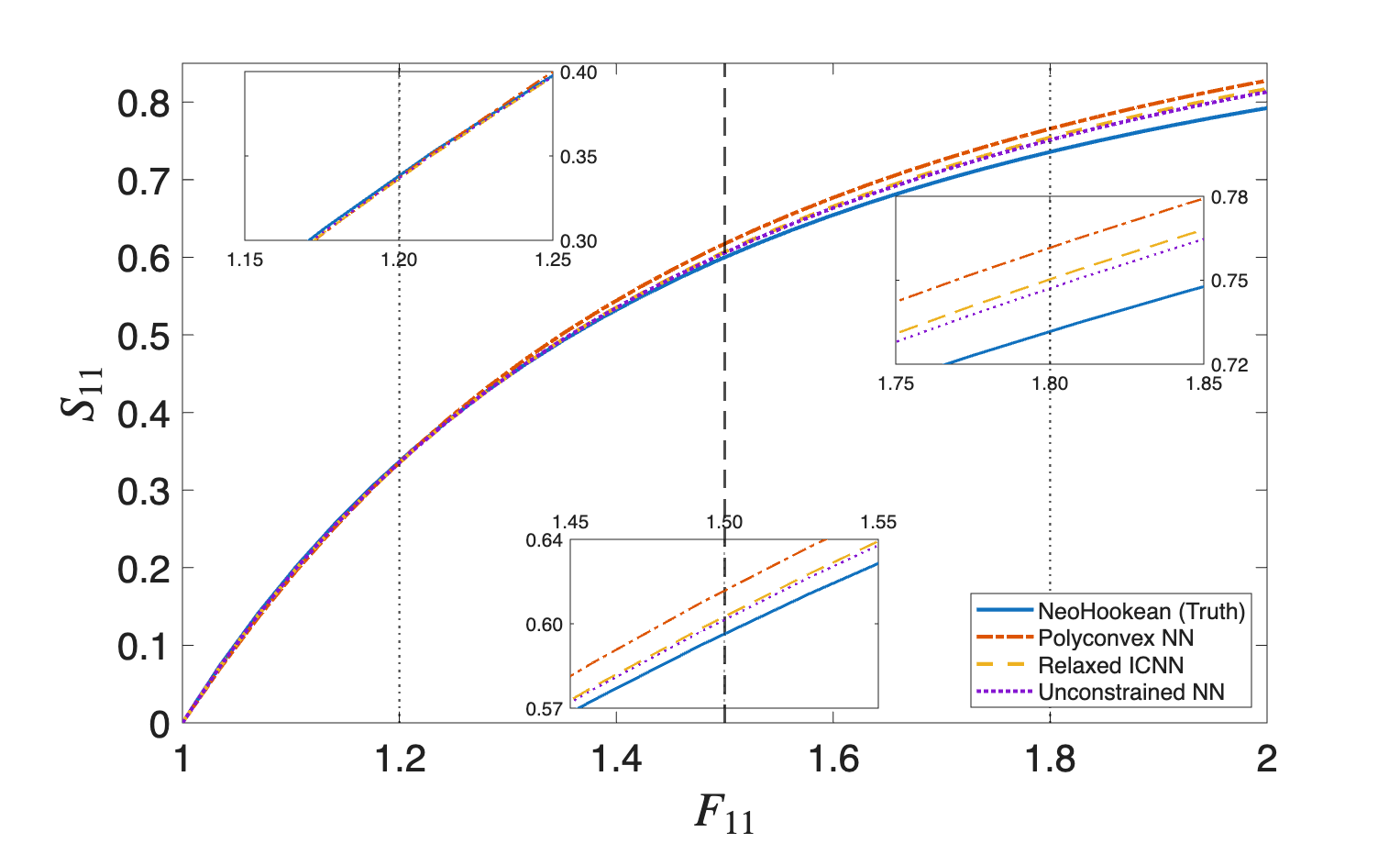}
    \caption{Transfer learning to Neo-Hookean DIC dataset: validation of learned models to the analytical Neo-Hookean model in a uniaxial tension setting via Newton-Raphson.}
    \label{fig:NR_neohooke}
\end{figure}

The transfer learning results show that all three ICNN variants successfully adapted to both new material behaviors, demonstrating impressive expressivity despite their extremely-sparsified forms. 
%
For the Neo-Hookean case, Fig.\ref{fig:loss_neohooke} shows that the optimization converged quickly for all three models, looking at different components of the loss, as well as the total loss. It is noted that the optimization scheme reaches a plateau in less than 20 iterations.
The final neural parameters (Tab.\ref{tab:neohooke}) differ from the pre-trained values (Tab.\ref{tab:three_sets}) in a way that reflects the simpler nature of Neo-Hookean compared to the Gent-Gent model. 
After transfer learning, all three models reproduced the Neo-Hookean response with high accuracy. 
To further validate the transferred ICNN models, we subjected them to a classical uniaxial tension test and compared the stress–strain behavior to the known Neo-Hookean analytical solution. 
%
Fig.\ref{fig:NR_neohooke},  shows close agreement for all three transferred ICNN models for uniaxial stress–stretch response in the Neo-Hookean target case. 
It is noteworthy that the Polyconvex ICNN appears to generally require more iterations to converge in the Newton Raphson scheme and exhibits slightly higher residual error than the constrained and unconstrained models, two of which perform virtually indistinguishable in this test.

\begin{figure}[h!]
    \centering
    \includegraphics[width=0.31\linewidth, clip=true, trim = 59.7mm 0mm 59.7mm 0mm]{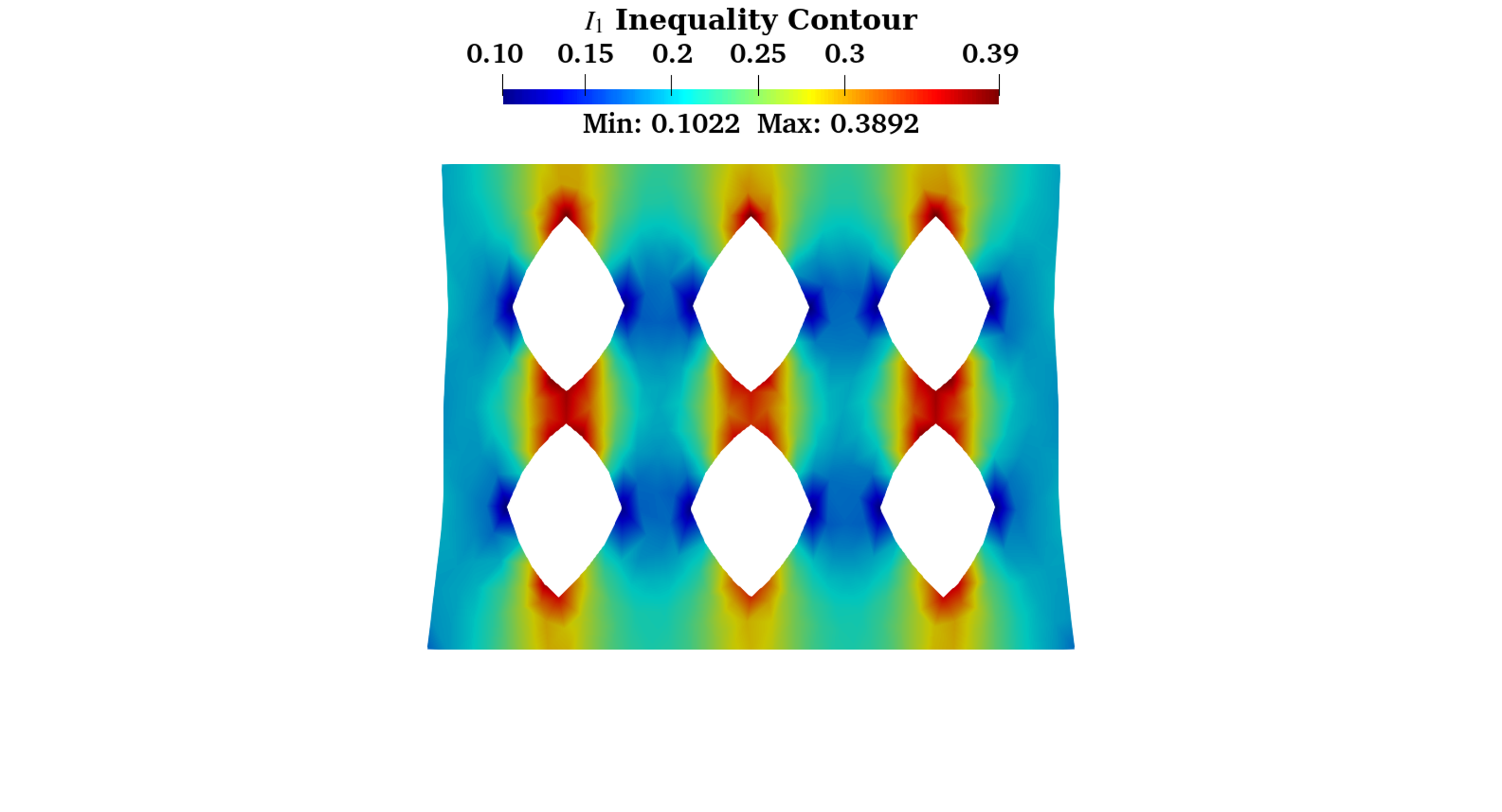}
    ~
    \includegraphics[width=0.31\linewidth, clip=true, trim = 155mm 0mm 155mm 0mm]{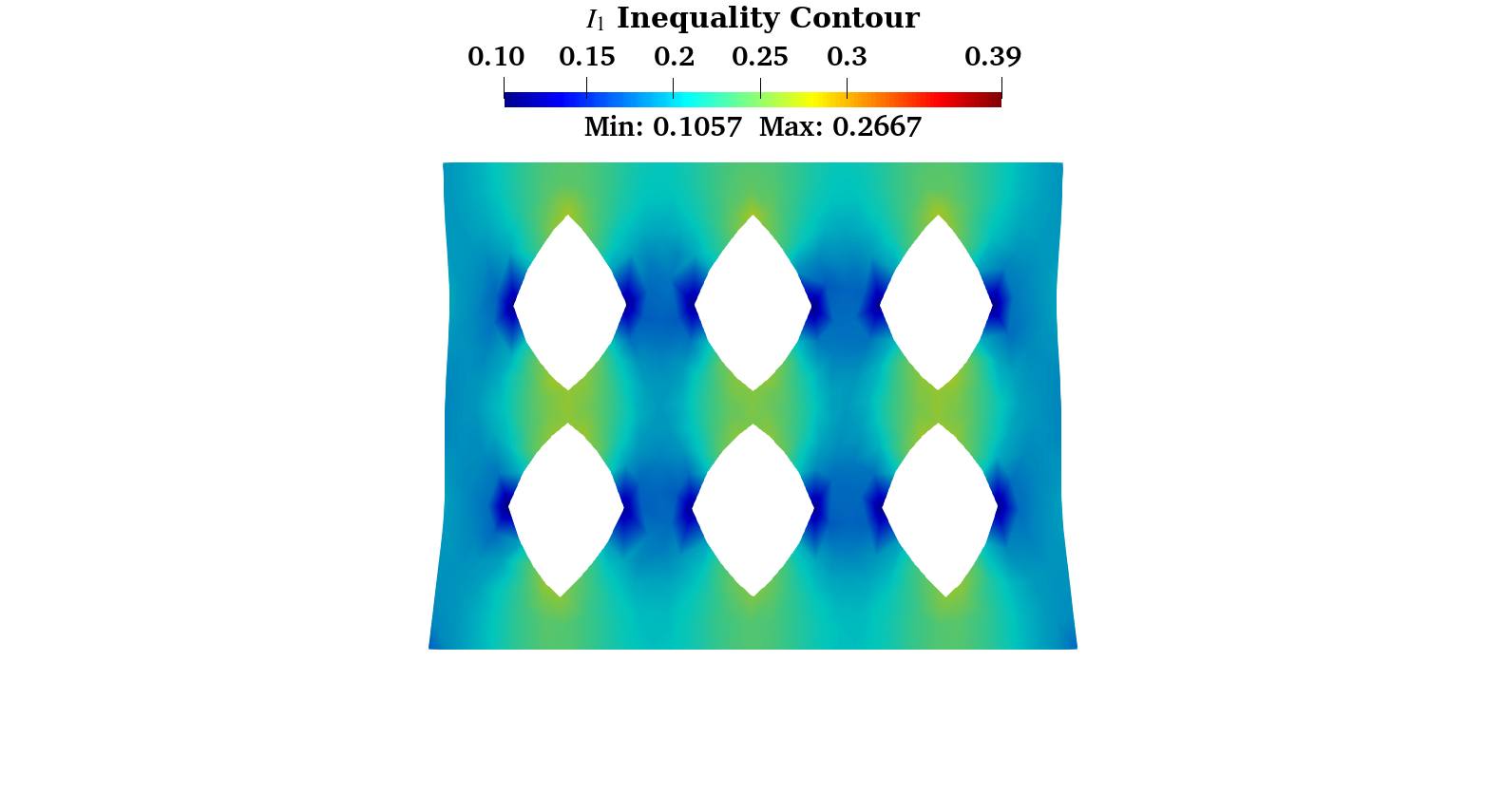}
    ~
    \includegraphics[width=0.31\linewidth, clip=true, trim = 155mm 0mm 155mm 0mm]{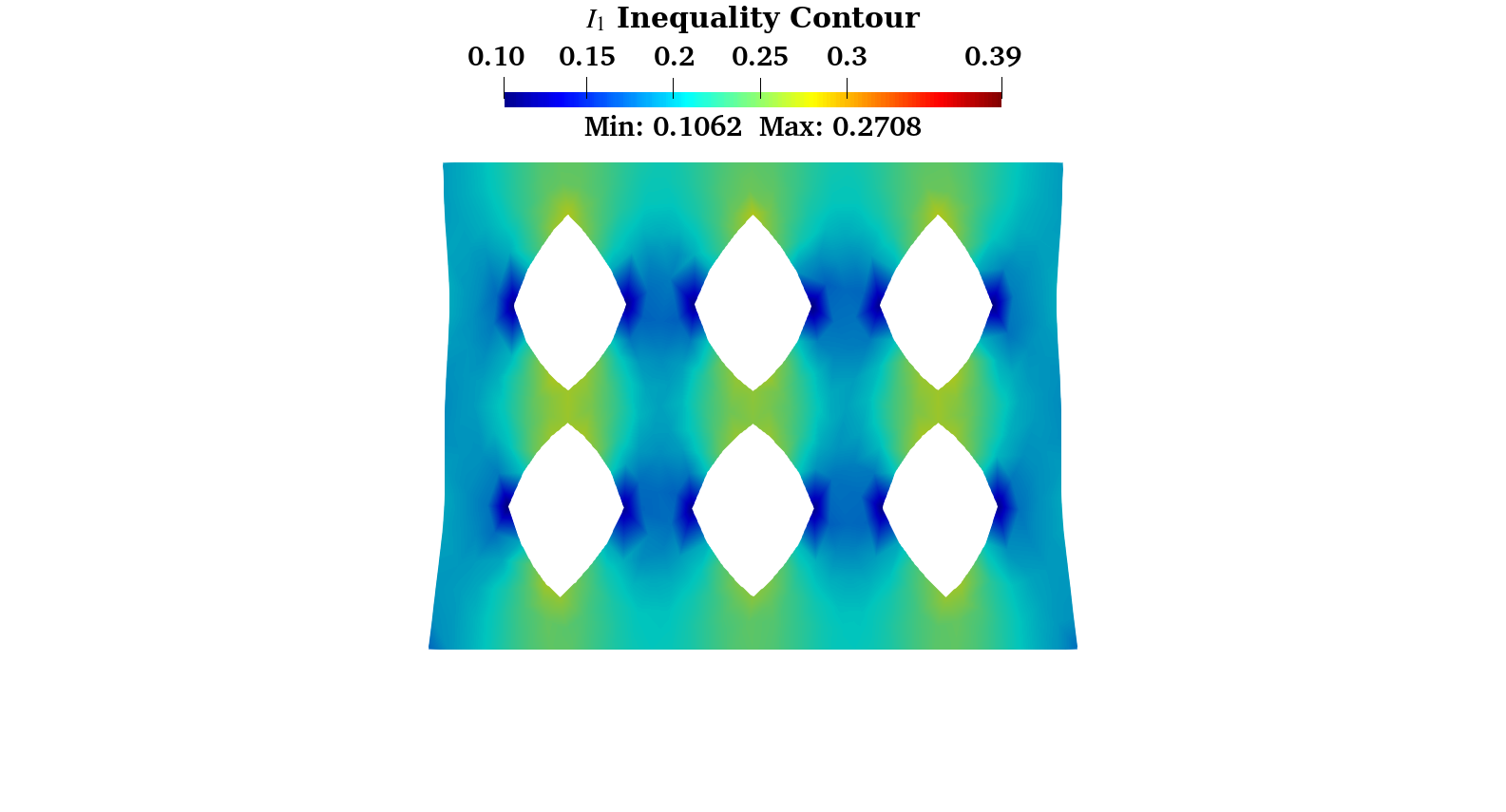}
    ~
    \includegraphics[width=0.31\linewidth, clip=true, trim = 155mm 0mm 155mm 0mm]{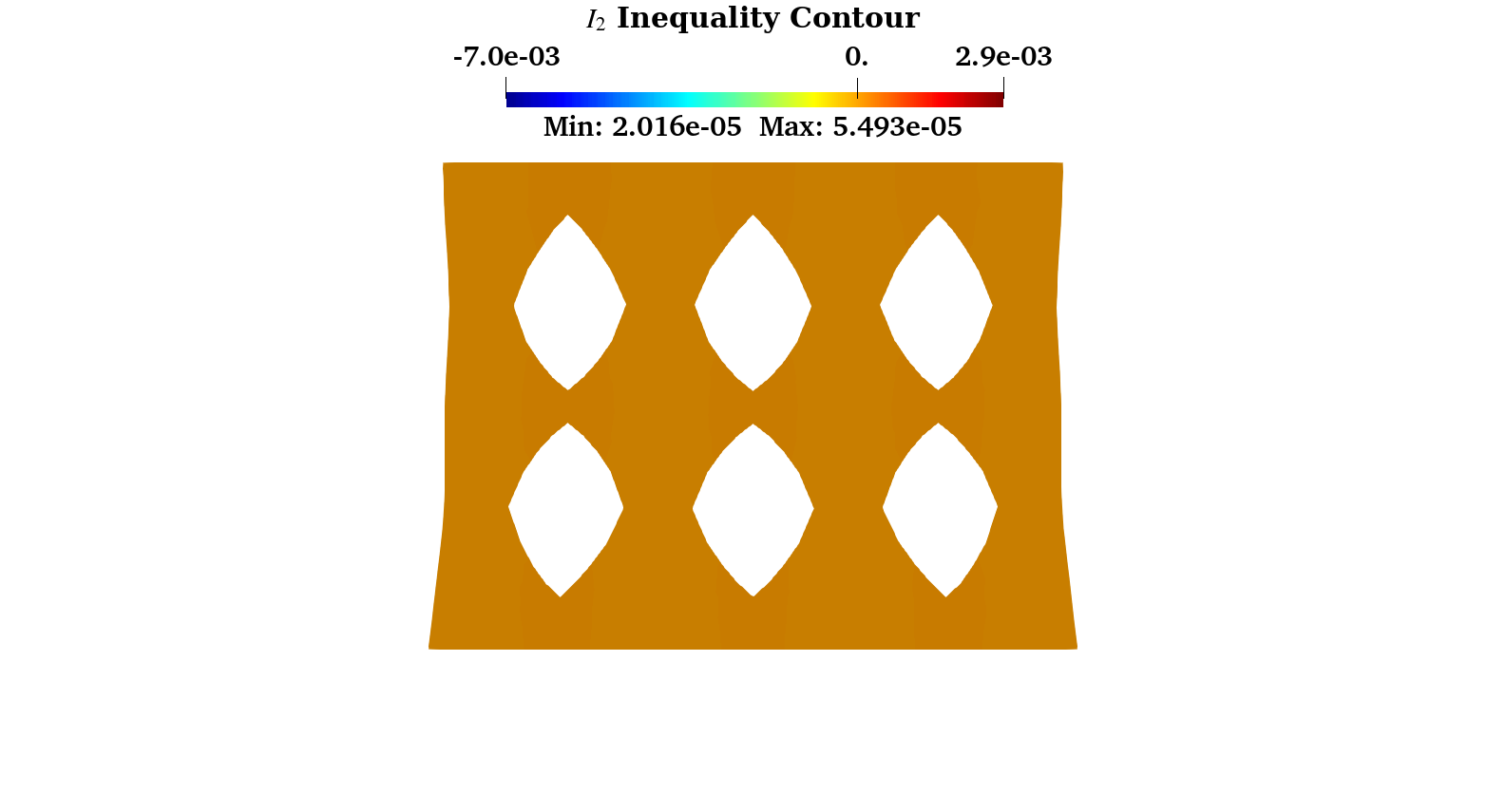}
    ~
    \includegraphics[width=0.31\linewidth, clip=true, trim = 155mm 0mm 155mm 0mm]{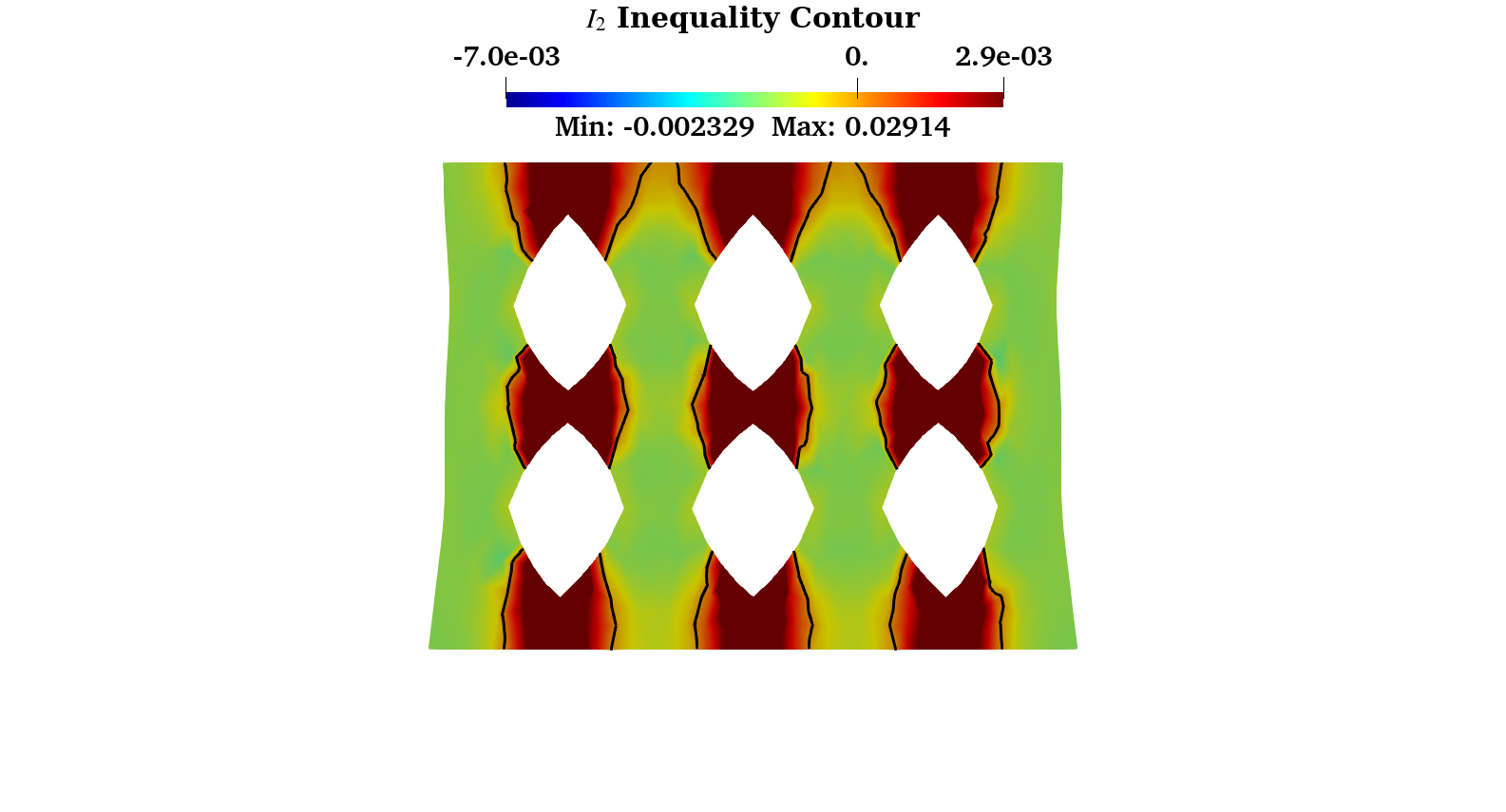}
    ~
    \includegraphics[width=0.31\linewidth, clip=true, trim = 155mm 0mm 155mm 0mm]{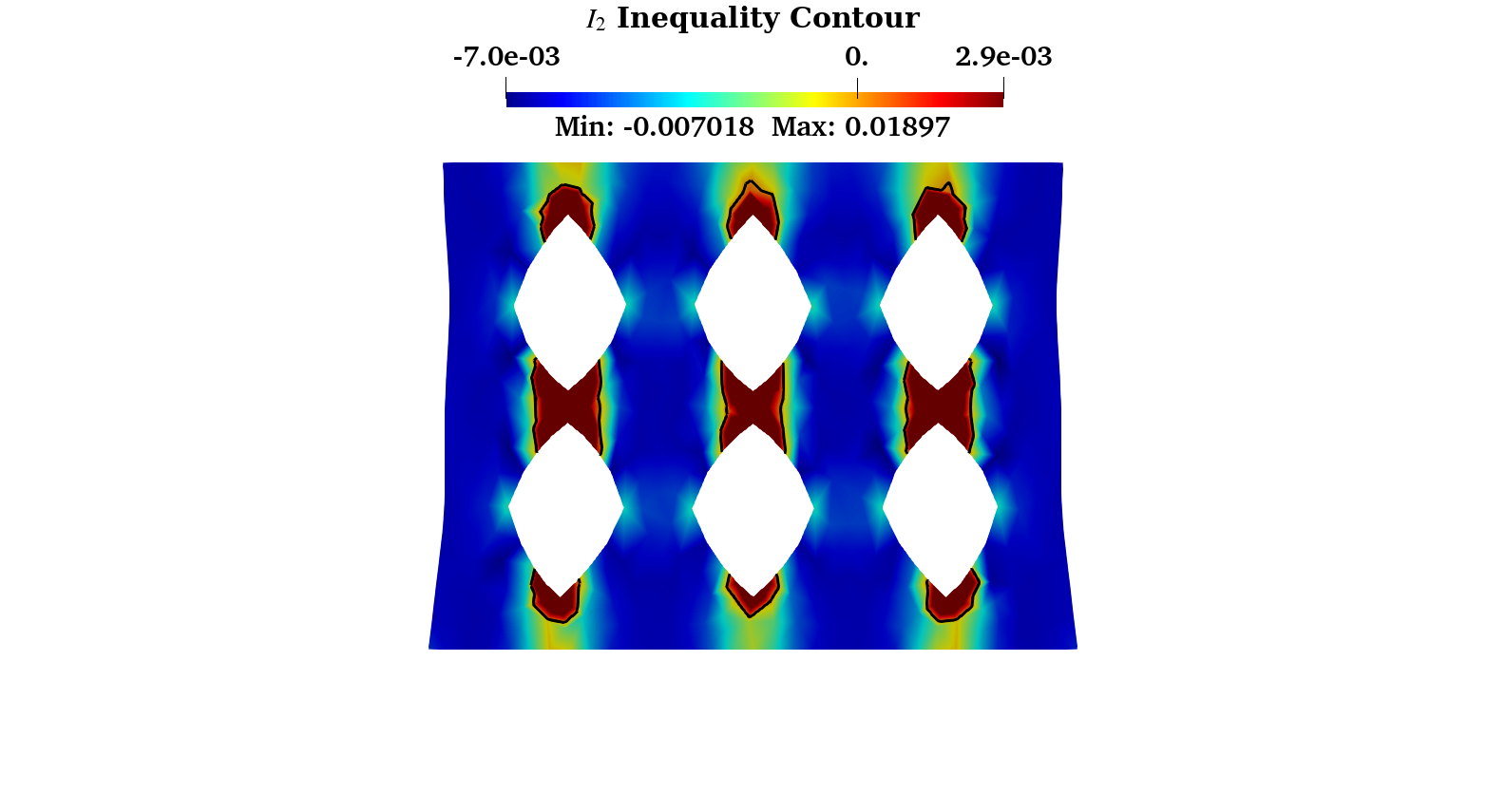}
    \caption{Polyconvexity indicator inequality evaluations, for polyconvex NN (left), reduced ICNN (middle) and unconstrained NN (right) models with the Neo-Hookean target.}
    \label{fig:trasf_neohooke}
\end{figure}

\begin{figure}[h!]
    \centering
    \includegraphics[width=0.32\linewidth, clip=true, trim = 140mm 50mm 140mm 0mm]{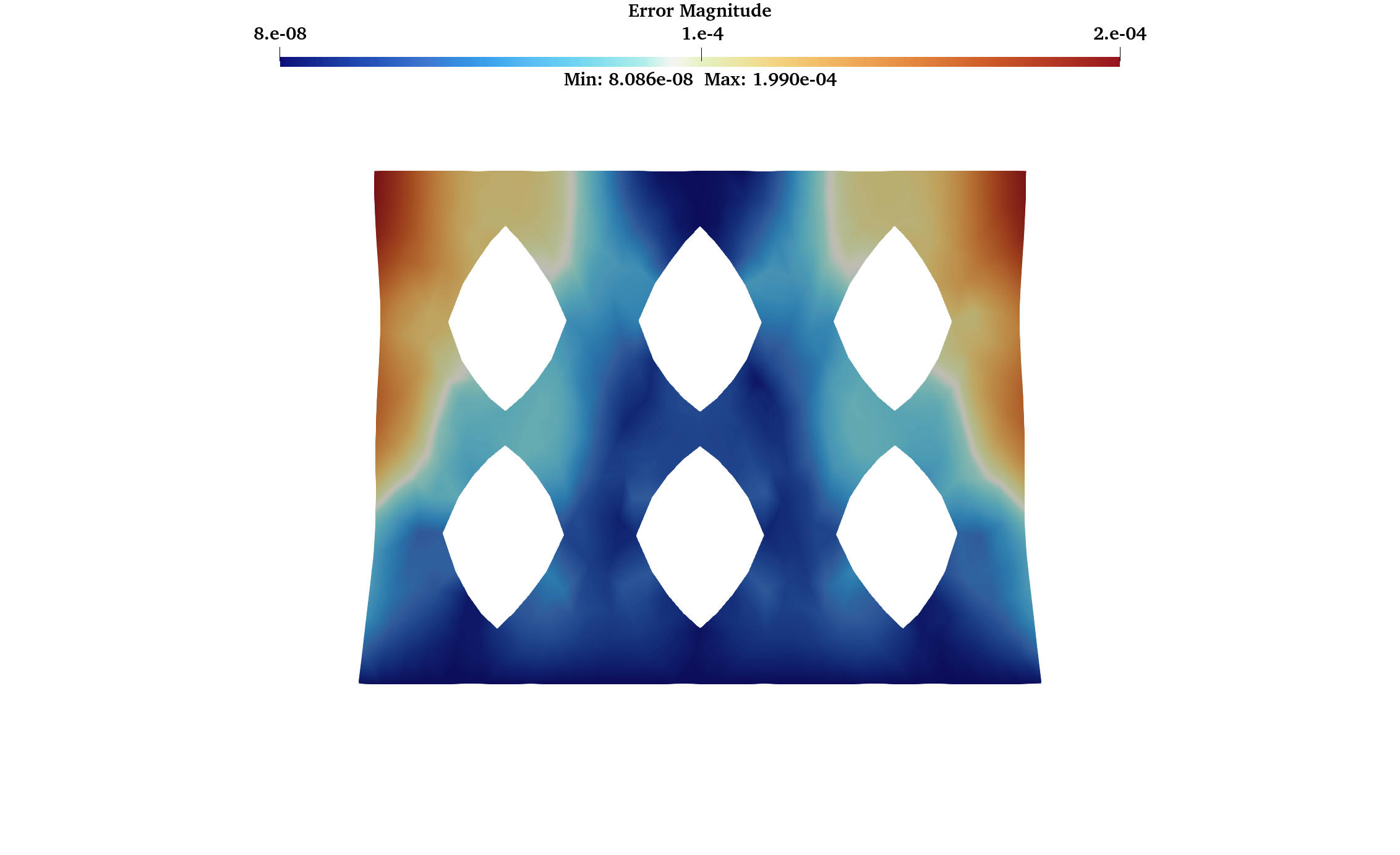}
    ~
    \includegraphics[width=0.32\linewidth, clip=true, trim = 140mm 50mm 140mm 0mm]{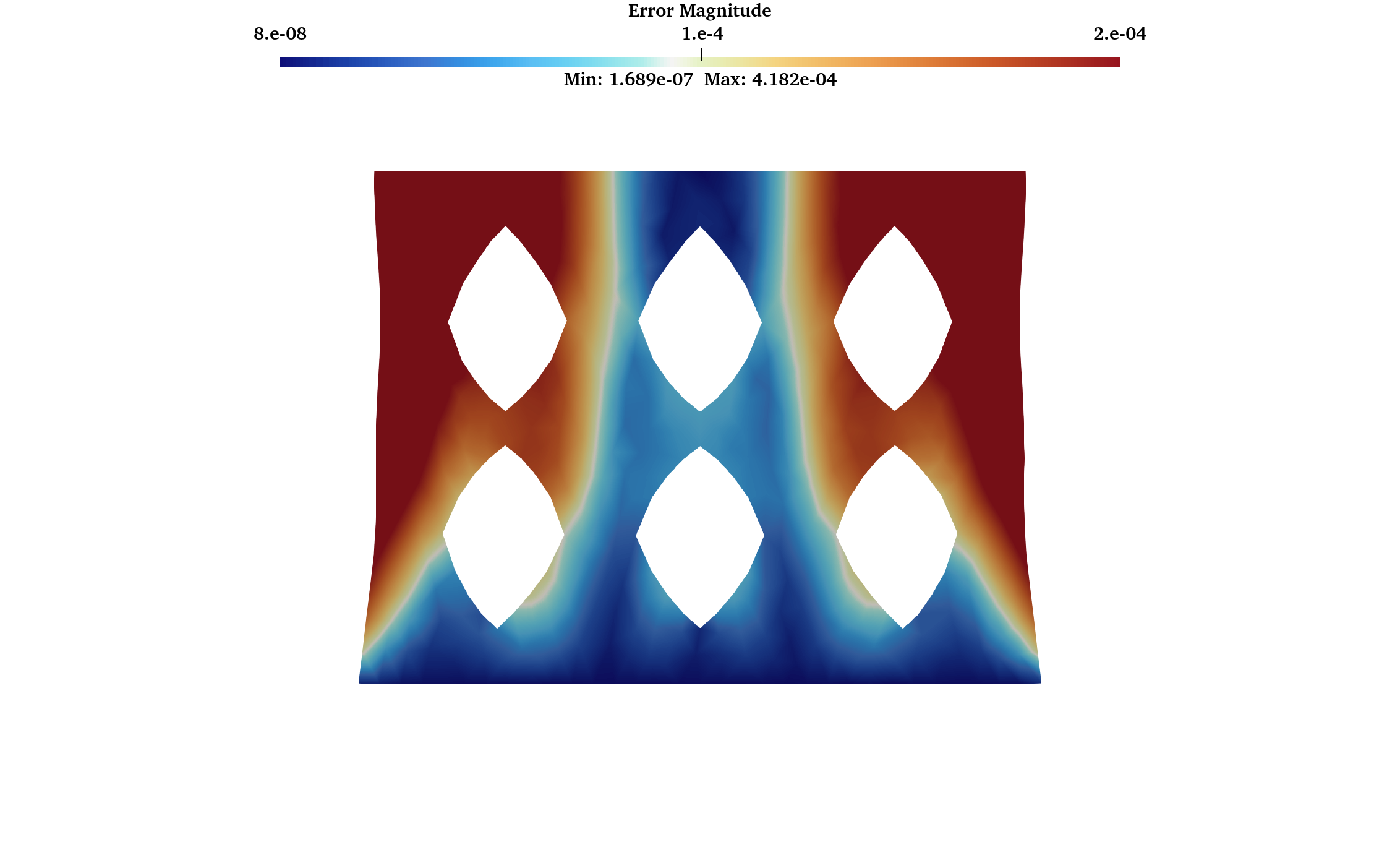}
    ~
    \includegraphics[width=0.32\linewidth, clip=true, trim = 140mm 50mm 140mm 0mm]{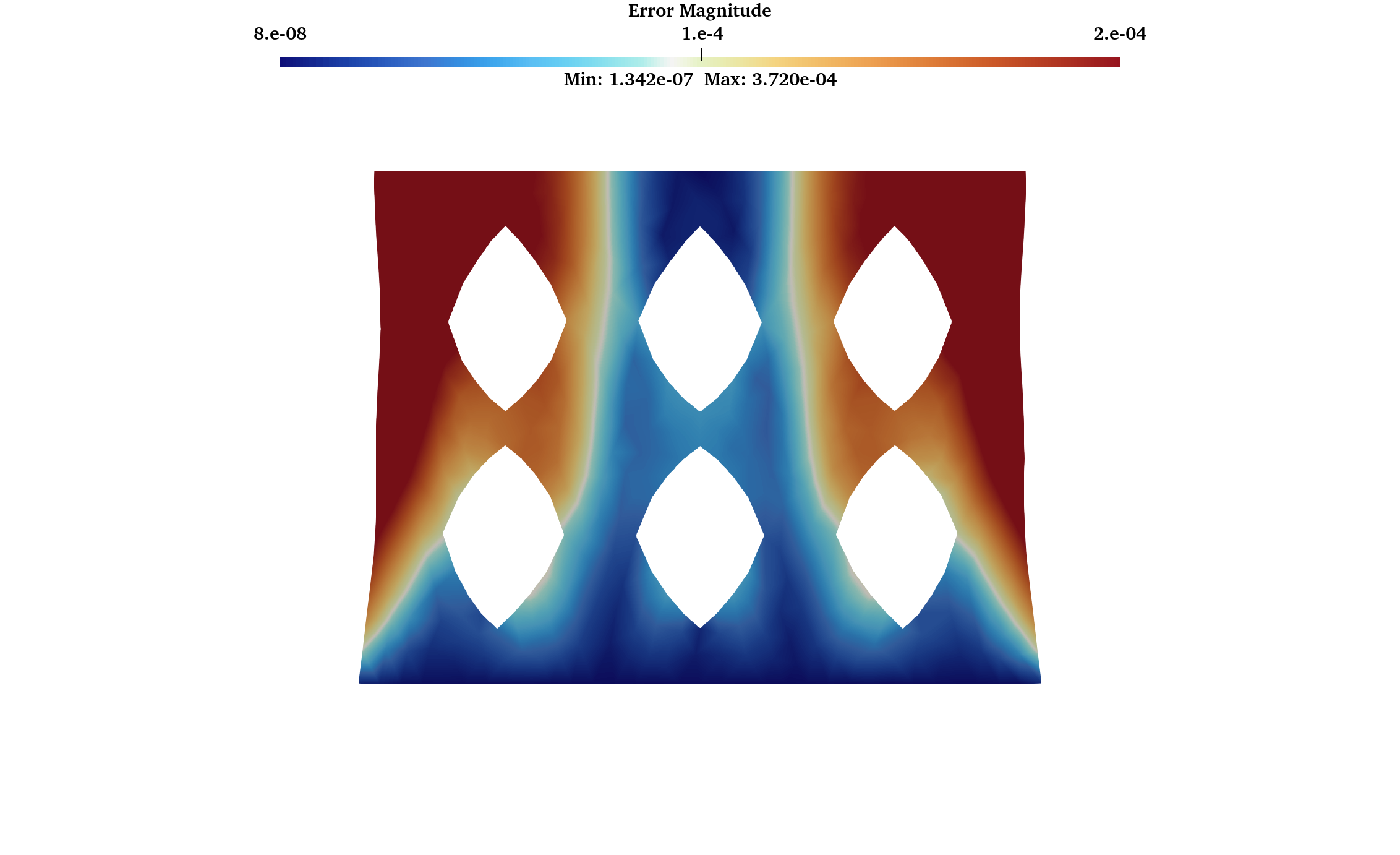}
    \caption{Transfer to Neo-Hookean Target, displacement error at $50\%$ applied strain}
    \label{fig:trasf_neo_displacement_error}
\end{figure}

Figure \ref{fig:trasf_neohooke} provides a deeper examination of the polyconvexity indicator inequalities by plotting the contours of \eqref{eq:polyconvex_I1} and \eqref{eq:polyconvex_I2} over the deformed finite element domain for all three models at $50\%$ elongation of the specimen. 
While all model satisfy the polyconvexity indicator $I_1$ inequality, both the relaxed and unconstrained NN variant show regions where the $I_2$ inaquality takes negative values which dictates that the polyconvexity requirements are violated. Nevertheless, none of the models had convergence issues that could be attributed to short wavelength instabilities for the chosen boundary value problem.  In Figure \ref{fig:trasf_neo_displacement_error}, the error of the final trained ICNN models compared to the Neo-Hookean ground truth is presented in a state of $50\%$ elongation of the specimen, showing that all ICNN variants have performed adequately.

\begin{figure}[h!]
    \centering
    \includegraphics[width=\linewidth]{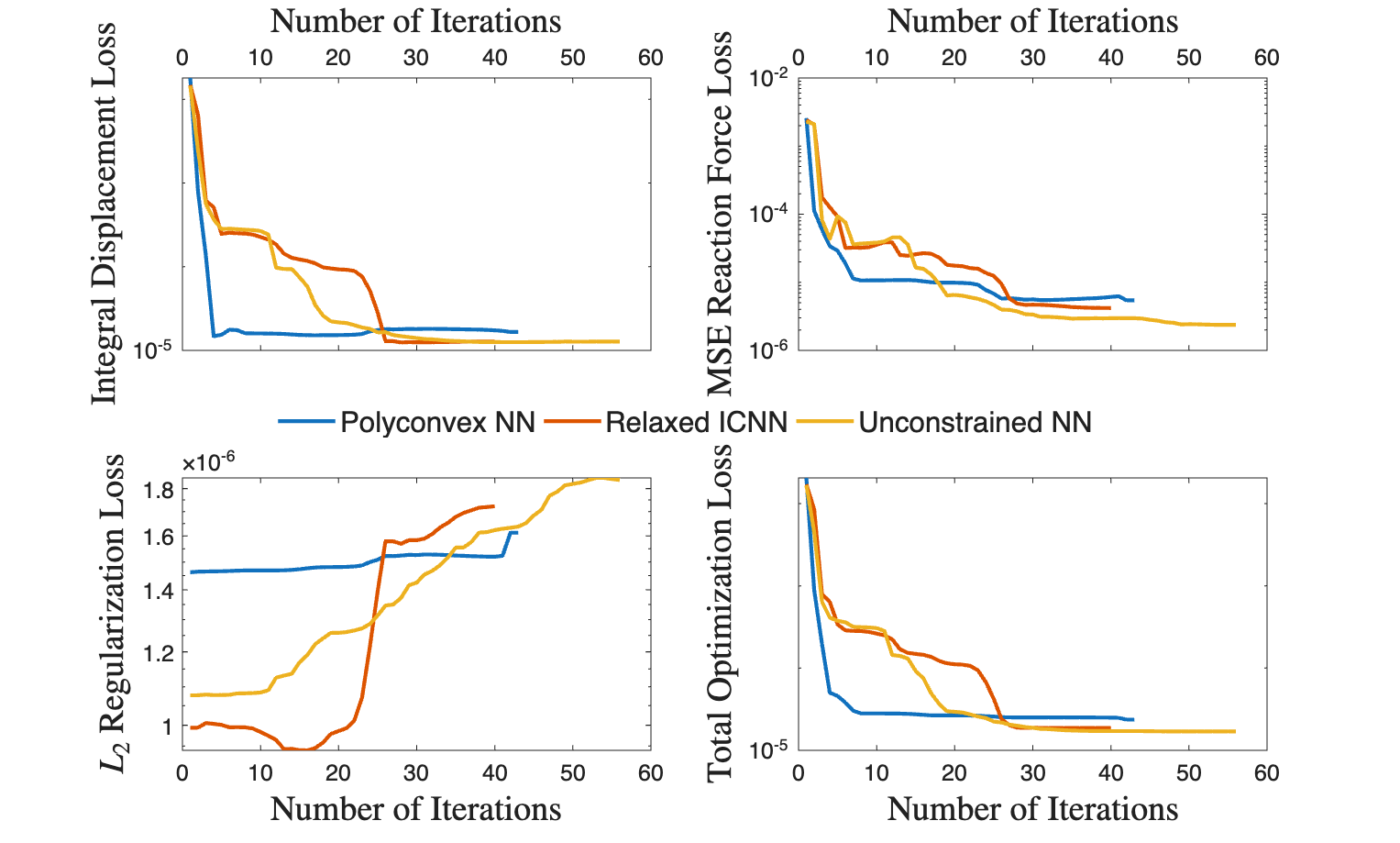}
    \caption{Loss components for transfer learning scheme, utilizing the Ogden synthetic DIC dataset}
    \label{fig:loss_gogden}
\end{figure}

\begin{figure}[h!]
    \centering
    \includegraphics[width=0.6\linewidth]{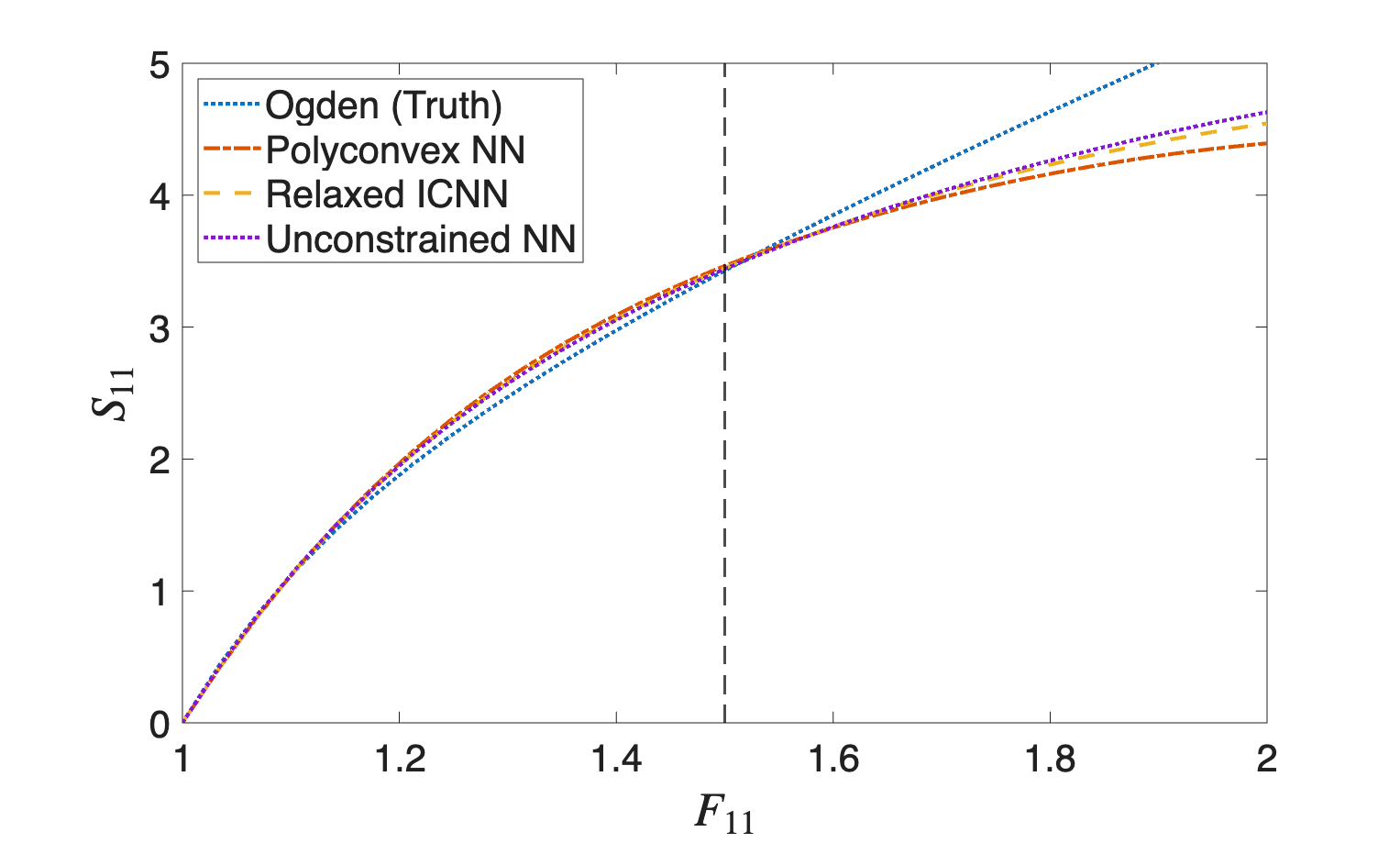}
    \caption{Transfer learning to Generalized Ogden DIC dataset: Validation of learned models to the analytical generalized Ogden model in a uniaxial tension setting via Newton-Raphson.}
    \label{fig:NR_gogden}
\end{figure}

\begin{figure}[h!]
    \centering
    \includegraphics[width=0.31\linewidth, clip=true, trim = 70mm 20mm 70mm 0mm]{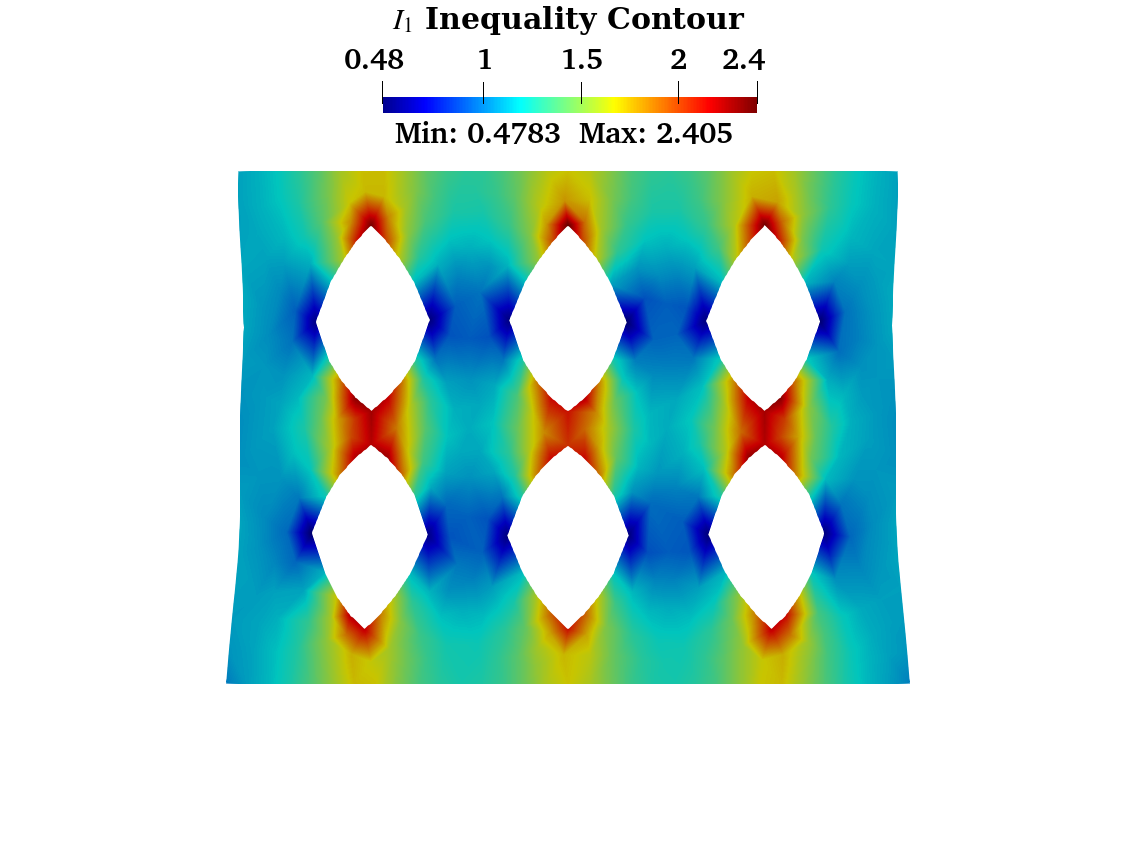}
    ~
    \includegraphics[width=0.31\linewidth, clip=true, trim = 70mm 20mm 70mm 0mm]{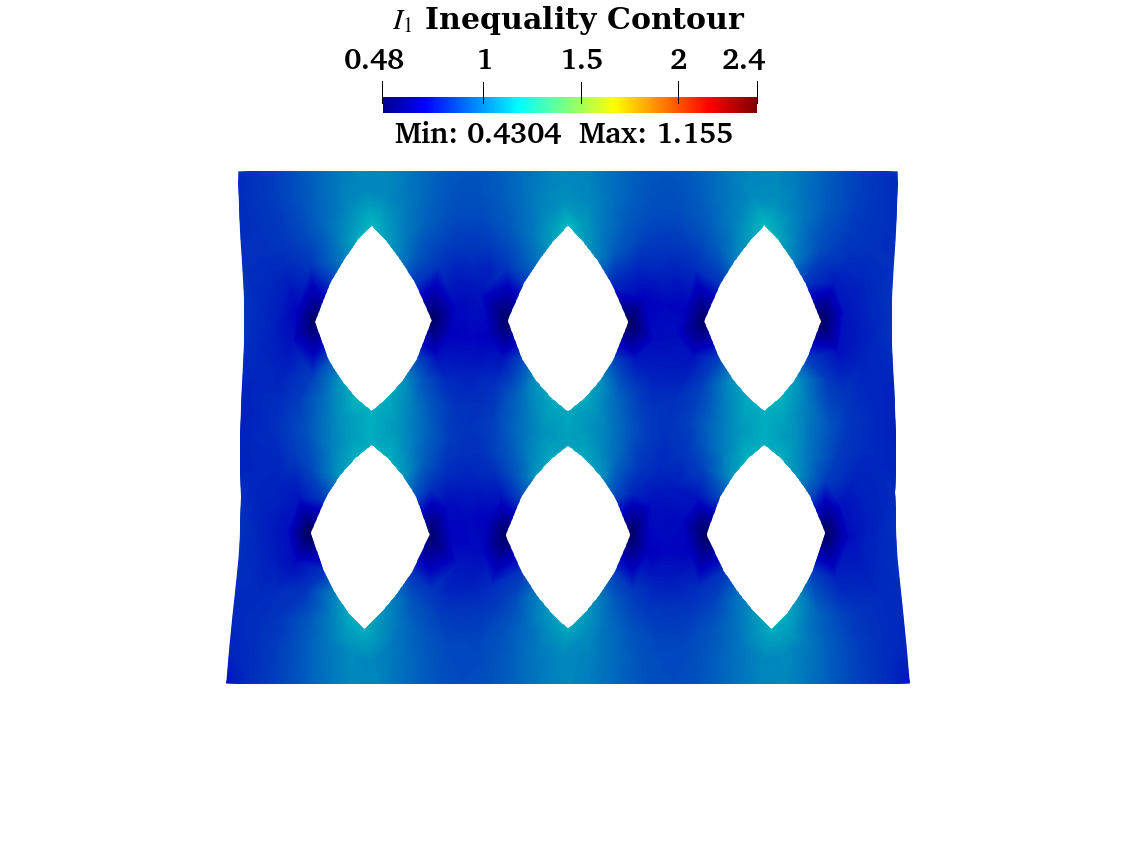}
    ~
    \includegraphics[width=0.31\linewidth, clip=true, trim = 70mm 20mm 70mm 0mm]{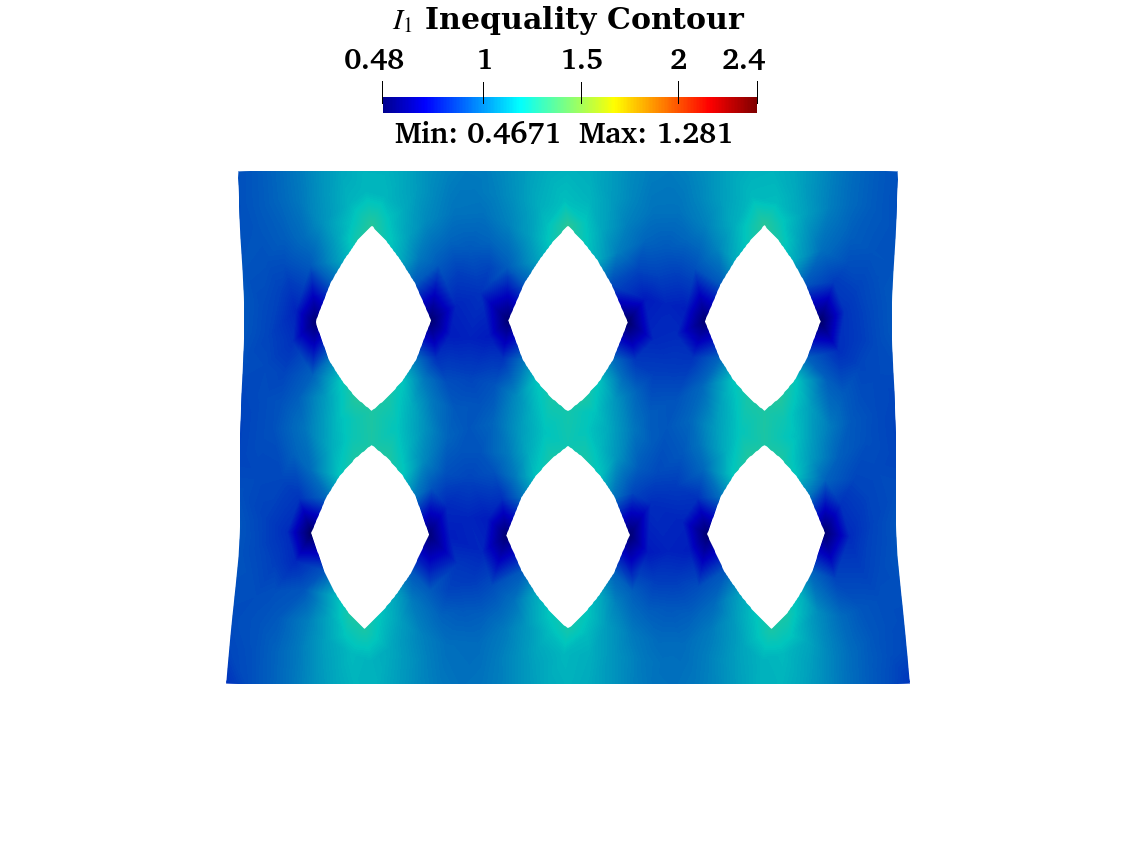}
    ~
    \includegraphics[width=0.31\linewidth, clip=true, trim = 70mm 20mm 70mm 0mm]{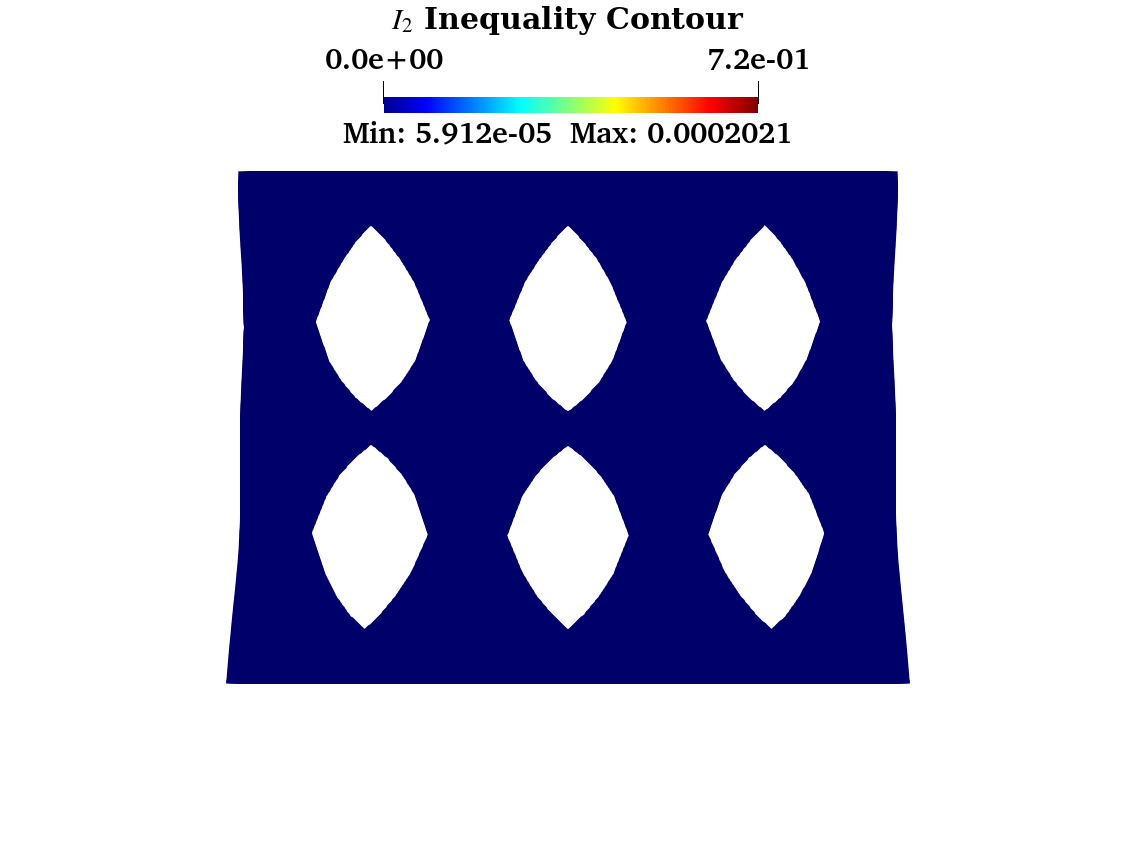}
    ~
    \includegraphics[width=0.31\linewidth, clip=true, trim = 70mm 20mm 70mm 0mm]{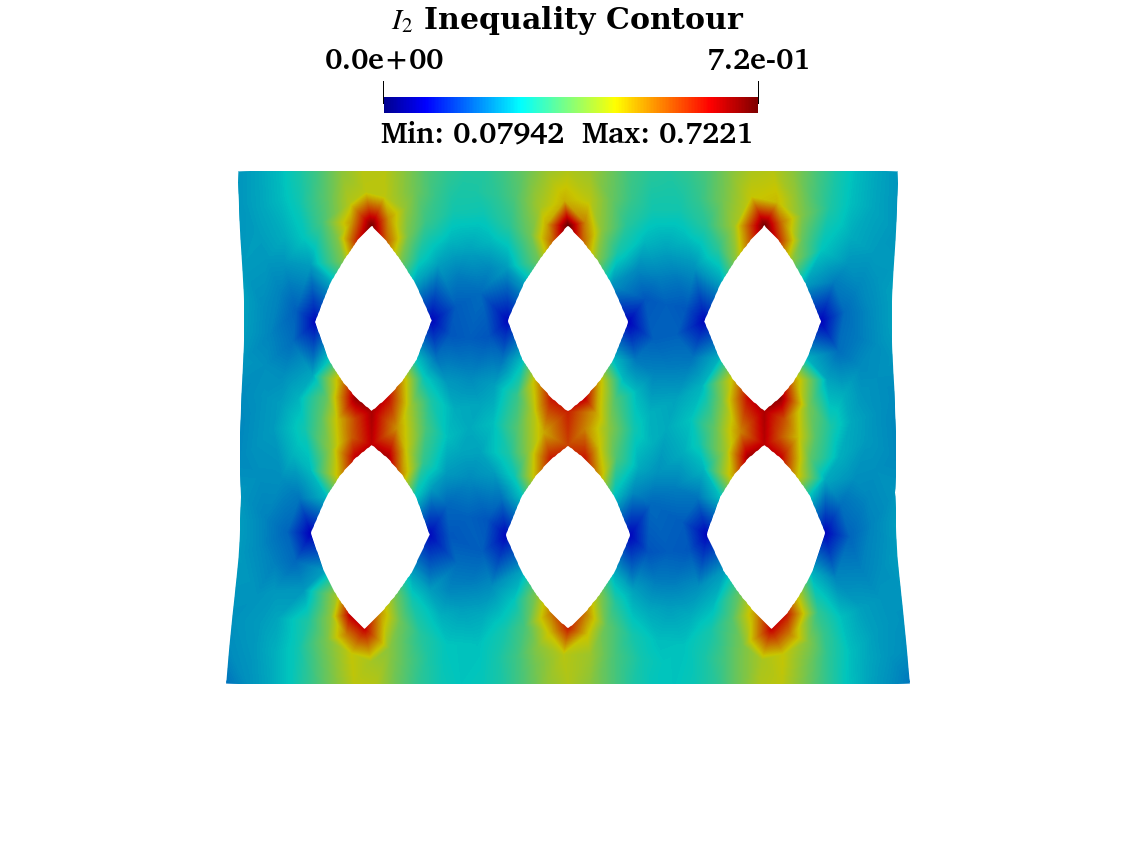}
    ~
    \includegraphics[width=0.31\linewidth, clip=true, trim = 70mm 20mm 70mm 0mm]{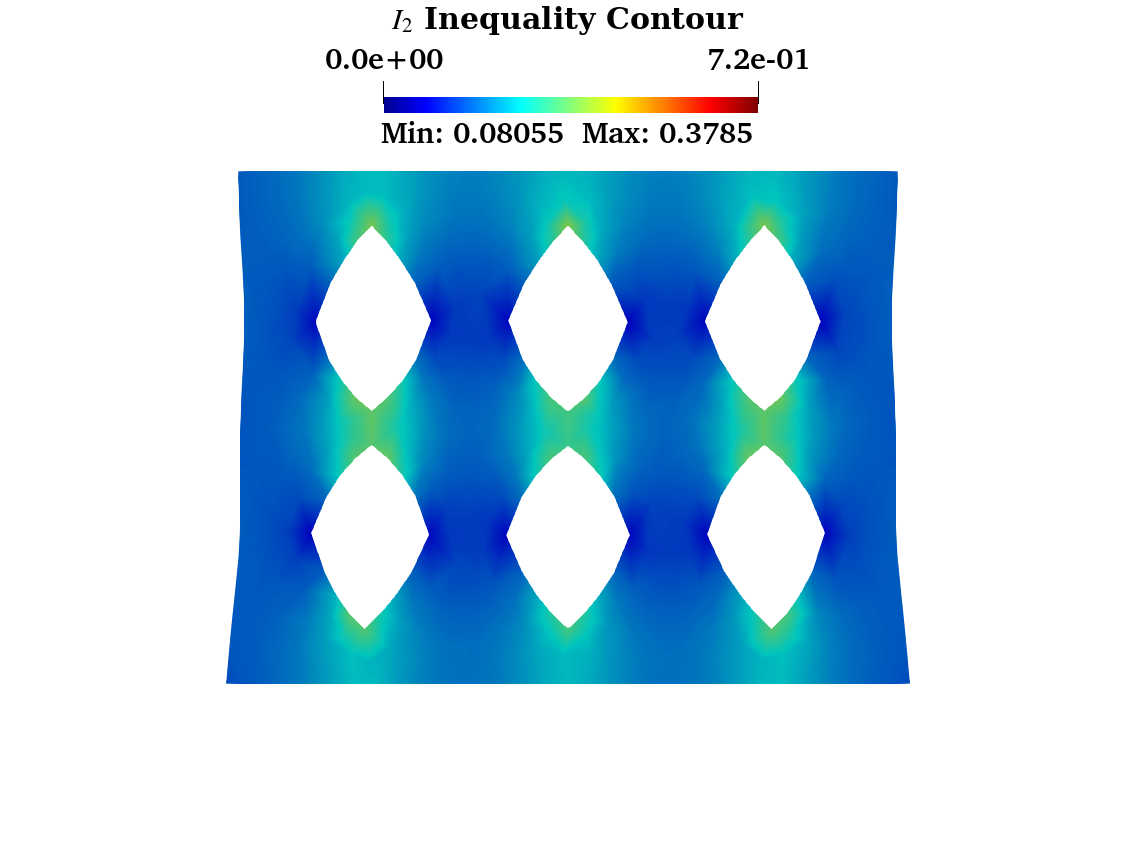}
    \caption{Polyconvexity indicator inequality evaluations, for polyconvex NN (left), reduced ICNN (middle), and unconstrained NN (right) models with the generalized Ogden target.}
    \label{fig:trasf_ogden}
\end{figure}

\begin{figure}[h!]
    \centering
    \includegraphics[width=0.31\linewidth, clip=true, trim = 70mm 20mm 70mm 0mm]{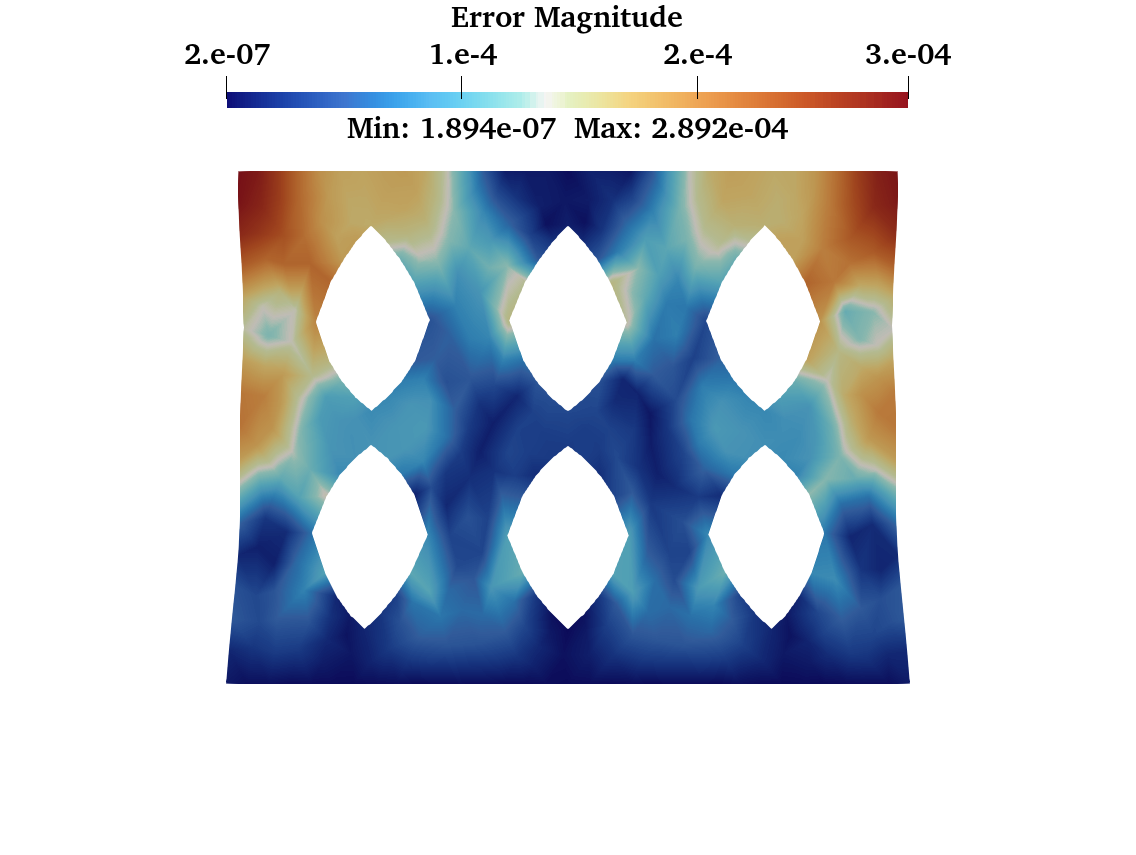}
    ~
    \includegraphics[width=0.31\linewidth, clip=true, trim = 70mm 20mm 70mm 0mm]{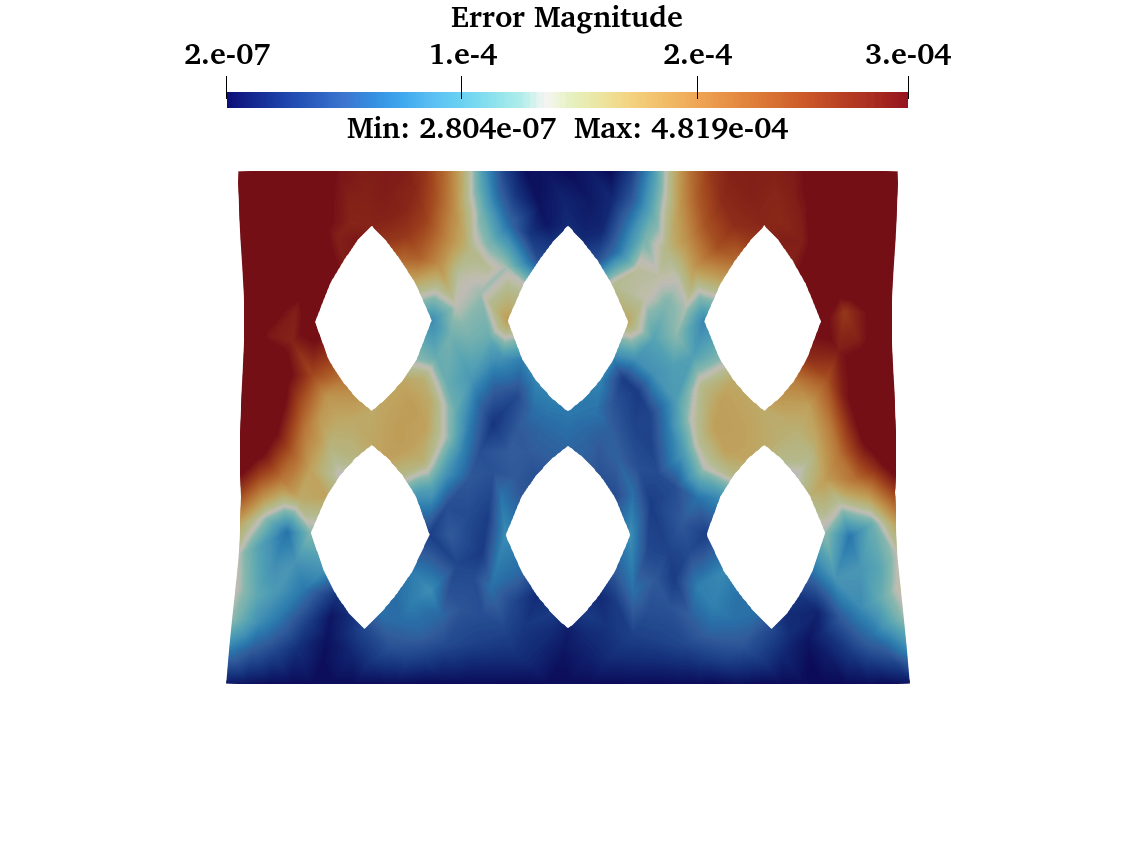}
    ~
    \includegraphics[width=0.31\linewidth, clip=true, trim = 70mm 20mm 70mm 0mm]{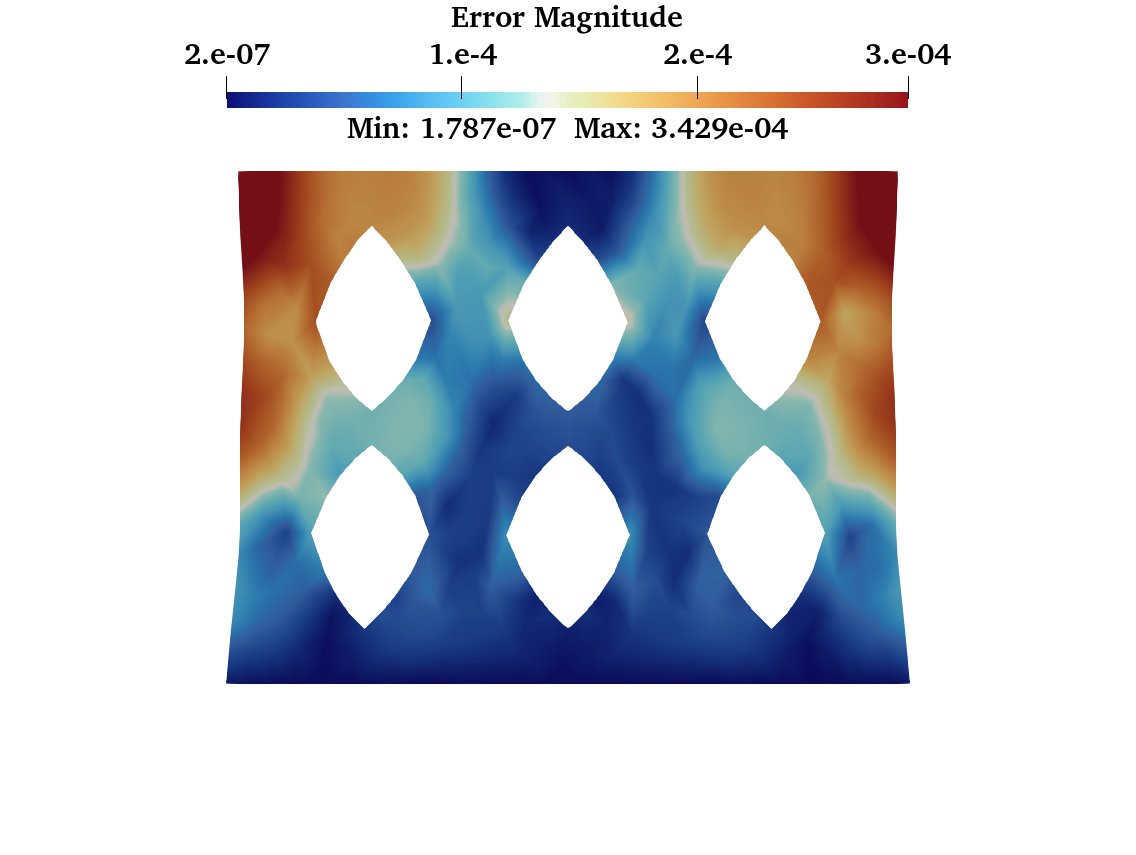}
    \caption{Transfer to Generalized Ogden, displacement error at $50\%$ applied strain}
    \label{fig:trasf_ogden_error}
\end{figure}


For the generalized Ogden case, Fig.\ref{fig:loss_gogden} shows that the optimization converged quickly for all three models, but the regularization loss did not further improve for any of the cases for the multi-objective optimization problem. It is noted that the optimization scheme reaches a plateau in slightly more than 20 iterations. The validation to the uniaxial tension test compared to the generalized Ogden model in Figure \ref{fig:NR_gogden}, showcases that in this case the model behaves sufficiently for strains that were predominantly observed in the DIC tests. This dictates the need for model augmentation strategies, as a means to improve expressivity and balance trustworthiness, in order to improve the error in generalization when the target material has a more complex response (in this case there was a deviatoric volumetric split that was present in the target but not part of the pre-training). The polyconvexity indicator inequality plots are shown for the generalized Ogden training in Figure \ref{fig:trasf_ogden}, with similar conclusion as for the Neo-Hookean transfer learning case discussed earlier. A visualization of the fit quality is given in Fig.\ref{fig:trasf_ogden_error}, which plots the spatial distribution of displacement error for the Ogden case at $50\%$ strain. 
The error magnitudes are very small on the order of a few percent of total displacement and are randomly distributed, with no systematic pattern of bias, indicating the model has captured both global and local behaviors well. 

\subsection{Deployment}
\begin{figure}[h!]
    \centering
    \includegraphics[width=\linewidth, clip=true, trim = 0mm 120mm 0mm 50mm]{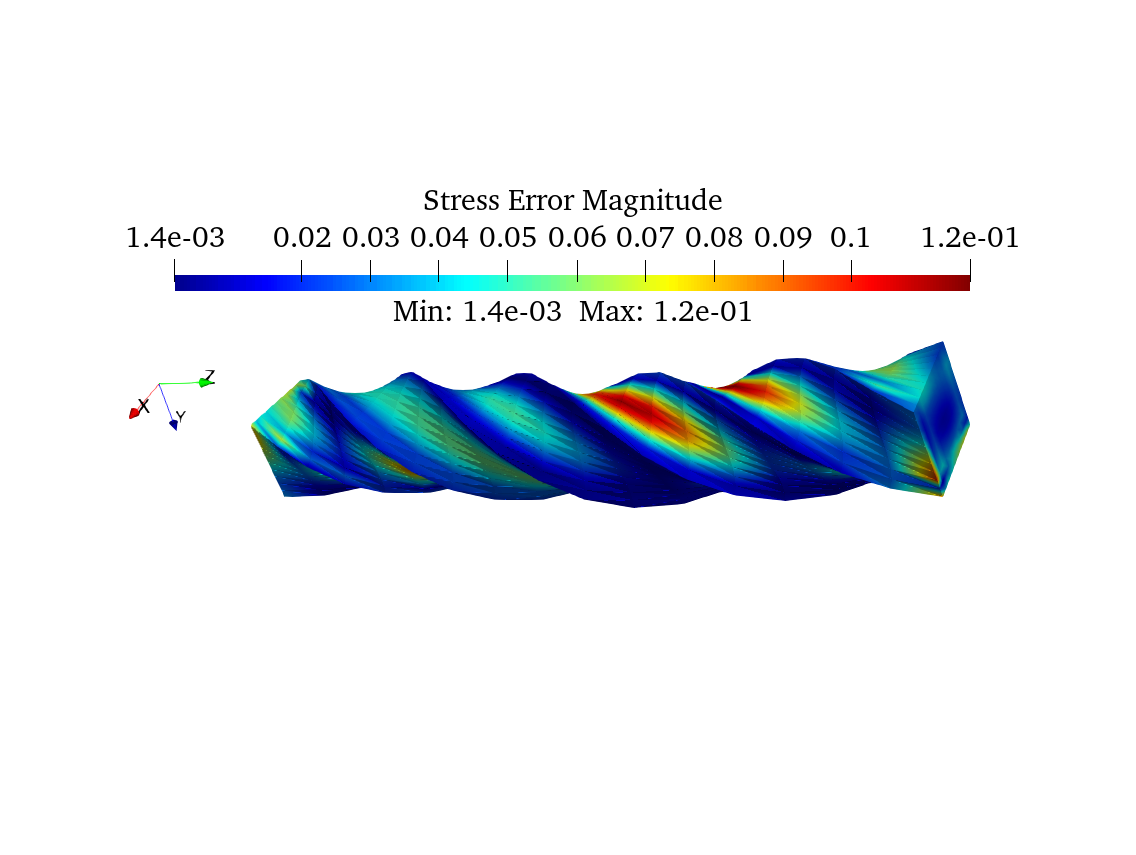}
    \caption{Stress error in the deployment of the transferred model in 3D torsion simulation with a maximum twist angle of $458^\circ$ compared to ground truth. The overall relative error at this state is $8.6\%$}
    \label{fig:twist}
\end{figure}

Finally,  we deploy the transferred constitutive model in a significantly more complex loading case to assess its predictive power,assuming no additional observation or re-calibration is available. 
This stage serves as the ultimate trustworthiness test: if the model can accurately predict outcomes in a complex unseen scenario with potentially many under-informed  modes,
it demonstrates that the pre-training to transfer learning process has yielded a truly reliable predictive tool. 
For this deployment, we selected a 3D torsional deformation as the deployment challenger case. 
Specifically, a rectangular columnar specimen with a square cross-section is subjected to a prescribed twist about the $z$-axis incrementally up to an extreme angular displacement of 458$^\circ$ on one end, while the other end is fixed and total elongation along the $z$-axis is prohibited. 
The material model for this simulation is the final trained  relaxed ICNN model with the Neo-Hookean target from Sec.\ref{subsec:transfer}, with no additional fitting or adjustment tailored for torsion. 
We then compared the model prediction in stress to that of the ground truth simulation of the same torsion scenario with the analytical ....model, resulting to the contour of absolute error shown in Fig.\ref{fig:twist}. 
The results of this deployment are very encouraging with the neural network model remaining stable and not showcasing any numerical issues for the large deformation twist. Ultimately the model was able to predict the stress distribution in the twisted cylinder with good accuracy relative to the ground truth. 
An overall relative error measure integrating the stress error over the entire 3D volume at this deformed state shows only $8.6\%$ relative error in the predicted stress, with maximum error as specific locations reaching $12\%$. It is worth noting that the maximum values of the magnitude of the deformation gradient in this test is $|\mathbf{F}|_\mathrm{max}=1.97$, far beyond the training domain.
%
%

\section{Conclusion}
This work introduces paFEMU, a transfer learning scheme to enable interpretable discovery of constitutive laws from multi-modal datasets. The key innovation lies in combining sparsified neural representations with differentiable finite element solvers to enable low-dimensional, physics-consistent transfer learning. In the pre-training stage, NN-based models are trained directly on labeled data pairs connecting the deformation state and the stress state through a learned potential. Three levels of enforcing and biasing the polyconvexity requirement are discussed: polyconvex ICNNs, relaxed ICNNs and unconstrained NNs, and the concept of a polyconvexity indicator that can be easily evaluated is introduced. The final training stage further refines the low-dimensional discovered expressions for the constitutive law, on a new material target utilizing an FE-based adjoint for PDE-constrained optimiation. The target data corresponds to full-field displacement and reaction force information, for a synthetic DIC test. Sparsification is key to completing this strategy in a trusted FE setting, without the burden of over-parameterized neural representations. This approach is designed to manage expressivity and trustworthiness, enabling the use of pre-trained models corresponding to materials in the same or adjacent material classes. Unlike prior approaches that either focus on purely data-driven discovery or calibration within fixed model classes, the proposed framework unifies model discovery, sparsification, and adjoint-based updating within a end-to-end. A central outcome of this formulation is the emergence of sparse, interpretable constitutive models that retain the expressivity of neural networks while enabling efficient integration within classical finite element workflows.

The transfer learning stage validates that our extremely-sparsified ICNN models retain sufficient expressiveness to learn new complex behaviors from full-field data. 
While all three ICNN variants could be transferred to the new materials, we observed some differences in their calibration efficiency and fidelity. 
The Unconstrained NN consistently achieved the lowest final error on the DIC data and required the fewest iterations to converge. This is expected, as its lack of restrictions allows it to reshape freely to the target behavior. 
The Polyconvex ICNN, in contrast, sometimes converged to a slightly higher misfit, especially for the generalized Ogden case 
However, it is important to note that the polyconvex model still captured the major trends of the data and remained physically plausible throughout and never violating material stability. 
The constrained ICNN proved to be a very effective compromise, it achieved almost the same accuracy as the unconstrained model in both cases, while still biased to not violate polyconvexity.

 Future work will focus on extending the framework to history-dependent and path-dependent material behavior, including plasticity and viscoelasticity, as well as integrating active experimental design to optimally select loading paths that maximize identifiability. Additional directions include scaling the approach to heterogeneous materials and multi-physics settings, and exploring real-time updating in closed-loop experimental systems.

Overall, this work opens the door to rapid material characterization in data-scarce regimes, where limited experimental measurements can be leveraged through transfer learning to construct reliable and physically admissible constitutive models.


\section{Acknowledgement}
Tha authors would like to thank D. Thomas Seidl from Sandia National Laboratories for the helpful discussions and suggestions regarding adjoints and DIC. J.T. and N.B. were supported by the Cornell SciAI Center, and funded by the Office of Naval Re-
search (ONR), under Grant No. N00014-23-1-2729. 
\appendix
\numberwithin{figure}{section}
\numberwithin{equation}{section}
\section{Appendix}
\subsection{Simple Deformation Modes}
To verify and illustrate constitutive behavior, one often examines canonical homogeneous deformations. We consider three basic modes – uniaxial stretch, equibiaxial stretch, and simple shear – under idealized constraints, plus a case of free lateral contraction. These allow deriving analytical stress–strain relations from a given $\varphi(\mathbf{F})$ and checking consistency (or providing data fits). All deformations below are taken with reference to an orthonormal basis and (for simplicity) assumed to be aligned with the principal material axes of an isotropic hyperelastic solid.

\subsubsection{Constrained uniaxial tension/compression}
Here the material is stretched by a factor $\lambda$ in one direction ($x_1$), while the lateral directions are held fixed (no strain in $x_2,x_3$). The deformation gradient is $\mathbf{F}=\mathrm{diag}(\lambda,\,1,\,1)$. Because the cross-section is not allowed to contract or expand, this is a plane-strain uniaxial test, and the stress response can be obtained from the energy by $S = \partial \varphi/\partial \mathbf{C}$. 
In an isotropic material, symmetry implies $S_{22}=S_{33}$ in this scenario. Generally, one finds a nonzero lateral stress developing: $S_{22}=S_{33}\neq 0$ because the material wants to contract laterally but is constrained. The axial stress $S_{11}$ increases with $\lambda$ according to the model’s specific form. 
Constrained uniaxial tests are useful for assessing the model’s predicted Poisson effect and verifying stress symmetry (here $S_{22}=S_{33}$) and energy consistency.

\subsubsection{Constrained equi-biaxial tension/compression}

In this mode, the sample is stretched by the same factor in two orthogonal directions while the thickness is held fixed. For instance, $\lambda_1=\lambda_2=\lambda$ and $\lambda_3=1$, so $\mathbf{F}=\mathrm{diag}(\lambda,\,\lambda,\,1)$. 
This could model a sheet stretched in two directions while preventing any thickness change.
%
%
The in-plane stresses $S_{11}=S_{22}$ will be tensile for tension loads, and a normal stress $S_{33}$ generally arises due to the constraint on thickness. 
If the material is incompressible, the constraint $\det F=1$ would actually require $\lambda_3=(\lambda^2)^{-1}=\lambda^{-2}$ (the thickness must contract when stretching in-plane). 
But here we consider the constrained case $F_{33}=1$, so incompressibility is violated and a positive $S_{33}$ develops to resist volume change. 
Equi-biaxial loading is a stringent test of the model's volumetric response and its strain-hardening characteristics. For example, the Gent model under equi-biaxial strain gives a stress–stretch curve that highlights the limiting chain extensibility as $\lambda$ approaches the limit (where $(I_1-3) \to J_m$)

\subsubsection{Simple shear}

This deformation gradient is 

\begin{equation}
    \mathbf{F} 
    = 
    \begin{bmatrix}
       1 & \gamma & 0 \\
       0 & 1 & 0 \\
       0 & 0 & 1
    \end{bmatrix},
\end{equation}
representing a shear of amount $\gamma$ in the $x_1$–$x_2$ plane with no change in lengths along axes. 
It is a volume-preserving isochoric deformation ($\det \mathbf{F}=1$) often used to probe shear response. 
%
%
The shear stress $S_{12}$ (or $\sigma_{12}$ in Cauchy form) can be derived from $\varphi$: 
for small $\gamma$, one expects $S_{12}\approx \mu\,\gamma$ where $\mu$ is the shear modulus;
however, simple shear is not purely simple – a nonzero normal stress can occur (the Poynting effect), meaning $S_{11}\neq S_{22}$ in general. In fact, many hyperelastic models predict that a material in simple shear will develop normal stress differences $S_{11}-S_{22}$ proportional to $\gamma^2$. 
For instance, neo-Hookean and Mooney–Rivlin materials both exhibit a tensile normal stress in the direction of shear for $\gamma>0$.
Experimentally, this effect is observed as a normal force pushing the shear plates apart. 
By analyzing simple shear, one checks the model's ability to capture such asymmetry. Stress symmetry is still maintained through symmetric stress in the shear plane ($S_{12}=S_{21}$), while generally $S_{11}\neq S_{22}$.
Simple shear tests thus reveal nonlinear shear behavior and are often used to fit the $I_2$-dependent terms in a model.

\subsubsection{Traction-free uniaxial tension/compression}

This case corresponds to standard uniaxial tension and compression tests where the material can freely contract laterally with no lateral forces/constraints applied. 
Here one prescribes an axial stretch $\lambda$ in $x_1$ and requires the lateral normal stresses to vanish ($S_{22}=S_{33}=0$). 
The deformation is not known apriori in $x_2,x_3$, instead, lateral stretches $\lambda_2,\lambda_3$ must be solved for such that the equilibrium condition (zero lateral stress) is satisfied. 
Incompressible behavior further dictates $\lambda_2=\lambda_3=\lambda^{-1/2}$. In the general compressible case, one finds $0<\lambda_2=\lambda_3<1$ for $\lambda>1$, meaning the specimen contracts laterally. 
The specific value comes from solving $S_{22}(\lambda,\lambda_2,\lambda_2)= 0$ given the specific form of $\varphi$.
In general, a nonlinear equation stemming from 
\begin{equation}
    S_{22} 
    = 
    2\frac{\partial\varphi}{\partial I_1}
    +
    2(\lambda^2 +  \lambda_2^2)\frac{\partial\varphi}{\partial I_2}
    +
    \lambda
    \frac{\partial\varphi}{\partial J}
    =0
\end{equation}
to find the equilibrium lateral stretch $\lambda_2$. 
If a differentiable form of $\varphi$ is available, the Newton–Raphson method is commonly employed to solve the nonlinear equation using the derivative (Jacobian) of $S_{ss}$.
The result is then used to compute the axial stress under true uniaxial conditions:
\begin{equation}
    S_{11}(\lambda)
    = 
    2\frac{\partial\varphi}{\partial I_1}
    +
    4\lambda_2^2
    \frac{\partial\varphi}{\partial I_2}
    +
    \frac{\lambda_2^2}{\lambda}
    \frac{\partial\varphi}{\partial J}
\end{equation}
This scenario highlights that, unlike the constrained tests, here the material’s natural Poisson contraction is realized, and no lateral reactive stress is present. It provides a consistency check between the predicted Poisson effect and the measured lateral contraction in experiments. Additionally, it probes the convexity and numerical stability of the chosen model $\varphi$.

\subsection{Displacement-Based Boundary Value Problems}
The above constitutive framework can be further incorporated into a displacement-based finite element method (FEM) to solve boundary value problems with high geometric complexity in hyperelasticity. We outline the standard formulation, starting from the strong form and proceeding to the numerical solution.
\subsubsection*{Strong form via momentum balance}
In the reference configuration $X\in\Omega_0$, the equilibrium equations are
\begin{equation}
    \nabla_X \cdot \mathbf{P}(X) + \mathbf{B}(X) = \mathbf{0} \quad \forall X\in\Omega_0,
\end{equation}
where $\mathbf{P}$ is the first Piola–Kirchhoff stress computed from a chosen constitutive model $\varphi$ through the differential relation in \eqref{eq:stress_from_phi} ; and $\mathbf{B}$ a body force (we omit inertia for static problems). 
In absence of body forces this simplifies to $\nabla\cdot \mathbf{P}=0$ in $\Omega_0$. 
Boundary conditions are prescribed as follow: 
displacements $\mathbf{u} = \mathbf{u}^\dagger$ on a portion of the boundary $\partial\Omega_0^u$ 
and 
traction $\mathbf{P}\mathbf{N}=\bar{\mathbf{T}}$ 
on the remainder 
$\partial\Omega_0^t$, with $\mathbf{N}$ the outward normal in the reference configuration.
\subsubsection{Weak form via principle of virtual work}
To derive the weak form, one multiplies the equilibrium equation by a virtual displacement field (test function) $\delta \mathbf{u}(X)$ and integrates over $\Omega_0$. Integration by parts (assuming sufficient smoothness) yields the virtual work principle: 
\begin{equation}
    \int_{\Omega_0} \mathbf{P} : \nabla_X(\delta \mathbf{u})\,\mathrm{d}V = \int_{\Omega_0} \mathbf{B}\cdot \delta \mathbf{u}\,\mathrm{d}V + \int_{\partial\Omega_0^t}\bar{\mathbf{T}}\cdot \delta \mathbf{u}\,\mathrm{d}A,
\end{equation}
for all admissible $\delta\mathbf{u}$. 
This weak form enforces equilibrium in an integral, weighted-average sense and naturally incorporates traction boundary conditions (via the surface integral). It is equivalent to the strong form for sufficiently smooth $\mathbf{P}$, and is the starting point for finite element discretization.
In hyperelasticity, $\mathbf{P}$ is derived from a strain-energy
as in \eqref{eq:stress_from_phi}, or often one works with the second Piola–Kirchhoff stress $\mathbf{S}=\frac{\partial \varphi}{\partial \boldsymbol{\varepsilon}}$ with the Green–Lagrange strain tensor $\boldsymbol{\varepsilon}$ and its conjugates.

\subsection{Hybrid farthest-point sampling and simulated annealing}\label{Appdx:FRS+SA}

\begin{algorithm}[H]
\caption{Hybrid farthest-point sampling and simulated annealing for triplet selection from a convex hull}
\label{alg:hybrid_fps_sa_hull}

\KwIn{Hull point set $\mathcal{X}\subset\mathbb{R}^3$, target cardinality $k$, weights $\alpha,\beta$, annealing parameters $(T_0,\gamma,I_{\max},I_{\mathrm{stall}})$, candidate-pool size $M_{\mathrm{cand}}$, swap attempts $n_{\mathrm{swap}}$, optional anchor}
\KwOut{Selected point set $\mathcal{X}_{\mathrm{sel}}$}

$\widetilde{\mathcal{X}} \gets (\mathcal{X}-\mu_{\mathcal{X}})./\sigma_{\mathcal{X}}$\;
$\mathcal{S}_{\mathrm{curr}} \gets \arg\max_{\mathcal{S}\in\{\mathrm{FPS\ runs}\}} d_{\min}(\mathcal{S})$\;
$\mathcal{S}_{\star}\gets \mathcal{S}_{\mathrm{curr}},\quad T\gets T_0,\quad i_{\star}\gets 0$\;

$J(\mathcal{S})=\alpha\, d_{\min}(\mathcal{S})+\beta\,\overline{d}_{\mathrm{NN}}(\mathcal{S})$\;

\For{$i=1$ \KwTo $I_{\max}$}{
    $r(x)=\min\limits_{s\in\mathcal{S}_{\mathrm{curr}}}\|x-s\|_2^2,\qquad x\in\widetilde{\mathcal{X}}\setminus\mathcal{S}_{\mathrm{curr}}$\;

    $\mathcal{C}\subset \widetilde{\mathcal{X}}\setminus\mathcal{S}_{\mathrm{curr}},\quad |\mathcal{C}|\le M_{\mathrm{cand}},\quad \Pr(x\in\mathcal{C})\propto r(x)$\;

    \For{$t=1$ \KwTo $n_{\mathrm{swap}}$}{
        $p\sim\mathcal{S}_{\mathrm{curr}},\quad q\sim\mathcal{C}$\;
        $\mathcal{S}_{\mathrm{prop}}=(\mathcal{S}_{\mathrm{curr}}\setminus\{p\})\cup\{q\}$\;
        $\Delta J = J(\mathcal{S}_{\mathrm{prop}})-J(\mathcal{S}_{\mathrm{curr}})$\;

        \If{$\Delta J\ge 0$ \textbf{or} $\mathrm{rand}(0,1)<\exp(\Delta J/T)$}{
            $\mathcal{S}_{\mathrm{curr}}\gets \mathcal{S}_{\mathrm{prop}}$\;
            \If{$J(\mathcal{S}_{\mathrm{curr}})>J(\mathcal{S}_{\star})$}{
                $\mathcal{S}_{\star}\gets \mathcal{S}_{\mathrm{curr}},\quad i_{\star}\gets i$\;
            }
            \textbf{break}\;
        }
    }

    $T\gets \gamma T$\;

    \If{$i\equiv 0 \pmod{500}$}{
        $\mathcal{S}_{\star}\gets \mathrm{Rescue}(\mathcal{S}_{\star})$\;
{\footnotesize$\triangleright$ targeted 1-swap updates to break closest pair(s), accept only if $d_{\min}$ increases\;}
    }

    \If{$i\equiv 0 \pmod{5000}$}{
        $\mathcal{S}_{\star}\gets \mathrm{HillClimb}(\mathcal{S}_{\star})$\;
{\footnotesize$\triangleright$ deterministic 1-swap ascent on $d_{\min}$\;}
    }

    \If{$i-i_{\star}>I_{\mathrm{stall}}$}{
        \textbf{break}\;
    }
}

$\mathcal{S}_{\star}\gets \mathrm{HillClimb}(\mathcal{S}_{\star})$\;
{\footnotesize$\triangleright$ deterministic 1-swap ascent on $d_{\min}$\;}
$\mathcal{X}_{\mathrm{sel}}=\mathcal{X}(\mathcal{S}_{\star},:)$\;
\end{algorithm}

\subsection{Reconstruction of diagonal right Cauchy-Green tensor from invariants}\label{appdx:rec_C}

\begin{equation}
    \begin{cases}
        H=\frac{1}{9}\left(I_1^2-3I_2\right)
        \\
        G=\frac{1}{3}I_1 I_2 - I_3 - \frac{2}{27}I_1^3
        \\
        \beta=\arccos\left(-\frac{G}{2H^{3/2}}\right)
    \end{cases}
\end{equation}

\begin{equation}
\lambda_{1,\mathrm{rec}}^{2}
=
\frac{1}{3}I_{1}
-2\sqrt{H}\cos\left(\frac{\pi-\beta}{3}\right)
\end{equation}

\begin{equation}
\lambda_{2,\mathrm{rec}}^{2}
=
\frac{1}{3}I_{1}
-2\sqrt{H}\cos\left(\frac{\pi+\beta}{3}\right)
\end{equation}

\begin{equation}
\lambda_{3,\mathrm{rec}}^{2}
=
\frac{1}{3}I_{1}
+2\sqrt{H}\cos\left(\frac{\beta}{3}\right)
\end{equation}

\begin{equation}
\mathbf{C}_{\mathrm{rec}}=
\begin{bmatrix}
\lambda_{1,\mathrm{rec}}^{2} & 0 & 0 \\
0 & \lambda_{2,\mathrm{rec}}^{2} & 0 \\
0 & 0 & \lambda_{3,\mathrm{rec}}^{2}
\end{bmatrix}
\end{equation}

\subsection{Model parameters in full precision}\label{appdx:params}
\begin{table}[h!]
  \centering
  \caption{Pretrained model parameter sets (full precision).}
  \label{tab:pretrained_full}
  \begin{tabular}{@{}lr lr lr@{}}
    \toprule
    \multicolumn{2}{c}{Polyconvex NN} &
    \multicolumn{2}{c}{Relaxed ICNN} &
    \multicolumn{2}{c}{Unconstrained NN} \\
    \cmidrule(lr){1-2}\cmidrule(lr){3-4}\cmidrule(lr){5-6}
    \textbf{Param} & \textbf{Value} &
    \textbf{Param} & \textbf{Value} &
    \textbf{Param} & \textbf{Value} \\
    \midrule
    $\theta_{1}$  & 0.998645544052124 & $\theta_{1}$ & 0.905308246612549 & $\theta_{1}$ & 0.782256841659546 \\
    $\theta_{2}$  & 5.22694253921509  & $\theta_{2}$ & 0.506476998329163 & $\theta_{2}$ & 0.694582641124725 \\
    $\theta_{3}$  & -0.695928037166595 & $\theta_{3}$ & 1.67390620708466 & $\theta_{3}$ & 3.12386536598206 \\
    $\theta_{4}$  & 0.149173066020012 & $\theta_{4}$ & -0.210611954331398 & $\theta_{4}$ & -0.192448511719704 \\
    $\theta_{5}$  & 1.57681846618652  & $\theta_{5}$ & 1.42899298667908 & $\theta_{5}$ & 1.51982772350311 \\
    $\theta_{6}$  & -0.584444403648376 & $\theta_{6}$ & 1.26648092269897 & $\theta_{6}$ & 1.28323090076447 \\
    $\theta_{7}$  & 0.0798970237374306 & $\theta_{7}$ & -0.772105455398560 & $\theta_{7}$ & -0.803252279758453 \\
    $\theta_{8}$  & 1.61656022071838  & $\theta_{8}$ & 3.65706539154053 & $\theta_{8}$ & 2.79724001884460 \\
    $\theta_{9}$  & -3.23169875144958 & $\theta_{9}$ & -0.521339654922485 & $\theta_{9}$ & -0.574892044067383 \\
    $\theta_{10}$ & 1.27855789661407  & \multicolumn{2}{c}{} & \multicolumn{2}{c}{} \\
    $\theta_{11}$ & 0.0287286546081305 & \multicolumn{2}{c}{} & \multicolumn{2}{c}{} \\
    $\theta_{12}$ & 0.0713559985160828 & \multicolumn{2}{c}{} & \multicolumn{2}{c}{} \\
    $\theta_{13}$ & 0.0418042466044426 & \multicolumn{2}{c}{} & \multicolumn{2}{c}{} \\
    \bottomrule
  \end{tabular}
\end{table}
\begin{table}[h!]
  \centering
  \caption{Calibrated model parameter sets for model transfer to NeoHookean material (full precision).}
  \label{tab:neohooke}
  \begin{tabular}{@{}lr lr lr@{}}
    \toprule
    \multicolumn{2}{c}{Polyconvex NN} &
    \multicolumn{2}{c}{Relaxed ICNN} &
    \multicolumn{2}{c}{Unconstrained NN} \\
    \cmidrule(lr){1-2}\cmidrule(lr){3-4}\cmidrule(lr){5-6}
    \textbf{Param} & \textbf{Value} &
    \textbf{Param} & \textbf{Value} &
    \textbf{Param} & \textbf{Value} \\
    \midrule
    $\theta_{1}$  & 0.8049869853289970401 & $\theta_{1}$ & 0.7789940295960968708 & $\theta_{1}$ & 0.6182716140982345010 \\
    $\theta_{2}$  & 5.195067564321128373  & $\theta_{2}$ & 0.3324961120417712079 & $\theta_{2}$ & 0.4267721052272835380 \\
    $\theta_{3}$  & -0.2838953915400435069 & $\theta_{3}$ & 1.661755726450315995 & $\theta_{3}$ & 3.140349947502798944 \\
    $\theta_{4}$  & 0.1072488259756032430 & $\theta_{4}$ & -0.003102093041861720031 & $\theta_{4}$ & -0.006083654397070120678 \\
    $\theta_{5}$  & 1.571875317538424355  & $\theta_{5}$ & 1.329829265689388640 & $\theta_{5}$ & 1.388819097573269490 \\
    $\theta_{6}$  & -0.5763841724275414746 & $\theta_{6}$ & 1.222523233407654342 & $\theta_{6}$ & 1.247441611058169642 \\
    $\theta_{7}$  & 0.05617533738633816165 & $\theta_{7}$ & -0.9549223356382565697 & $\theta_{7}$ & -0.9777724100235600790 \\
    $\theta_{8}$  & 1.609687410067021096  & $\theta_{8}$ & 3.631512561877058953 & $\theta_{8}$ & 2.724926948310436803 \\
    $\theta_{9}$  & -3.184162587963997204 & $\theta_{9}$ & -0.2757196402806860180 & $\theta_{9}$ & -0.3285828895261063143 \\
    $\theta_{10}$ & 1.432467616732921778  & \multicolumn{2}{c}{} & \multicolumn{2}{c}{} \\
    $\theta_{11}$ & 0.02813968622180360382 & \multicolumn{2}{c}{} & \multicolumn{2}{c}{} \\
    $\theta_{12}$ & 0.07109716712678922079 & \multicolumn{2}{c}{} & \multicolumn{2}{c}{} \\
    $\theta_{13}$ & 0.04120009311925439122 & \multicolumn{2}{c}{} & \multicolumn{2}{c}{} \\
    \bottomrule
  \end{tabular}
\end{table}
\begin{table}[h!]
  \centering
  \caption{Calibrated model parameter sets for model transfer to Ogden material (full precision).}
  \label{tab:ogden}
  \begin{tabular}{@{}lr lr lr@{}}
    \toprule
    \multicolumn{2}{c}{Polyconvex NN} &
    \multicolumn{2}{c}{Relaxed ICNN} &
    \multicolumn{2}{c}{Unconstrained NN} \\
    \cmidrule(lr){1-2}\cmidrule(lr){3-4}\cmidrule(lr){5-6}
    \textbf{Param} & \textbf{Value} &
    \textbf{Param} & \textbf{Value} &
    \textbf{Param} & \textbf{Value} \\
    \midrule
    $\theta_{1}$  & 1.157275360655964258e+00 & $\theta_{1}$ & 5.645962189730805436e-01 & $\theta_{1}$ & 7.928536877065486266e-01 \\
    $\theta_{2}$  & 5.028109292028598354e+00 & $\theta_{2}$ & 1.982058889749746644e+00 & $\theta_{2}$ & 1.564181103306155896e+00 \\
    $\theta_{3}$  & -8.059457928952151740e-01 & $\theta_{3}$ & 4.711667943240553935e-01 & $\theta_{3}$ & 2.906866912983426587e+00 \\
    $\theta_{4}$  & 3.782093258970941063e-01 & $\theta_{4}$ & 7.098786078326054794e-01 & $\theta_{4}$ & 7.722878520017870119e-02 \\
    $\theta_{5}$  & 1.579239498975667289e+00 & $\theta_{5}$ & 3.000210769605224481e+00 & $\theta_{5}$ & 2.988771051719846916e+00 \\
    $\theta_{6}$  & -5.079725980572348254e-01 & $\theta_{6}$ & 3.316988866896799948e+00 & $\theta_{6}$ & 2.472296935852781097e+00 \\
    $\theta_{7}$  & 3.136053668569337982e-01 & $\theta_{7}$ & -8.509482812298809762e-01 & $\theta_{7}$ & -8.896302472541730566e-01 \\
    $\theta_{8}$  & 1.374119396630650414e+00 & $\theta_{8}$ & 3.471552104220098744e+00 & $\theta_{8}$ & 3.243040987696523381e+00 \\
    $\theta_{9}$  & -3.994605454095531361e+00 & $\theta_{9}$ & 2.552982794876417771e-01 & $\theta_{9}$ & -1.575907780538438718e+00 \\
    $\theta_{10}$ & 1.740566716704961658e+00 & \multicolumn{2}{c}{} & \multicolumn{2}{c}{} \\
    $\theta_{11}$ & 5.080664188423940353e-02 & \multicolumn{2}{c}{} & \multicolumn{2}{c}{} \\
    $\theta_{12}$ & 8.095671679163782275e-02 & \multicolumn{2}{c}{} & \multicolumn{2}{c}{} \\
    $\theta_{13}$ & 6.536249210448823177e-02 & \multicolumn{2}{c}{} & \multicolumn{2}{c}{} \\
    \bottomrule
  \end{tabular}
\end{table}
\begin{figure}[h!]
    \centering
    \includegraphics[width=0.8\linewidth]{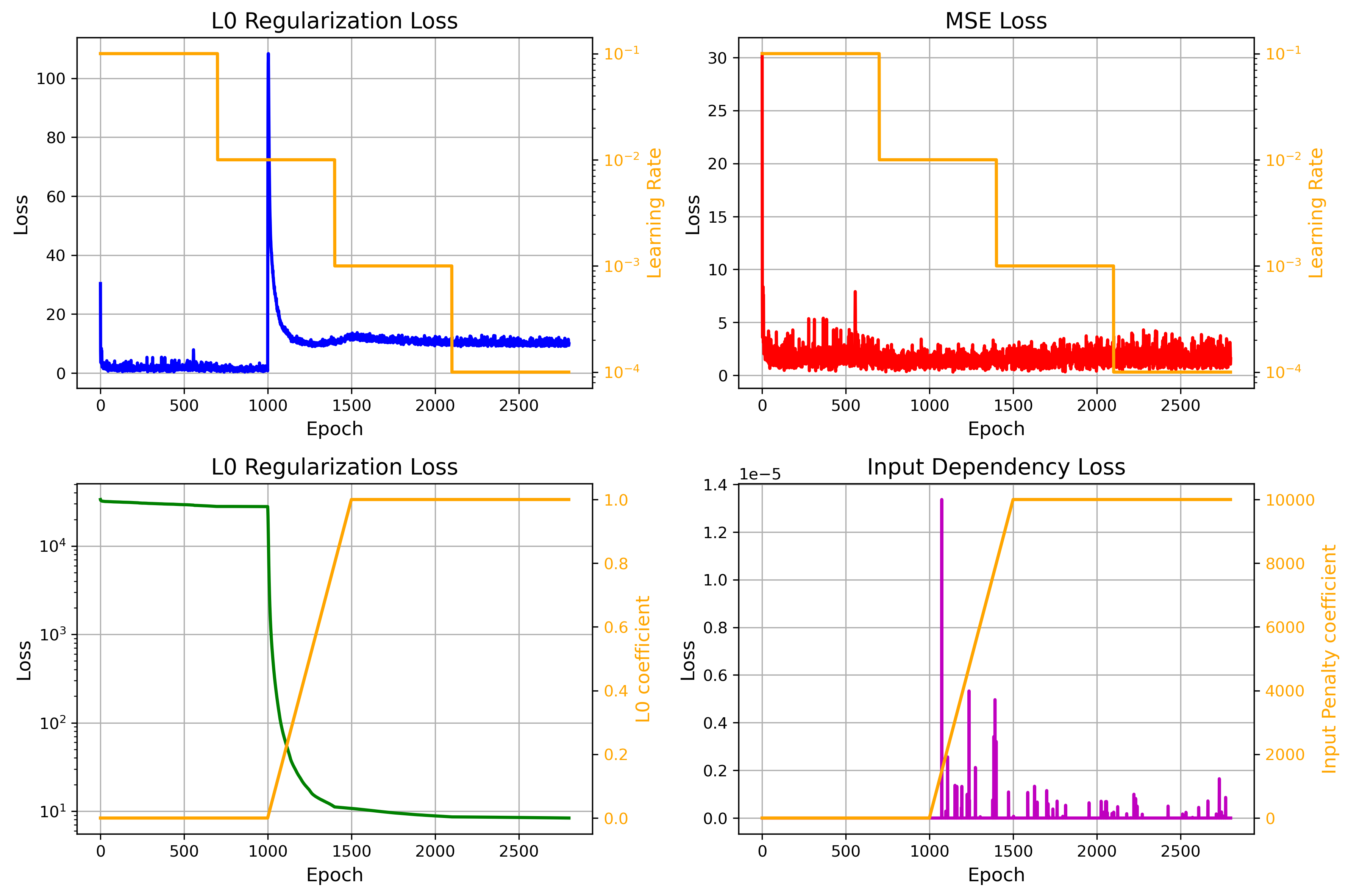}
    ~
    \includegraphics[width=0.8\linewidth]{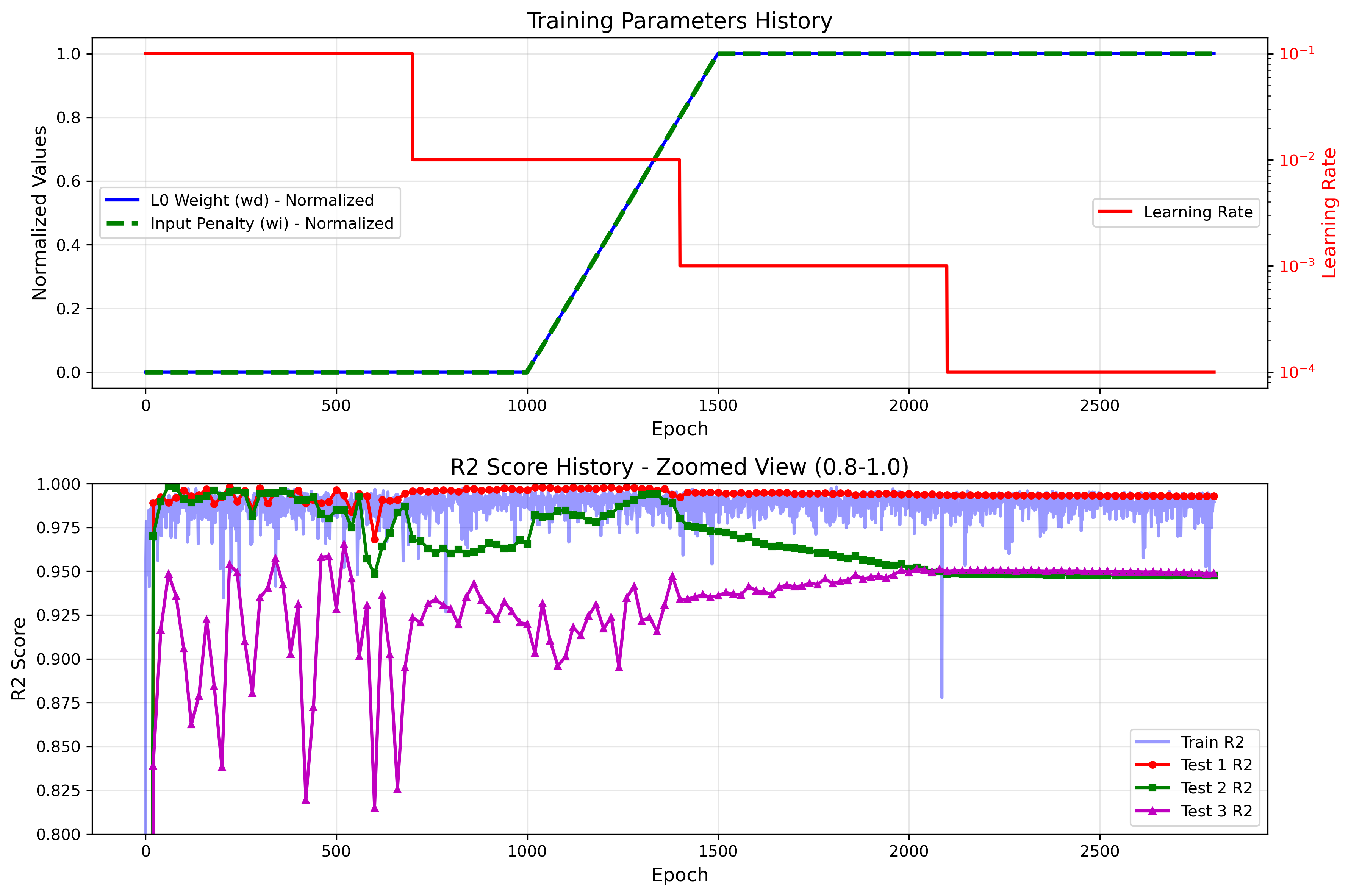}
    \caption{Losses and R2 score for Polyconvex PCNN model}
    \label{fig:history_poly}
\end{figure}
\begin{figure}[h!]
    \centering
    \includegraphics[width=0.8\linewidth]{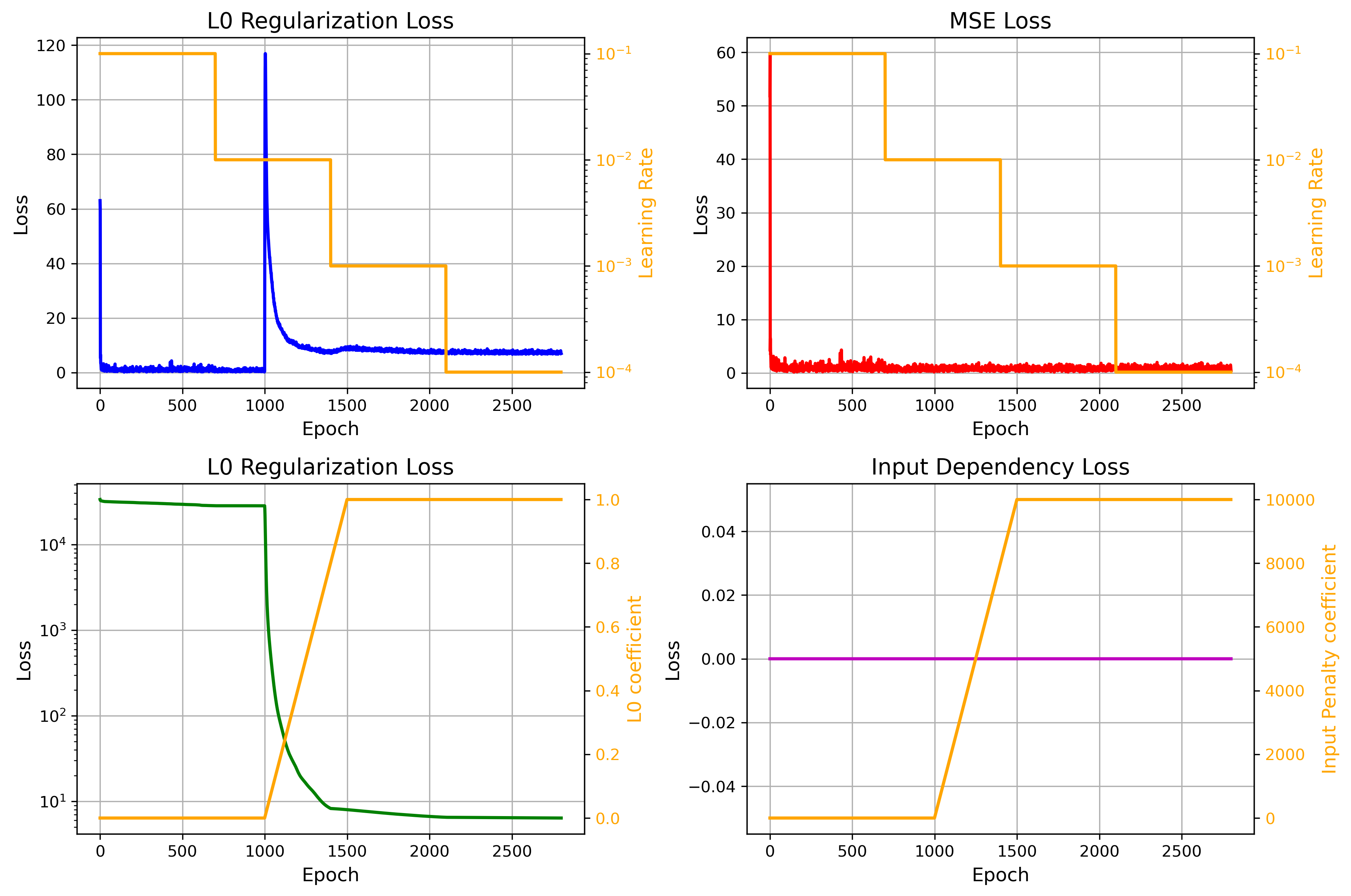}
    ~
    \includegraphics[width=0.8\linewidth]{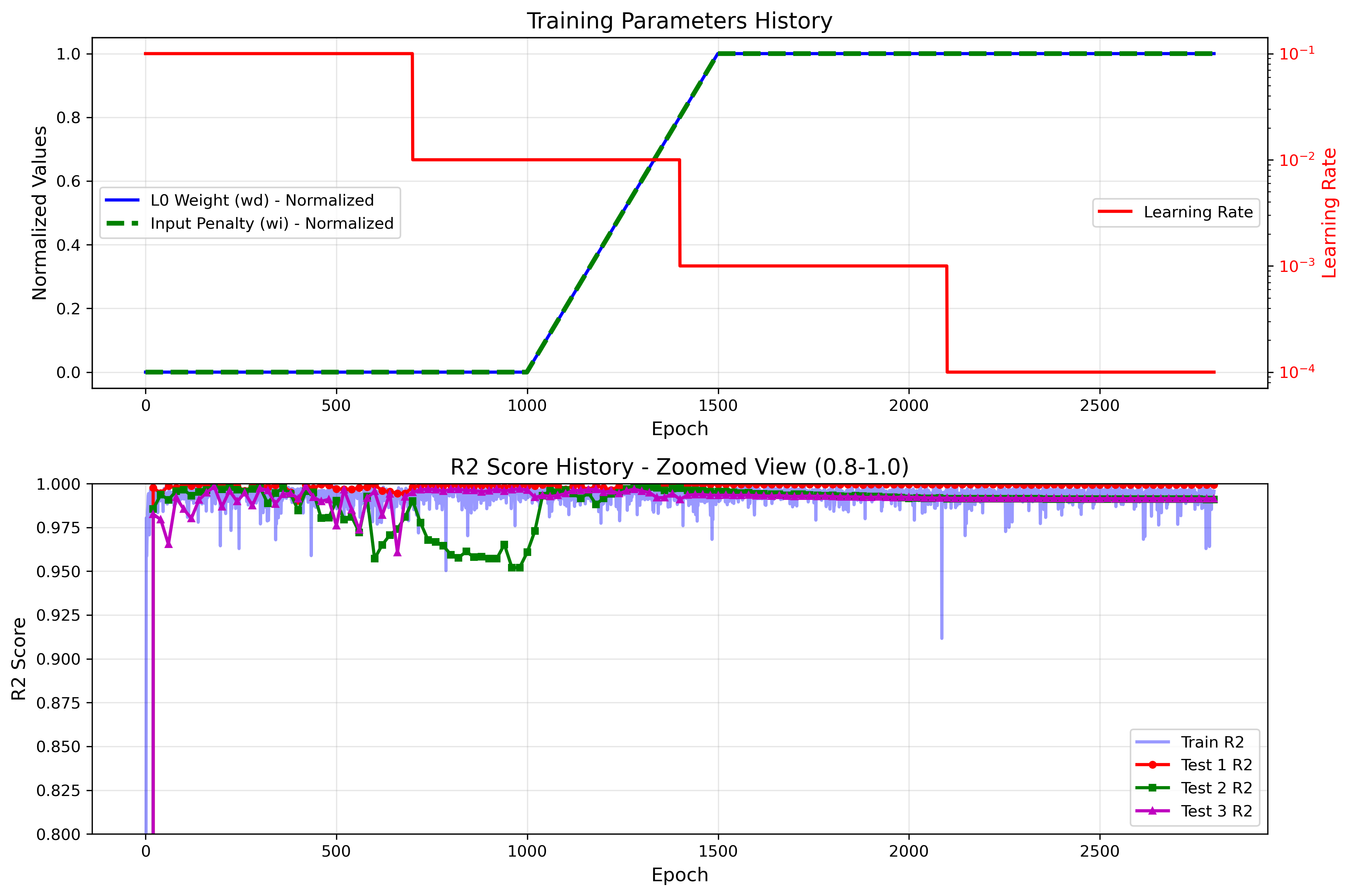}
    \caption{Losses and R2 score for Relaxed ICNN model}
    \label{fig:history_relax}
\end{figure}
\begin{figure}[h!]
    \centering
    \includegraphics[width=0.8\linewidth]{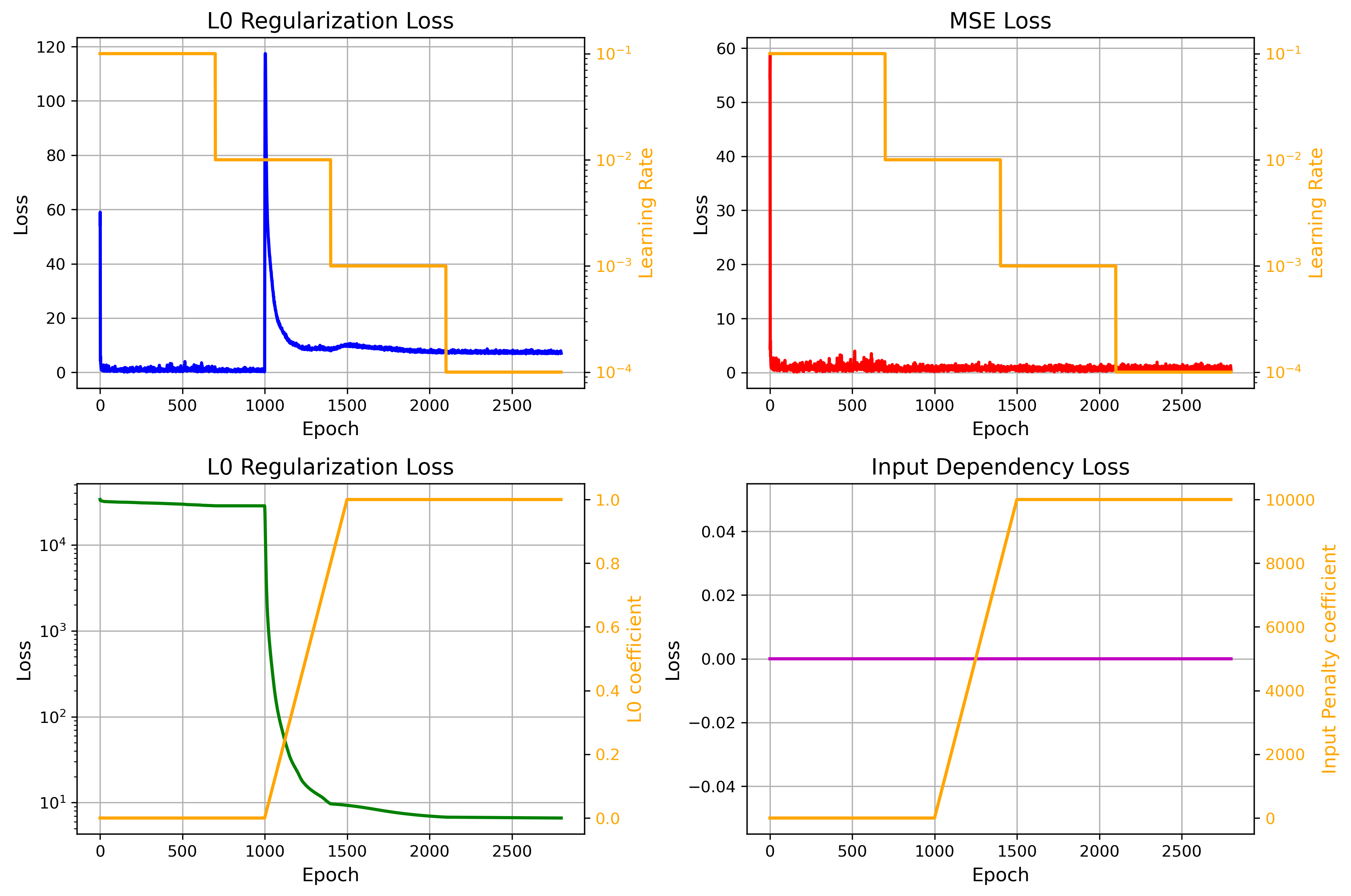}
    ~
    \includegraphics[width=0.8\linewidth]{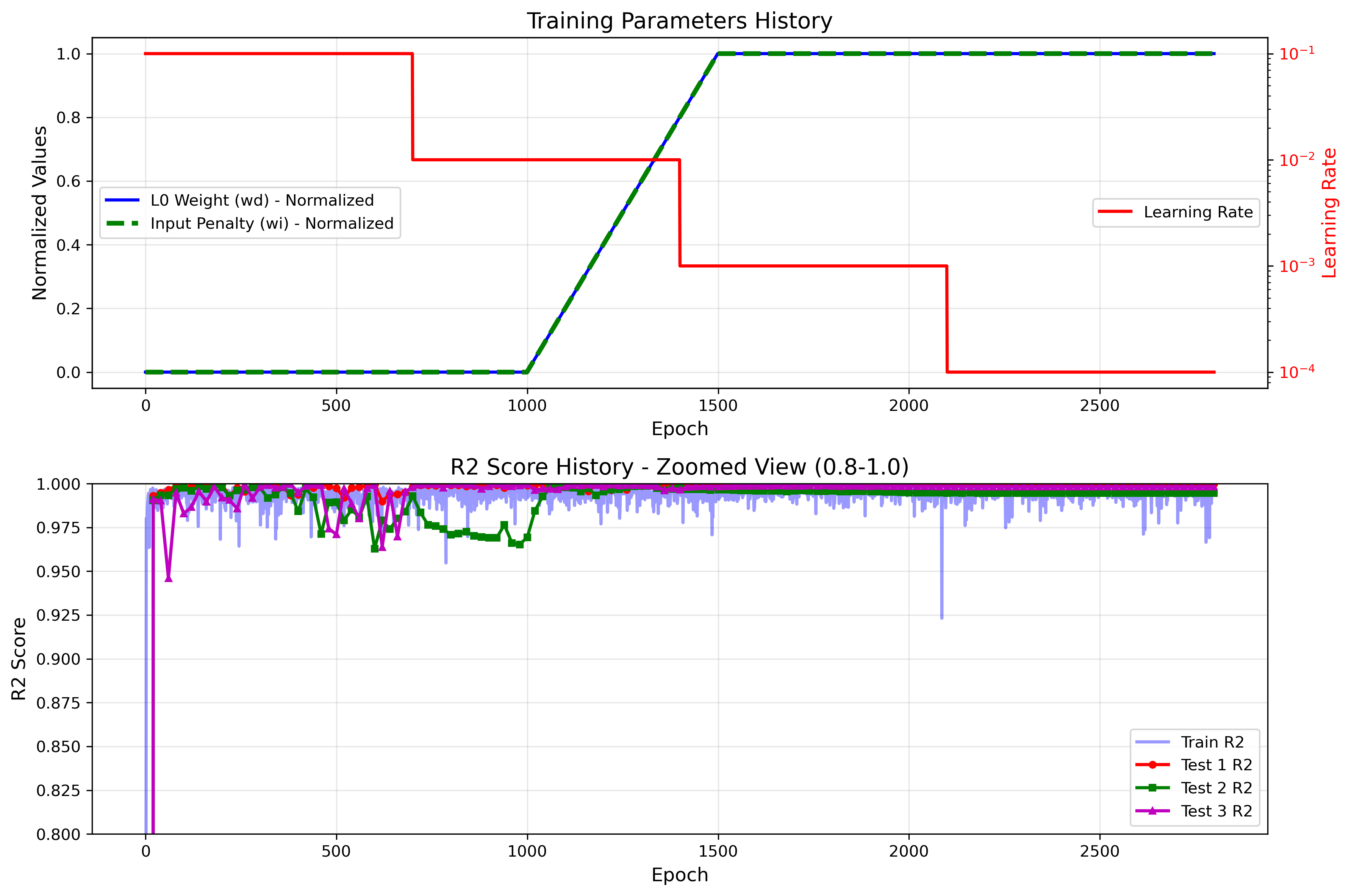}
    \caption{Losses and R2 score for Unconstrained NN model}
    
    \label{fig:history_unc}
\end{figure}
\clearpage
\bibliographystyle{unsrt}  
\bibliography{references}

\end{document}